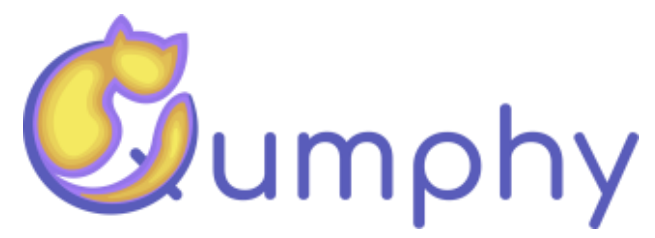

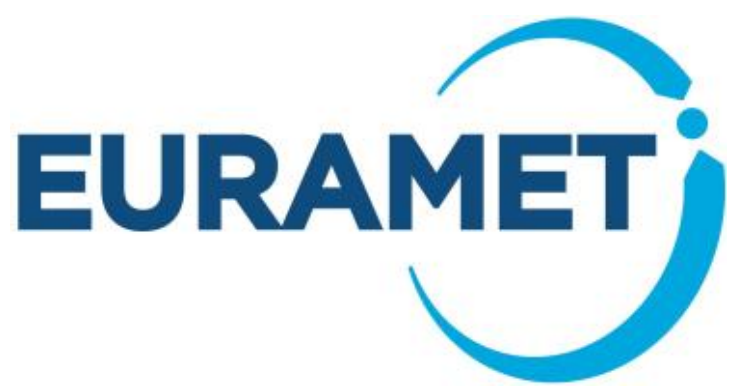

# 22HLT01 - QUMPHY

## Good Practice Guide for quantifying uncertainties for machine learning models applied to photoplethysmography signals

P. Harris, C. Bench, M. Rinkevičius, V. Marozas, L. Coquelin,
A. Thompson, M. Nandi, U. Hackstein, P.J. Aston

**Confidentiality Status:** PU - Public, fully open

Funded by the European Union. Views and opinions expressed are however those of the author(s) only and do not necessarily reflect those of the European Union or EURAMET. Neither the European Union nor the granting authority can be held responsible for them.

The project has received funding from the European Partnership on Metrology, co-financed from the European Union's Horizon Europe Research and Innovation Programme and by the Participating States. Funding for the UK participants was provided by Innovate UK under the Horizon Europe Guarantee Extension.

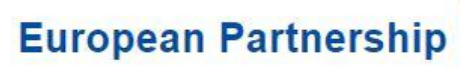

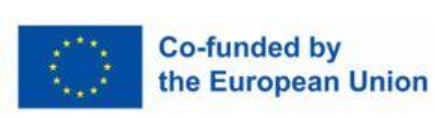

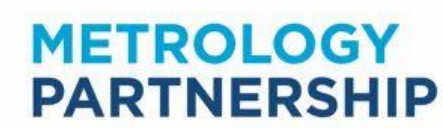

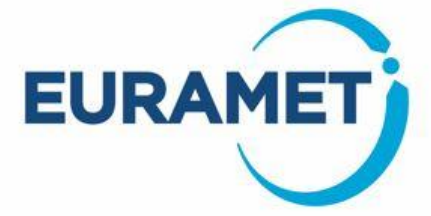

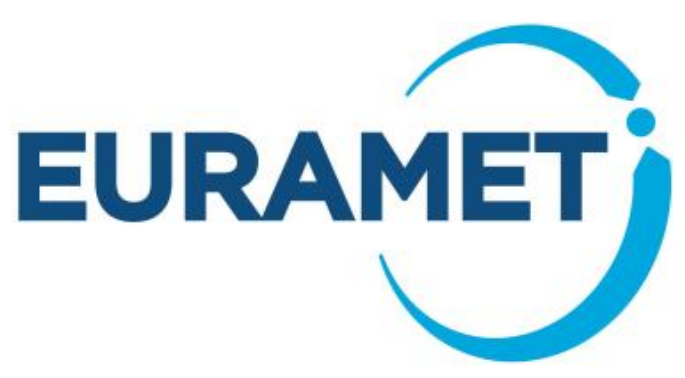

## Contributors

| | |
|---|---|
| PTB | Physikalisch-Technische Bundesanstalt |
| NPL | National Physical Laboratory |
| CMI | Cesky Metrologicky Institut |
| IMBiH | Institut za mjeriteljstvo Bosne i Hercegovine |
| IPQ | Instituto Português da Qualidade, I.P. |
| LNE | Laboratoire National de métrologie et d'Essais |
| FC | Qompium NV |
| FVB | Forschungsverbund Berlin e.V. |
| KTU | Kauno Technologijos Universitetas |
| THM | Technische Hochschule Mittelhessen |
| UGent | Universiteit Gent |
| Uni-Oldenburg | Carl von Ossietzky Universität Oldenburg |
| KCL | King's College London |
| SectorHealth | Sector Health Ltd |
| SURREY | University of Surrey |
| UCAM | University of Cambridge |

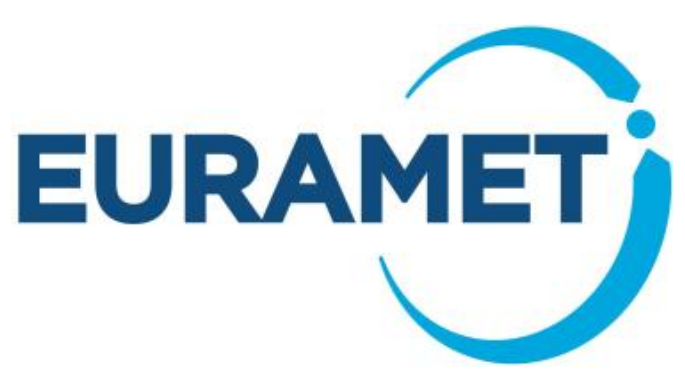

## Glossary

| | |
|---|---|
| ACE | Adaptive Calibration Error |
| AF | Atrial Fibrillation |
| APS | Adaptive Prediction Sets |
| BCE | Binary Cross Entropy |
| BP | Blood Pressure |
| CCE | Coverage Calibration Error |
| CNN | Convolutional Neural Network |
| CRPS | Continuous Ranked Probability Score |
| CWT | Continuous Wavelet Transform |
| CVD | Cardiovascular Disease |
| DBP | Diastolic Blood Pressure |
| DE | Deep Ensemble |
| DL | Deep Learning |
| DWPT | Discrete Wavelet Packet Transform |
| DWT | Discrete Wavelet Transform |
| EC | European Commission |
| ECE | Expected Calibration Error |
| ECG | Electrocardiogram |
| ENCE | Expected Normalised Calibration Error |
| FAIR | Findable, Accessible, Interoperable, and Reusable |
| GDPR | General Data Protection Regulation |
| GP (model) | Gaussian Process (model) |
| GPG | Good Practice Guide |
| ICA | Independent Component Analysis |
| ITU | International Telecommunication Union |
| LASSO | Least Absolute Shrinkage and Selection Operator |

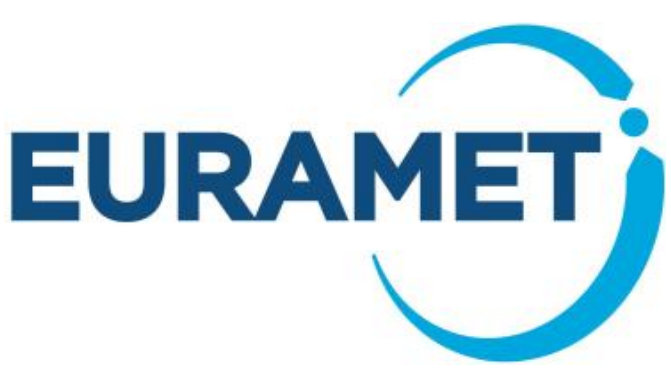

| | |
|---|---|
| LIME | Local Interpretable Model-agnostic Explanations |
| MAP (estimate) | Maximum A Posteriori (estimate) |
| MCD | Monte Carlo Dropout |
| MCE | Maximum Calibration Error |
| ML | Machine Learning |
| MLE | Maximum Likelihood Estimate |
| MLP | Multi-Layer Perceptron |
| MPIW | Mean Prediction Interval Width |
| MRMR | Minimum Redundancy Maximum Relevance |
| NLL | Negative-Log-Likelihood |
| NN | Neural Network |
| PCA | Principal Component Analysis |
| PICP | Prediction Interval Coverage Probability |
| PMF | Probability Mass Function |
| PP (interval) | Peak-to-Peak (interval) |
| PPFS | Predictive Permutation Feature Selection |
| PPG (signal) | Photoplethysmogram (signal) |
| QR | Quantile Regression |
| ReLU | Rectified Linear Unit |
| RGB (colour map) | Red, Green and Blue (colour map) |
| RNN | Recurrent Neural Network |
| SBP | Systolic Blood Pressure |
| SHAP | SHapley Additive exPlanations |
| smECE | smoothed Expected Calibration Error |
| SPAR (method) | Symmetric Projection Attractor Reconstruction (method) |
| SVM | Support Vector Machine |
| TCN | Temporal Convolutional Network |
| UCE | Uncertainty Calibration Error |

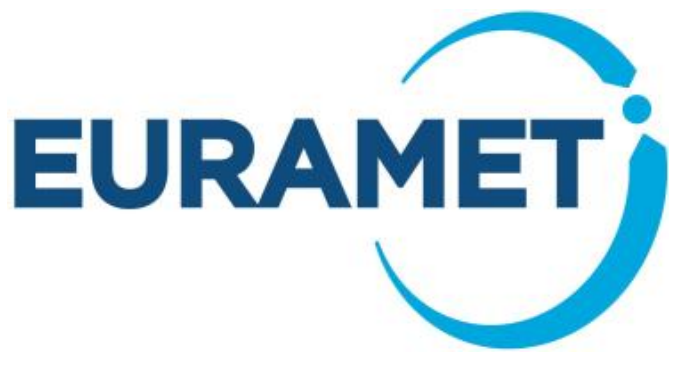

| | |
|---|---|
| UQ | Uncertainty Quantification |
| VCE | Variation Calibration Error |
| WHO | World Health Organisation |

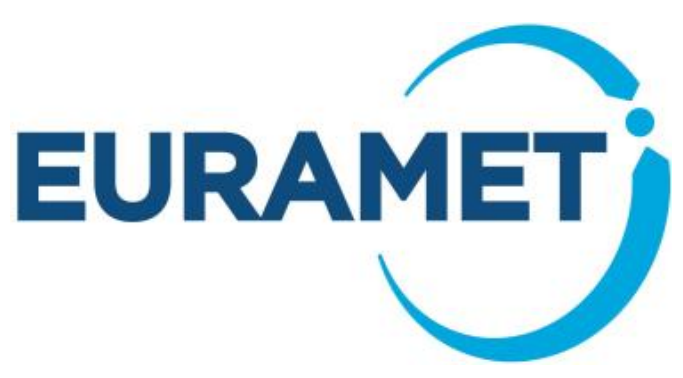

**TABLE OF CONTENTS**

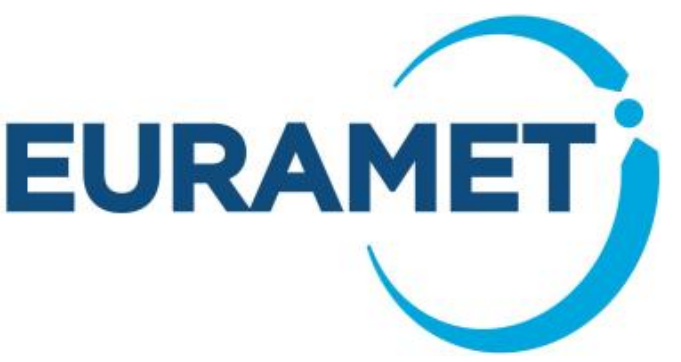

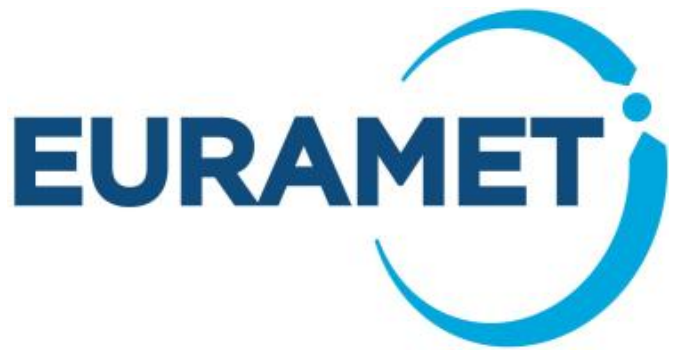

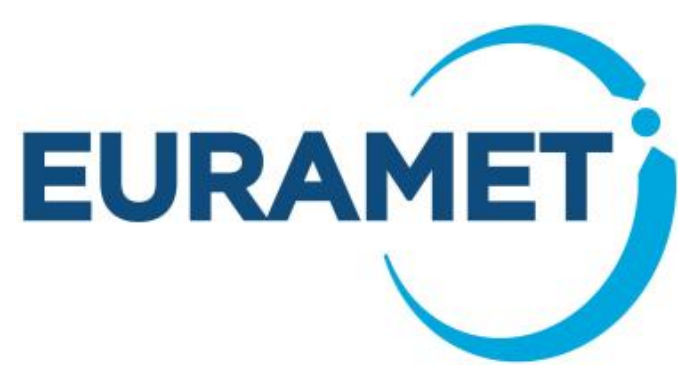

# 1 Introduction

## 1.1 Background

Photoplethysmography is a widely used physiological sensing technique. It consists of shining light (typically red, green or infrared) onto the skin, and measuring the amount of light either reflected from, or transmitted through, a region of tissue. The simplicity and non-invasive nature of the technique, as well as its ability to provide information about physiological parameters, make it an attractive sensing modality, and it is widely used in clinical devices such as pulse oximeters and consumer devices such as smartwatches.

The *photoplethysmogram* (PPG) signal measures the fluctuations in blood volume which occur with each heartbeat. PPG signals contain a wealth of physiological information relating primarily to the heart and blood vessels. Figure 1 (adapted from [1]) shows an idealised PPG signal: the shape of the pulse wave contains information relating to blood flow, and the inter-beat intervals are related to heart rhythm.

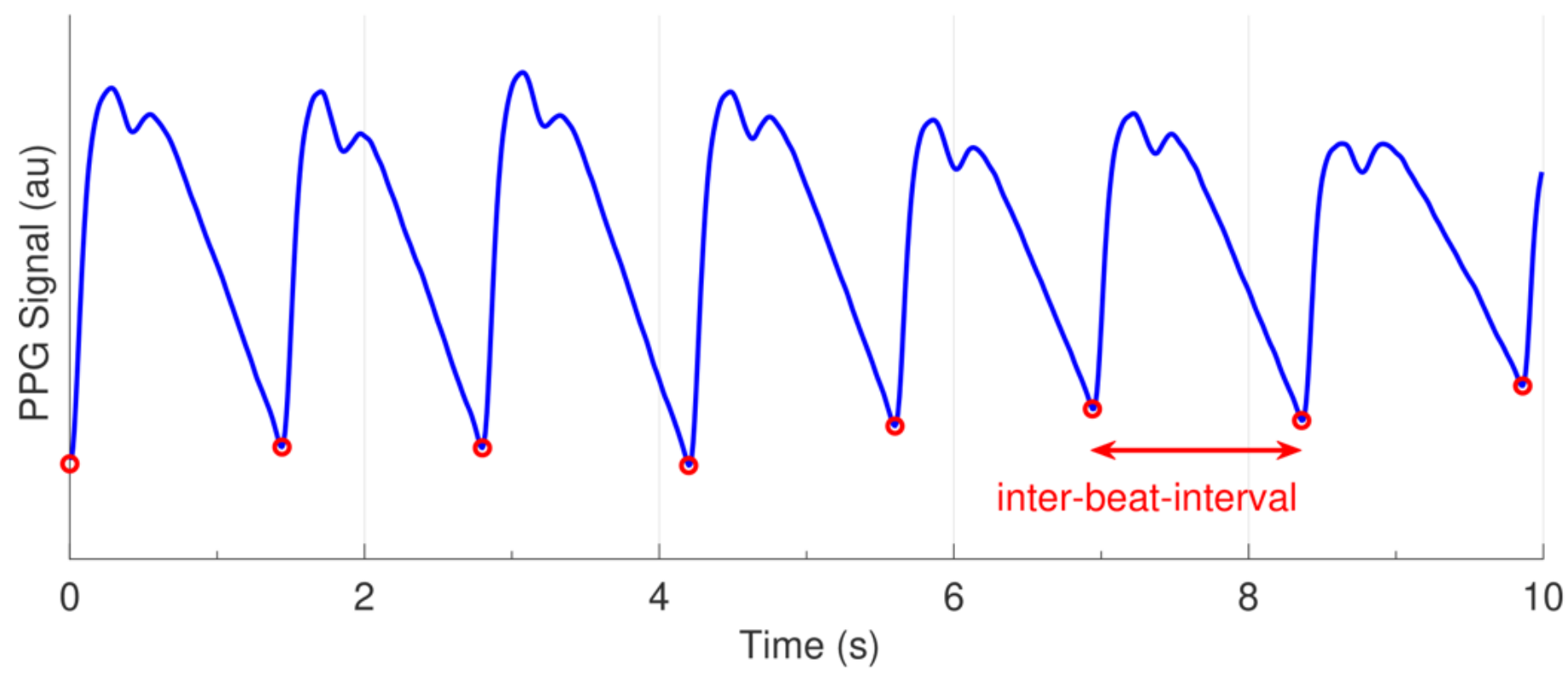

Figure 1: An idealised PPG signal showing a pulse wave for each heartbeat. Pulse onsets, representing individual heartbeats, are shown as red circles. An inter-beat interval is labelled, corresponding to the time between consecutive heartbeats. (Adapted from [1].)

*Machine learning* (ML) is often applied using PPG signal inputs in order to estimate various physiological parameters, such as *blood pressure* (BP), or to detect diseases associated with the cardiac system, such as *atrial fibrillation* (AF). Blood pressure estimation involves the estimation of a quantitative variable, often referred to as a *regression task*. Atrial fibrillation detection, on the other hand, involves the prediction of a categorical variable, often referred to as a *classification task*. Section 1.3 lists examples of problems of clinical relevance, treated within the QUMPHY project as benchmark problems, that can be addressed by a combination of ML and PPG signal inputs. Three approaches are commonly used: *deep learning* (DL) using the raw PPG signal as input; DL using the PPG signal converted into an image representation such as a scalogram; ML using features extracted from the PPG signal that relate to the signal morphology or variability. The last of these approaches is attractive as the features are easy to understand and interpret, as well as generally being computationally less costly compared to the other approaches. However, this reductionist approach will not necessarily include all the morphological features relevant to predicting the physiological parameters of interest and so results can be comparatively poor unless the most relevant information in the signal for predicting those parameters can be identified. In contrast, DL approaches using either the raw signals or image representations make use of the entire signal, extracting the most relevant features on their own, and so often give better results but are not so easy to interpret.

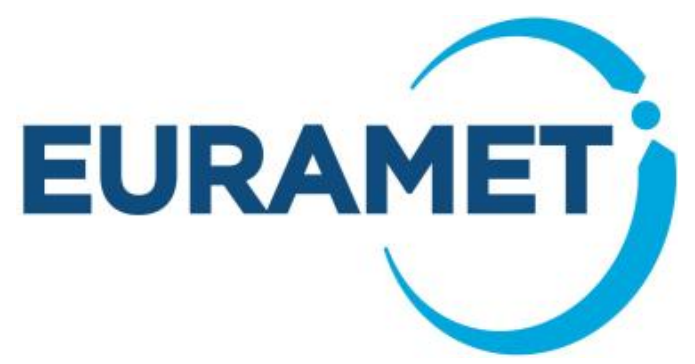

Whichever approach to analysing a PPG signal is adopted, a model output in the form of a prediction is incomplete without a quantitative statement of the quality of the prediction in terms of an uncertainty. The uncertainty is needed to establish whether the prediction can be used to reliably inform diagnosis and as the basis of decision-making. But to realise the practical benefits of using the prediction, it is essential that the uncertainty is reliable and conveys meaningful information about the prediction. For example, when the predicted quantity is a physiological parameter, which is quantitative, the uncertainty should reflect the ‘true’ dispersion of the possible values of the quantity based on the available knowledge about the quantity. Indeed, the quality of uncertainty values must be assessed carefully, and should cater to the practical realities of the model’s use, in particular when just a single measurement, or a few measurements, will be used to inform diagnosis. It follows that the practical realisation of the benefits that wearable PPG-sensing technologies could provide to patients hinges on the development of robust *uncertainty quantification* (UQ) methods and a framework for the *validation of uncertainty*.

As a simple illustration, consider that a quantity describing a physiological parameter for a patient is measured to decide if a specification limit is exceeded. In the case that the limit is exceeded, the clinician will decide to treat the patient by administering a drug and, otherwise, no treatment is started. The decision has health implications for the patient as well as cost implications for the clinician who might control the financial budget that pays for the drug. Figure 2 shows four scenarios.

In scenario (a), only the prediction of the quantity is available to the clinician. Under the weak assumption that the quantity is as likely to be below the prediction as above it, the clinician can conclude that the probability of the quantity exceeding the specification limit is greater than 0.5. But that is all they can conclude and, without any quantitative information about the quality of the prediction, they cannot say whether that probability is nearer to 0.5 or to 1.0. If the clinician chooses to make a decision based on the prediction alone, they have no way to judge how confident they are in the decision.

In scenarios (b) and (c), the prediction is accompanied by a statement of uncertainty in the form of an ‘uncertainty bar’, which defines an interval that contains the value of the quantity with a stated probability (0.95, say). In scenario (c), the clinician concludes with high confidence (with probability greater than 0.95) that the quantity exceeds the specification limit and decides to start treatment. In scenario (b), the clinician cannot conclude with high confidence that the limit is exceeded. But that is still useful information, and they might then decide to collect more information, for example, by undertaking further measurements, or to collect better information, for example, by repeating the measurement with a more accurate measuring device in order to reduce the uncertainty.

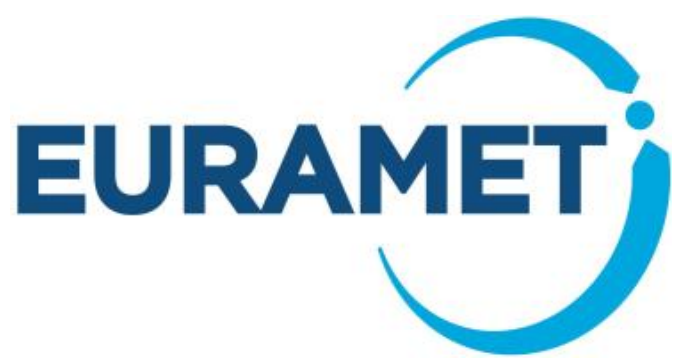

The information in scenario (c) is a summary of the more complete information about the prediction depicted in scenario (d), which is in the form of a probability distribution. This information emphasises that even in scenario (c) there is a small probability that the value of the quantity falls below the specification limit, and the probability distribution allows that probability to be evaluated. The probability quantifies the risk of the clinician making an incorrect decision and administering a drug unnecessarily.

However, good decisions depend on the reliability of the uncertainty in whatever form it is provided. For example, in scenario (c), if the stated probability for the uncertainty bar shown is 0.80 (rather than 0.95) then the information is misleading. The clinician is more confident in their decision than they should be, and they do not correctly understand the risks from making a wrong decision.

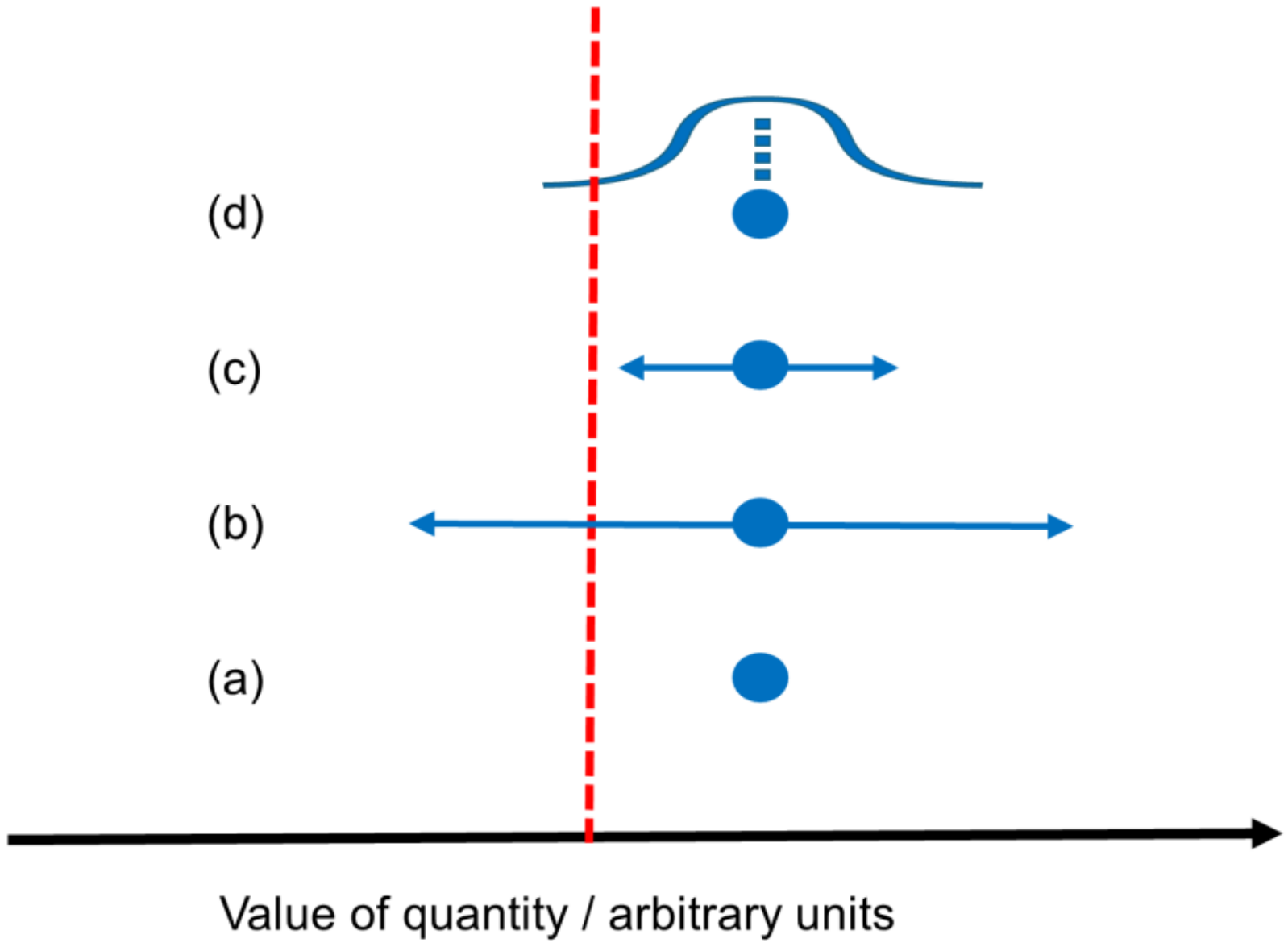

Figure 2: Illustration of four measurement results (in blue) to be used to decide whether the value of a quantity, such as a physiological parameter, exceeds a specification limit (dashed red line).

## 1.2 Aims and Organisation of the Good Practice Guide

This *Good Practice Guide* (GPG) is aimed at *providers* of information derived from PPG signals, such as device manufacturers who want to ensure that model predictions are accompanied by reliable statements of uncertainty. It is also aimed at *users* of such information, such as clinicians who want to understand the basis of the provided information and how to interpret it. The audience of the GPG is not expected to be an expert in ML and UQ but to have a working knowledge of both topics and of some methods for ML and UQ.

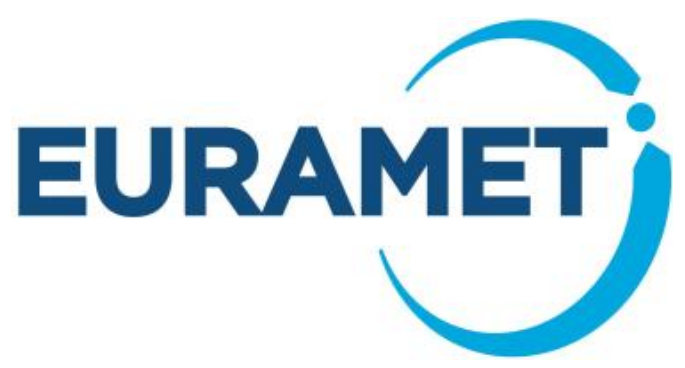

The GPG provides high-level guidance on what types of ML model might be used and how different models compare when applied to different tasks, both regression and classification tasks (section 2). It provides guidance on the implementation of different methods for UQ, covering both model-dependent and model-independent techniques (section 3) and on the validation of the results provided by those methods for UQ (section 4). The GPG also provides practical support to practitioners by providing pointers to different benchmark datasets (section 5) for the benchmark problems listed in section 1.3, and introduces software that can assist practitioners in implementing the guidance of this GPG (section 7). Considerations of ethics are discussed in section 8, and a summary together with recommendations is given in section 9.

For more in-depth presentations on these topics, the reader is referred to the following deliverables of the QUMPHY project and references contained therein.

- Paper on supervised ML and/or DL models applied to PPG signals for one prototypical regression (estimation of blood pressure) and one prototypical classification problem (detection of atrial fibrillation), including a comparison of model performance [2].
- Paper on methods of uncertainty quantification for one prototypical regression (estimation of blood pressure) and one prototypical classification problem (detection of atrial fibrillation), including evaluations of uncertainty of model predictions [3].
- Report on validating the uncertainties obtained for ML and/or DL models [4].
- Report describing six datasets using real or synthetic PPG data that can be used to benchmark accuracy and uncertainty of supervised ML and DL models applied to PPG signals [5].

## 1.3 Benchmark Problems

As cardiovascular disease (CVD) remains a leading global health challenge, the limitations of traditional diagnostic models, which often rely on infrequent "snapshot" data collected during brief clinic visits, have become increasingly apparent. Photoplethysmography offers a transformative approach by utilising the optical sensors already embedded in community wearables.

Unlike traditional electrocardiogram (ECG) patches or Holter monitors that can be cumbersome and stigmatising, PPG-enabled devices allow for 24/7 longitudinal data collection without overtly disrupting the user’s routine. This continuous stream of data enables the detection of subtle physiological shifts, such as an increased burden of atrial fibrillation or clinically concerning changes in arterial stiffness that could otherwise be missed. However, PPG data from wearables can be subject to artefacts [6] caused by sensor displacement and poor contact, which could lead to false positive or false negative results. Therefore, further research is needed to determine whether ML models trained on raw PPG data or those incorporating feature extraction and signal quality evaluation are more effective in reducing uncertainty.

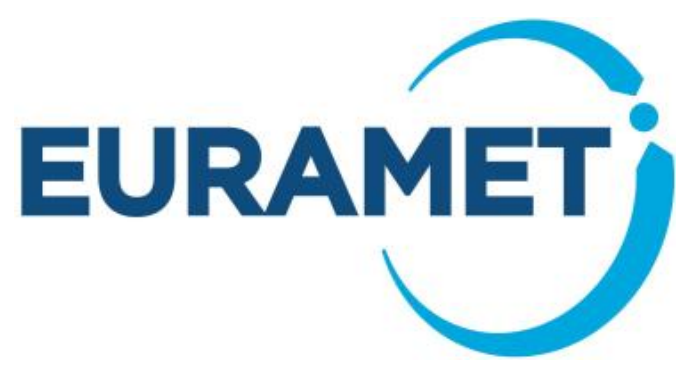

The report [5] describes six benchmark problems covering both regression and classification tasks applied to PPG signals. For each benchmark problem the report identifies benchmark datasets that are designed to support the development and evaluation of ML models and UQ methods for PPG signal analysis (see section 5). Here, only high-level summary descriptions of the benchmark problems are provided.

#### 1.3.1 Determination of Systolic and Diastolic Blood Pressure

BP is one of the most commonly taken physiological measurements as it is used to monitor health, predict adverse events and is essential for the selection and monitoring of patients and their response to treatments in numerous clinical settings [7]. BP measurement involves the determination of systolic BP and diastolic BP, and the estimation of these BPs from PPG signals is a regression task.

#### 1.3.2 Detection of Atrial Fibrillation

More than 33 million people worldwide are diagnosed with AF, and this number is projected to double by 2050 [8]. AF is a risk factor for stroke, hence early detection could support more timely clinical management. AF can be asymptomatic and difficult to detect due to its paroxysmal nature [9]. Additionally, irregular rhythms such as premature atrial contractions, bigeminy, atrial flutter, and tachycardia contribute to false positives [10][11]. Opportunistic pulse palpation followed by a twelve-lead ECG is a recommended screening method for individuals over 65 but is ineffective for asymptomatic cases as it is performed only when symptoms are reported. Therefore, developing technologies for screening high-risk populations in a convenient and affordable way is crucial [12].

The development of methods for the reliable detection of AF using PPG signals is hindered by the limited availability of high-quality labelled datasets. The lack of guidelines for arrhythmia interpretation in photoplethysmography necessitates alternative annotation strategies such as the simultaneous acquisition of ECGs. As a result, existing PPG datasets are either too small or may contain inaccurate labels due to reliance on automated annotations rather than expert labelling. The detection of AF from PPG signals is a (binary) classification task.

#### 1.3.3 Classification of Hypertension

Elevated BP and hypertension have a very high prevalence, with about 30 % of people in Europe affected by the condition. It is one of the major risk factors for many diseases, including hypertension-mediated organ damage, stroke, cognitive impairment, heart failure, AF and diabetes [7]. Three classes of blood pressure levels are defined in terms of values of systolic BP (SBP) and diastolic BP (DBP) in accordance with the 2024 ESC Guidelines [7] as follows: non-elevated (SBP < 120 mmHg and DBP < 70 mmHg); elevated (120 ≤ SBP < 140 mmHg or 70 ≤ DBP < 90 mmHg); hypertensive (SBP ≥ 140 mmHg or DBP ≥ 90 mmHg). This is a three-class classification task.

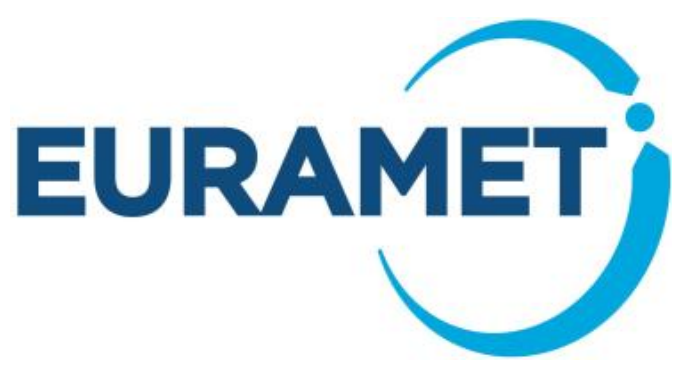

### 1.3.4 Determination of Vascular Age

Vascular aging is a complex process that involves the gradual deterioration of arterial structure and function over time and negatively impacts organ function [13]. The gold standard measurement for vascular age is carotid-femoral pulse wave velocity, which requires trained personnel and is not routinely available [14]. In healthy ageing, a person's vascular age typically corresponds to their chronological age [15]. Early detection of premature vascular aging is critical for the timely identification and treatment of cardiovascular disease. Photoplethysmography can help to determine vascular age by analysing the shape of the PPG pulse wave, which changes with age due to arterial stiffening. By comparing signal-based estimates of vascular age to a person's chronological age, it is hypothesised that early-stage vascular aging might be diagnosed in community-based settings allowing primordial prevention strategies to be implemented. We allocated subjects into decadal groups, thus converting this essentially regression task into a multiclass classification task.

### 1.3.5 Detection of Sleep Apnoea

Obstructive sleep apnoea is a prevalent sleep disorder characterised by repeated episodes of partial or complete upper airway obstruction during sleep, leading to disrupted breathing, oxygen desaturation, and fragmented sleep [16][17]. Untreated obstructive sleep apnoea is associated with a range of comorbidities, including hypertension, cardiovascular disease, stroke, type 2 diabetes, and neurocognitive impairments such as memory deficits and impaired executive function [18][19]. In addition, the condition contributes to daytime fatigue, thus increasing the risk of car and workplace accidents [20]. The chronic physiological stress from recurrent apnoea and hypopnea triggers systemic inflammation and oxidative stress, further exacerbating metabolic and cardiovascular complications. The condition's impact on quality of life, coupled with its economic burden due to healthcare costs and lost productivity, underscores the need for timely diagnosis and effective management [21].

Polysomnography, the gold standard for diagnosis, is resource-intensive, requiring overnight monitoring in a sleep laboratory with specialised equipment and trained personnel [22]. Home sleep apnoea testing offers a more accessible alternative, but its sensitivity and specificity vary, and it may miss milder cases or misclassify complex sleep disorders [23]. Moreover, many individuals with obstructive sleep apnoea remain asymptomatic or attribute symptoms like snoring or daytime sleepiness to other causes, delaying presentation to healthcare providers. Public awareness of obstructive sleep apnoea is low, and screening tools, such as questionnaires, while useful, lack precision and can lead to over- or under-referral for testing [24][25]. These barriers to detection highlight the need for scalable, cost-effective diagnostic solutions, such as wearable devices with AI-driven screening tools, to improve early identification and treatment. This is another classification task.

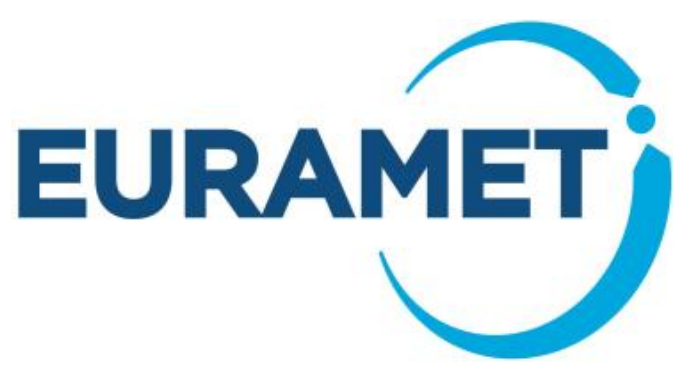

### 1.3.6 Determination of Respiratory Rate

Respiratory rate provides much information about a patient's physiological state [26][27] and is one of the most sensitive indicators of clinical deterioration [28][29][30][31]. It is known that physiological mechanisms cause ECG and PPG signals to be modulated by respiration [32][33], which motivates the possibility to gain insight about respiratory rate from the analysis of PPG signals. Indeed, approaches to measure respiratory rate have been incorporated recently into consumer wearables that record PPG signals, and it has become important to investigate the performance of such measurement of respiratory rate and to optimise it. This is a regression task.

# 2 Models for Supervised Learning

## 2.1 Background

*Supervised learning* is a type of ML where models are built to make predictions of some (output) variable given knowledge of other (input) variables. The key distinctive feature of supervised learning models is that they are learned from *training data*, which comprises values for the output variable corresponding to values of the input variables. Two types of supervised learning are often distinguished: *regression* and *classification*, and the distinction is based upon the type of variable that is being predicted, whether it is a continuous, quantitative variable or a discrete, categorical variable.

Good practice [34] involves splitting the training data into data to optimise the parameters of the model (the *training set*), to evaluate the performance of the model during the training phase (the *validation set*), and to evaluate the performance of the final model obtained from the training phase (the *test set*). For each cycle of the optimisation, the training set enables the model learning, and the validation set is used to monitor the model performance, such as detecting overfitting and early stopping to prevent wasted computation. A variety of metrics are used to quantify the performance of the model during the testing phase. Additionally, the split of the data into training, validation and test sets may also comprise an additional set, a so-called *calibration set*, which is required by some of the methods for UQ considered in this report.

For a regression problem, $y \in \mathcal{R}$ denotes the value of a quantitative output variable corresponding to a value $x \in \mathcal{R}^d$ of the input variables. For example, $y$ might be the (systolic or diastolic) BP of an individual and $x$ might comprise values of $d$ features extracted from a PPG signal recorded by a smartwatch worn by the individual. The training set is composed of input values $\{x_1, \ldots, x_N\}$, for example recorded for different individuals, and corresponding numeric labels $\{y_1, \ldots, y_N\}$ that are obtained by measurement of the output variable for those individuals, for example recorded by a BP monitor. The prediction task is to determine an estimate $y$ of the output variable for a new individual from the value $x$ of the input variables recorded for that individual.

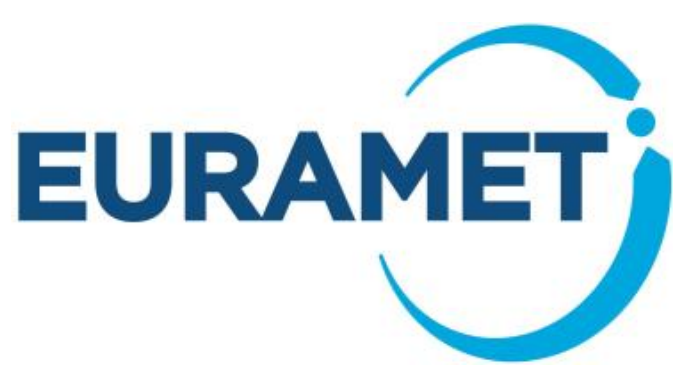

For a classification task, $y \in \{1, 2, \dots, K\}$, where $K$ is the number of classes, denotes the value of an unordered categorical output variable that identifies the class to which the corresponding value $x \in \mathcal{R}^d$ of the input variables belongs. For example, $y \in \{1, 2\}$ might represent whether a subject is experiencing AF ($y = 2$) or not ($y = 1$) based on values of $d$ features extracted from a PPG signal recorded by a pulse oximeter. The training set is composed of input values $\{x_1, \dots, x_N\}$, for example recorded for different subjects, and corresponding class labels $\{y_1, \dots, y_N\}$ that are obtained by examination by a clinician, perhaps based on the output from an independent measurement, such as that made using a blood pressure monitor, that determines a ground truth. The prediction task is to determine an estimate $y$ of the class label for a new subject from the value $x$ of the input variables for that subject.

There are three common approaches to applying ML to these sorts of tasks using PPG signals, which are discussed in the following sections.

## 2.2 Deep Learning Applied to Raw Signals

Recent advancements in DL have significantly impacted healthcare by enabling the analysis of raw physiological data. The key advantage of DL is its ability to automatically recognise complicated patterns directly from raw data such as PPG signals, thus eliminating the need for extensive manual feature engineering and extraction. Architectures for DL models, such as *Convolutional Neural Networks* (CNNs) and *Recurrent Neural Networks* (RNNs), have shown remarkable performance in tasks such as BP estimation by capturing the temporal nuances inherent in raw physiological signals. DL models are being used for many essential healthcare tasks, such as predicting patient outcomes, diagnosing various cardiovascular conditions, and monitoring chronic diseases. DL benefits from the availability of large and diverse datasets, which help in training robust and generalisable models. As the availability of high-quality physiological data increases, DL is expected to play a crucial role in advancing personalised medicine and improving patient care.

Some examples of DL models used for regression and classification tasks adapted to time-series data include:

1. *LeNet1d* [35], which is a one-dimensional CNN adapted from the original *LeNet* model architecture, and optimised for time-series data. By considering a greater number of convolutional and pooling layers it is more appropriate for extracting features from univariate time-series data.
2. *Inception1d* [36], which is inspired by the *Inception* model architecture but adapted for one-dimensional signals. The model uses parallel convolutional layers with different kernel sizes to capture information at multiple scales enabling the model to understand complex patterns in time-series data.

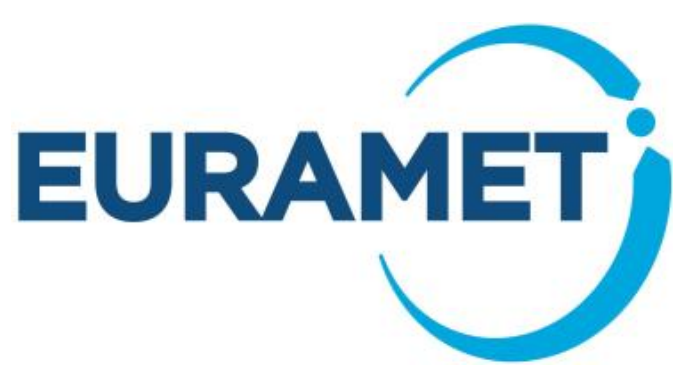

3. *XresNet1d50* and *XResNet1d101* [37], which represent state-of-the-art models for DL derived from the *ResNet* model architecture. The models incorporate more layers (50 and 101, respectively) and include a number of other enhancements, such as group normalisation and the option for selective kernel sizes. Such enhancements allow the models to learn more intricate hierarchical features and are very suited to tasks involving complex time-series data.
4. *AlexNet1d* [38], which is a one-dimensional adaptation of the popular *AlexNet* CNN model architecture traditionally used for two-dimensional image classification. The model consists of two primary components. The first component is a feature extractor comprising a stacked series of convolutional, rectified linear unit (ReLU), and 'max pooling' layers. The second component is a classifier comprising a fully connected *neural network* (NN) made of two linear layers, each followed by a ReLU activation function and a dropout layer. The final layer is a linear layer whose output is the model prediction.
5. *MiniRocket* [39], which is an almost deterministic transform for time-series data followed by a linear classifier. The model uses dilated convolutions with kernels of length nine and all possible permutations of the weights $-1$ and $2$ that sum up to zero, where the dilation values are selected to be within a fixed range relative to the input length. Following the convolutions, features are extracted by calculating the proportion of positive values in each convolution output, which is the number of positive values divided by the sequence length. This process yields a total of 9,996 features, which are then fed into a linear regressor to predict, for example, the mean and standard deviation of a Gaussian predictive probability distribution. The kernels do not require training, allowing for a one-time feature extraction. This approach enables efficient model training as only the linear regressor needs to be trained, and makes the model a very fast and effective baseline for time-series classification and regression tasks.
6. *Temporal Convolutional Networks* (TCNs) [40], which are a class of NNs designed for sequential data processing. Their ability to capture both short-term and long-term dependencies makes them particularly effective for analysing complex time-series data. TCNs leverage the strengths of CNNs and adapt them for temporal data. Unlike RNNs that process data sequentially, TCNs use convolutional layers to capture temporal dependencies allowing for parallel processing and typically achieving faster training times. The main features of TCNs are:
    a. *Causal convolutions* that ensure that predictions at time $t$ are only influenced by inputs from time $t$ and earlier to maintain the temporal order.
    b. *Dilated convolutions* that allow long-range, time-separated dependencies to be captured using a large receptive field without increasing computational load.

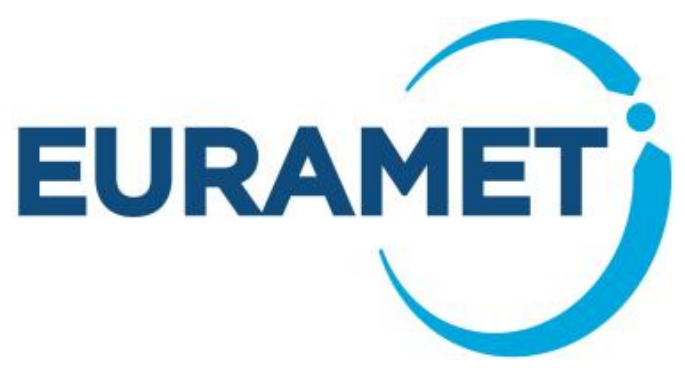

c. *Residual connections* that help to mitigate the ‘vanishing gradient problem’, where gradients used to update model weights during training become very small leading to slow or stalled convergence, and so improve the training of deeper networks.

Generally, it is good practice to ‘clean’ the raw signals before using them to train a DL model and as the basis for predicting the label for a new input to the model. Some examples of data pre-processing steps used in the context of extracting features from raw signals are described in section 2.4.4.

## 2.3 Deep Learning Applied to Image Representations

The idea here is to apply signal processing tools to turn raw PPG signals into two-dimensional image representations, and to use such representations of the signals to train DL models for regression and classification tasks. In particular, established DL models for image processing, such as CNNs, can be used for this purpose. For example, the *ResNet18* model [41] is a popular choice for image classification tasks, but many others are available. Furthermore, *transfer learning*, which uses or adapts a NN pre-trained for a different image processing task, can be used to remove the requirement for large training sets. We describe below two such signal processing tools, the *Continuous Wavelet Transform* (CWT) and the *Symmetric Projection Attractor Reconstruction* (SPAR) method.

### 2.3.1 Continuous Wavelet Transform

A Fourier analysis of a time signal $s(t)$ provides information about the frequency content of the signal in terms of a decomposition of the signal into the component signals $A\sin(2\pi f t + \phi)$ with amplitude parameter $A$ and phase parameter $\phi$ for different frequencies $f$. The information can be displayed as a plot of amplitude $A$ against frequency $f$, which is the so-called amplitude spectrum, revealing which frequency components are dominant. Such an analysis assumes the signal is stationary and provides no information about time-localisation of the different frequency components. Windowing the signal and applying a Fourier analysis within each window, which is the basis of the Short-Term Fourier Transform, is a step towards providing such time-localised information in the form of a plot of amplitude against frequency and window position. However, since the chosen windows are the same for each frequency, the degree of localisation remains the same for different frequencies.

The CWT [42] provides a more flexible time-frequency description of the signal by considering a decomposition into component signals that are dilated and shifted versions of a so-called *mother wavelet function* $\psi(t)$. The representation of the signal in terms of individual wavelet functions defined by

$$\psi_{a,b}(t) = \frac{1}{\sqrt{|a|}}\psi\left(\frac{t-b}{a}\right),$$

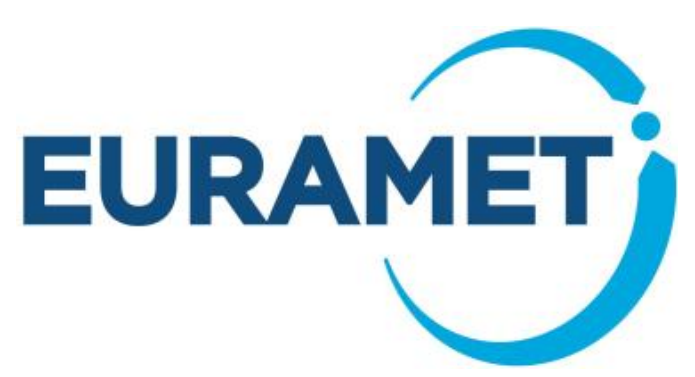

which represent a localised oscillation defined by scale parameter $a$ and translation parameter $b$, allows features in the signal to be identified. Here, low scales are akin to high frequencies in a Fourier analysis and high scales to low frequencies. A classical choice for the mother wavelet function is the 'Mexican hat' function $\psi(t) = (1 - t^2)\exp(-t^2/2)$, but there are many other possible choices (see, for example, [42]) distinguished by their compact support and how fast they decay to zero, and their smoothness.

To apply the CWT requires the selection of the mother wavelet function and the number of scales. The studies conducted in this work considered a generalised Morse wavelet [42] of length 1,024 sample points as the mother wavelet function and 128 logarithmically-spaced scales. Applying the CWT to a PPG signal of length $m$ yields a two-dimensional array of size $n_a \times m$ of wavelet coefficients, where $n_a$ is the number of scales. The CWT coefficients can be displayed as an RGB image, which is stored as a three-dimensional array of size $n_a \times m \times 3$, and that is used as the input to a DL model. Some resizing of the RGB image array may be necessary to align it with the size of the input layer of the model, particularly if transfer learning is applied.

### 2.3.2 Symmetric Projection Attractor Reconstruction Method

The SPAR method [43][44][45] is a novel non-fiducial-point-based method that uses the entire waveform to transform PPG signals into image representations that amplify differences in the morphology and variability of the signals. The SPAR method comprises four basic steps, which are illustrated in Figure 3: (a) given a raw signal to be analysed, the average cycle length is computed; (b) the number of delay coordinates is selected, which will determine the dimension of the attractor [45], and the attractor is generated by plotting these delay coordinates against each other; (c) the multi-dimensional attractor is projected into two-dimensions, with the projection being orthogonal to the unitary vector in the multi-dimensional space containing the attractor; (d) the two-dimensional plot of the attractor is post-processed to generate a density plot, which is the final SPAR image used as input to a DL model.

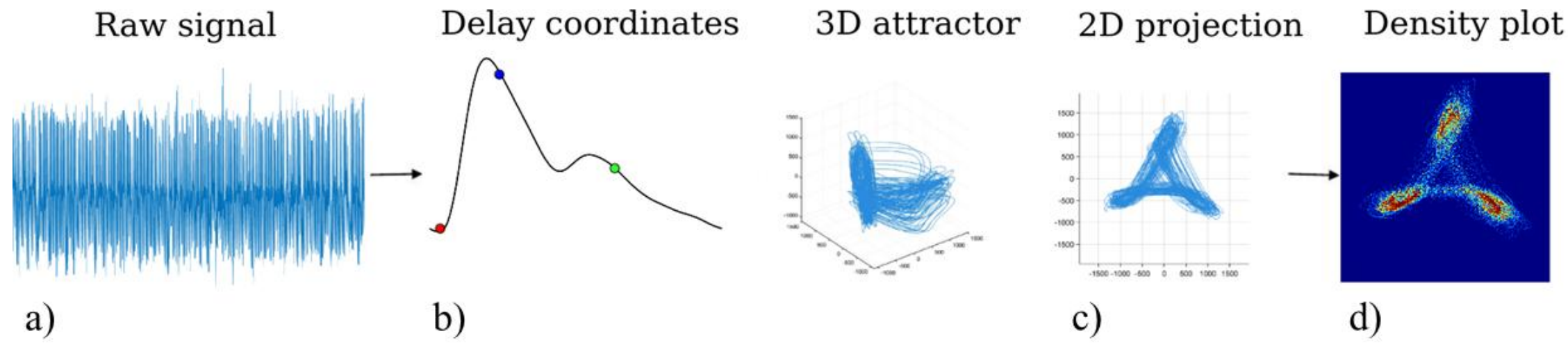

Figure 3: Symmetric Projection Attractor Reconstruction (SPAR) pipeline, from the raw PPG signal to the final density plot, which is the image representation used as input to a DL model. (Adapted from [46].)

The method has been applied in a wide variety of clinical contexts and to many cyclic waveforms, including for the classification of chronological aging as a surrogate of vascular aging from PPG signals. Those studies showed marked visual differences between the SPAR images for healthy human male and female subjects in the '18-39' and 'over 70' age ranges [47] (see Figure 4).

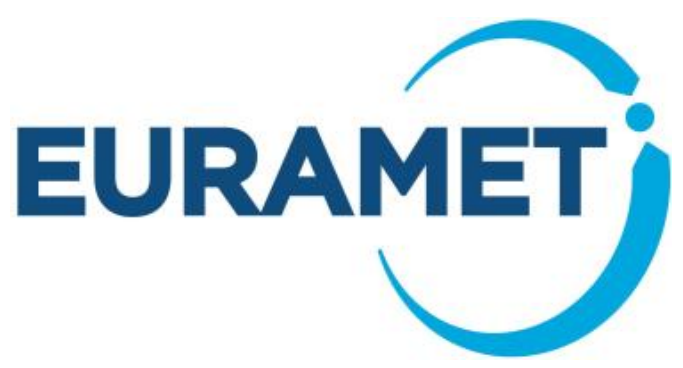

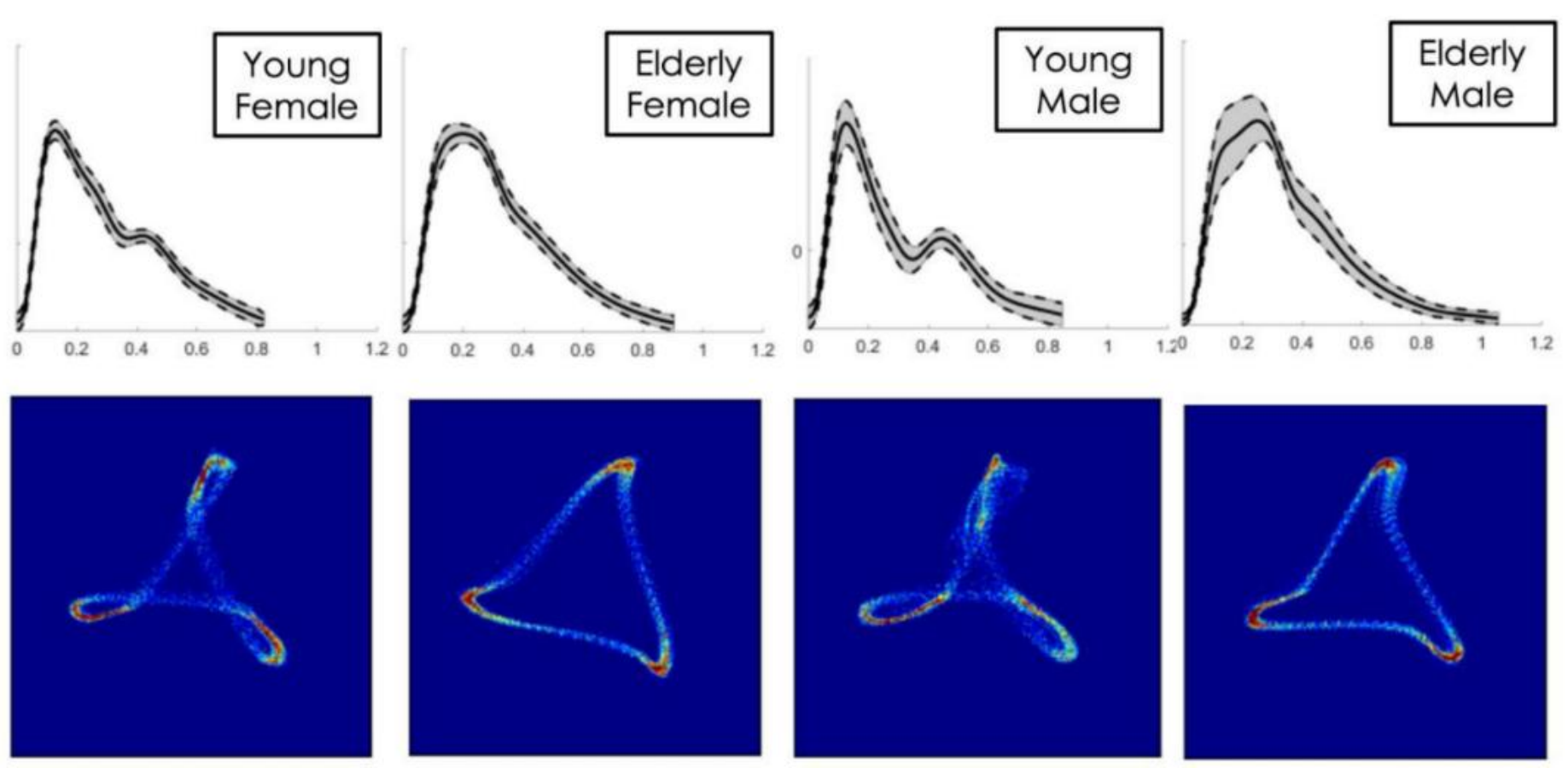

Figure 4: Effect of age and sex on PPG wave morphology for healthy human subjects (where 'young' corresponds to 18-39 years and 'old' refers to over 70 years): ensemble averaged 120 s PPG waveforms (top) and corresponding SPAR images (bottom). (Adapted from [47].)

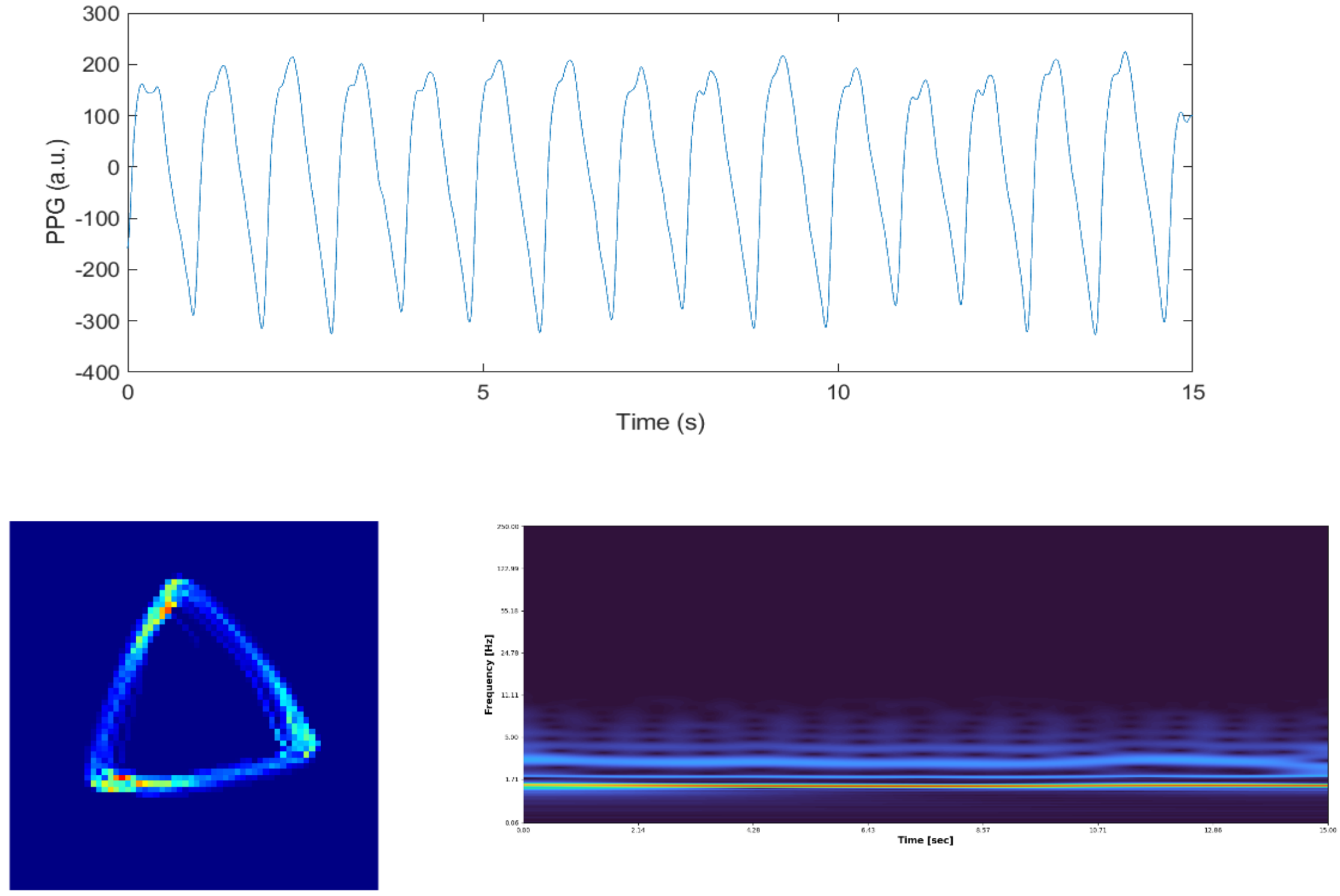

Figure 5: The SPAR attractor (bottom left) and the CWT image (bottom right) derived from the same PPG signal (top).

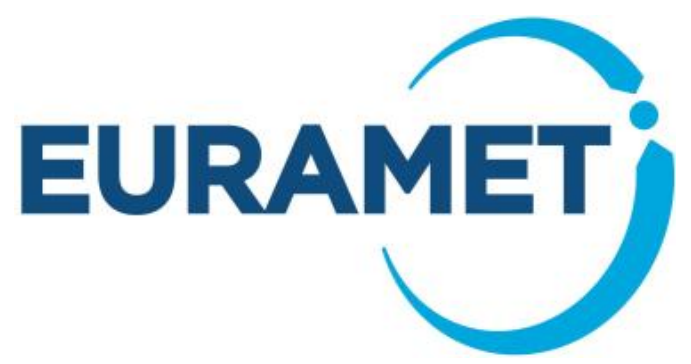

Examples of images obtained from the morphology-based SPAR method and the frequency-based CWT method are illustrated in Figure 5, where each is derived from the same PPG signal (also shown).

## 2.4 Machine Learning Applied to Extracted Features

The idea here is to apply ML to features extracted from PPG signals that describe signal morphology or variability that are likely to encode information relevant to estimating the physiological parameter of interest. Features capturing the morphology of PPG pulses include amplitude-related features, for example, the amplitudes of the systolic and diastolic peaks, and time-related features, for example, pulse, diastole and systole durations. Other possible features include those that capture characteristics of the *peak-to-peak* (PP) intervals extracted from PPG signals, for example, features that measure the randomness, variability, and complexity of sequences of such intervals. The choice of which features to use can be strongly dependent on the particular task of interest.

Using BP estimation (section 2.4.1) and AF detection (section 2.4.2) as illustrative of respectively, a regression task and a classification task, a selection of different features for the development of ML models are described. PPG signals can show a strict periodic behaviour or, in general, a quasi-periodic behaviour, which motivates the use of features derived from Fourier, wavelet or Hilbert-Huang transformations for training ML models. The use of the *Discrete Wavelet Transform* (DWT) and the related *Discrete Wavelet Packet Transform* (DWPT) for this purpose are described in section 2.4.3. Feature-based ML methods often benefit from signal pre-processing as well as the selection of those features that are most informative about the output variable of interest, and these topics are discussed in section 2.4.4. Finally, feature-based ML models are not restricted to deep NNs; other possibilities are considered in section 2.4.5.

### 2.4.1 Estimation of Blood Pressure

For the regression task of BP estimation, different features that capture the morphology of a PPG pulse can be used [48][49]. These features are derived from the PPG signal itself and its derivatives (Figure 6) or as higher order statistics such as skewness and kurtosis (Figure 7).

Those derived from the PPG signal include:

1. Amplitude-related features:
    a. The amplitudes of the first, second and third pulse waveforms [50], $P_1$, $P_2$, $P_3$.
    b. The ratio of the amplitudes of the first and second pulse waveforms, $P_2/P_1$.
    c. The *reflection index*, $P_3/P_1$.
    d. The *augmentation index*, $(P_1 - P_3)/P_1$.

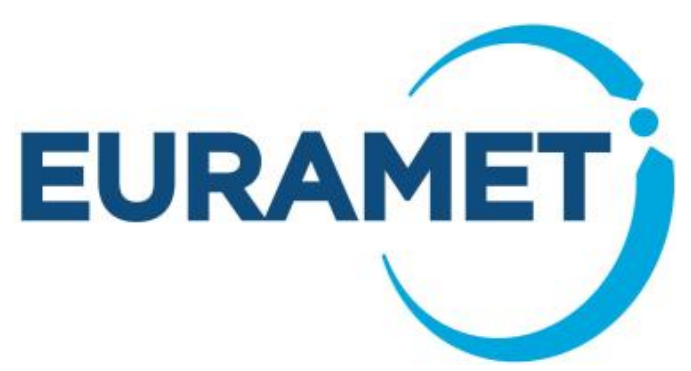

2. Time-related features:
    a. The *pulse duration*, which is the time interval from the pulse onset to the pulse endpoint, $T_p$.
    b. The *diastole duration*, which is the time interval from the dicrotic notch to the pulse endpoint, $T_d$.
    c. The *systole duration*, which is the time interval from the pulse onset to the first systolic peak, $T_1$.
    d. The time interval from the first systolic peak to the diastolic peak, $\Delta t$.
3. Area-related features:
    a. The *systolic area*, which is the area under the signal between the pulse onset and the dicrotic notch, $A_1$.
    b. The *diastolic area*, which is the area under the signal between the dicrotic notch and the pulse endpoint, $A_2$.
    c. The *inflection point area ratio*, which is the ratio of the systolic and diastolic areas, $A_2/A_1$.
    d. The inflection point area ratio $A_2/A_1$ plus the amplitude $d$ of the second derivative of the PPG signal at the second systolic peak.

Those derived from the second derivative of the PPG signal include:

1. Amplitude-related features:
    a. The ratios of the amplitudes of the second derivative of the PPG signal at points along the PPG signal, $b/a$, $c/a$, $d/a$ and $e/a$.
    b. The *cardiovascular aging index*, which is used for the estimation of arterial stiffness and the evaluation of cardiovascular aging, $(b - c - d - e)/a$.
    c. The *interval aging index*, $(b - e)/a$.
    d. The *modified aging index*, $(b - c - d)/a$.
2. Time-related features:
    a. The time interval between the $a$ and $b$ features, $t_{b-a}$.
    b. The time interval between the $b$ and $c$ features, $t_{b-c}$.
    c. The time interval between the $b$ and $d$ features, $t_{b-d}$.
3. Slope-related features:

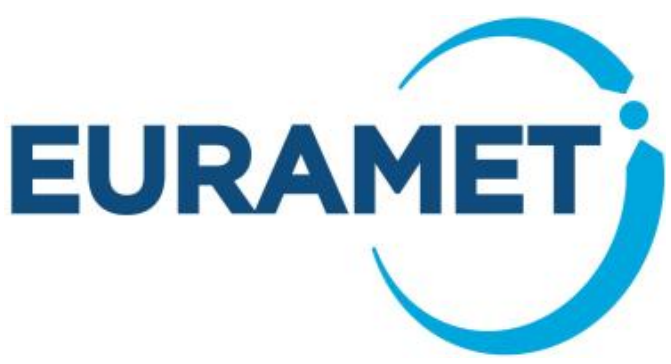

a. The slope of the straight-line passing through the $b$ and $c$ features, $S_{b-c}$.
b. The slope of the straight-line passing through the $b$ and $d$ features, $S_{b-d}$.

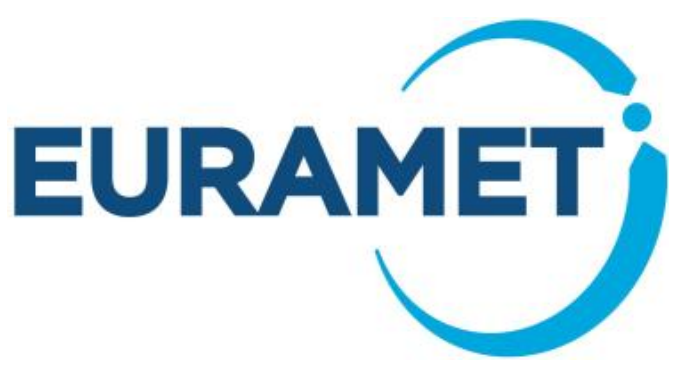

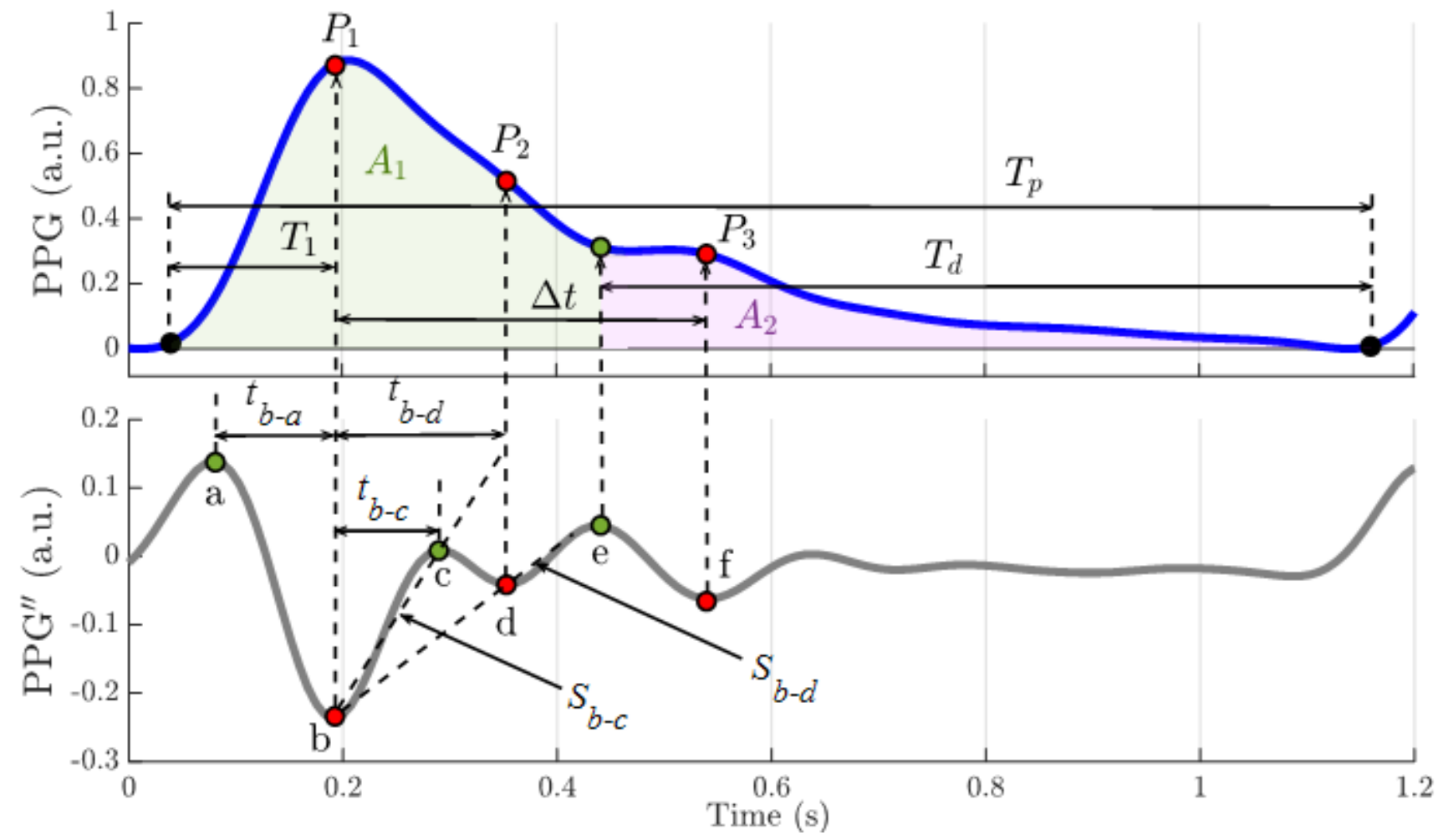

Figure 6: Illustration of PPG signal features based on waveform morphology. The (normalised) PPG signal is shown at the top, and the second derivative of the signal below, with various amplitude-related, time-related and area-related features. (Reproduced from [2].)

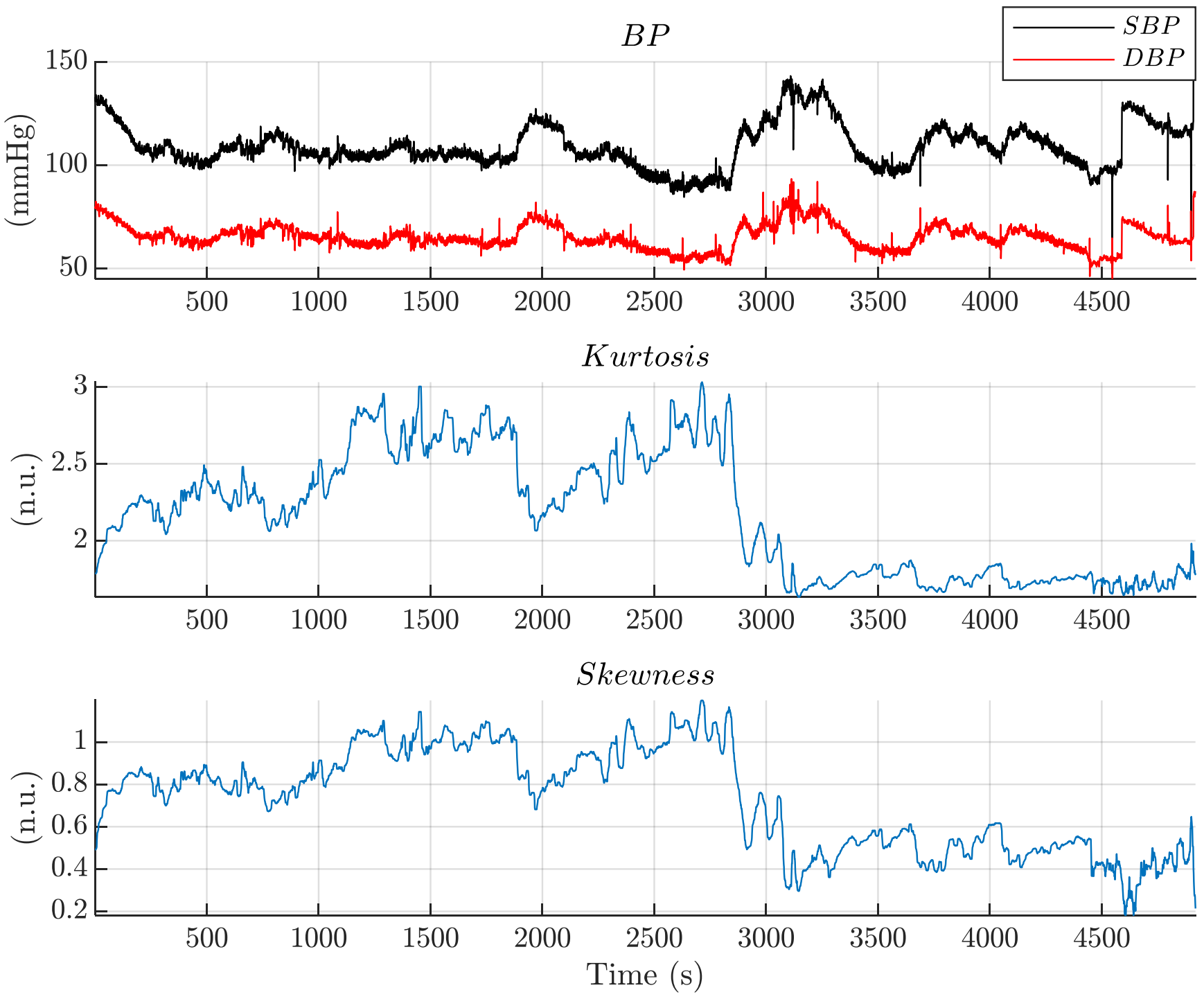

Figure 7: Changes in beat-to-beat systolic BP (SBP) and diastolic BP (DBP) are shown at the top, and corresponding trends in the kurtosis and skewness are shown in the middle and bottom, respectively. (Reproduced from [2].)

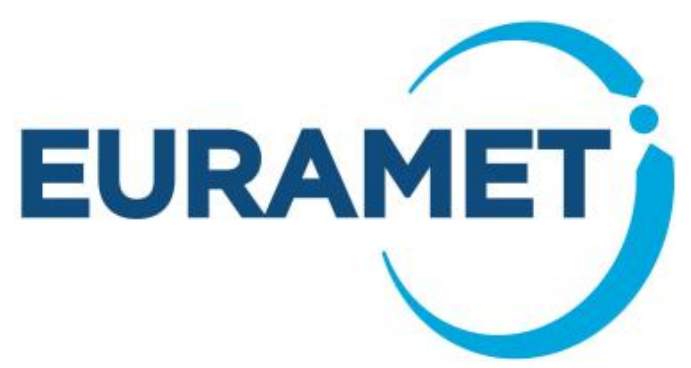

### 2.4.2 Detection of Atrial Fibrillation

To diagnose AF, a clinician typically examines a 12-lead *electrocardiogram* (ECG) for three distinctive features: a highly irregular rhythm, the absence of $P$ waves, and the presence of $f$ waves [51]. However, atrial activity is barely recognisable in PPG signals, and the only clinically interpretable feature is an irregular rhythm. For the classification task of AF detection, the irregularities in PP intervals extracted from PPG signals are described by features that measure randomness, variability, and complexity [52]. Some of those features include:

1. The *turning point ratio*, which is a measure of the randomness of a series of PP intervals. A PP interval is considered as a turning point if the length of the interval is greater than (or less than) the lengths of both intervals that are its nearest neighbours. The turning point ratio is evaluated as the ratio of the number of turning points to the total number of PP intervals within the analysis time. A higher turning point ratio signifies a greater degree of randomness and serves as an indicator of AF.
2. The *coefficient of variation*, which is a measure of variability and is evaluated as the ratio of the standard deviation of the lengths of the PP intervals within the analysis time to the average of those lengths. When AF occurs, the standard deviation of the lengths of the PP intervals increases whereas the mean duration of the PP intervals decreases, yielding a larger coefficient of variation.
3. The *mean successive beat interval difference*, which is a measure of variability and, similarly to the coefficient of variation, relates the dispersion of PP interval lengths to the mean PP interval length.
4. The *root-mean-square of successive differences*, which is a measure of variability that does not account for changes in mean heart rate. It tends to increase during AF.
5. The *Shannon entropy*, which quantifies the 'unpredictability' of the information contained within a signal. It increases as the probability distribution of the values of the signal becomes more uniform and decreases when the distribution becomes more centred around a specific value. In the detection of AF, the Shannon entropy is often considered due to its tendency to be substantially higher during AF compared to normal sinus rhythm.
6. The *sample entropy*, which assesses the 'self-similarity' of sequences of samples within a signal and can be interpreted as a measure of signal complexity. It is expressed as the negative natural logarithm of the conditional probability that two sequences of $m$ samples that are similar within a tolerance $r$ remain similar for the next sample, with self-matches excluded.
7. A *Poincare plot*, which is a scatter plot of consecutive pairs of PP interval lengths. In the case of AF, the Poincare plot displays a substantially higher level of dispersion compared to plots corresponding to a normal sinus rhythm or to atrial or ventricular premature beats containing rhythms.

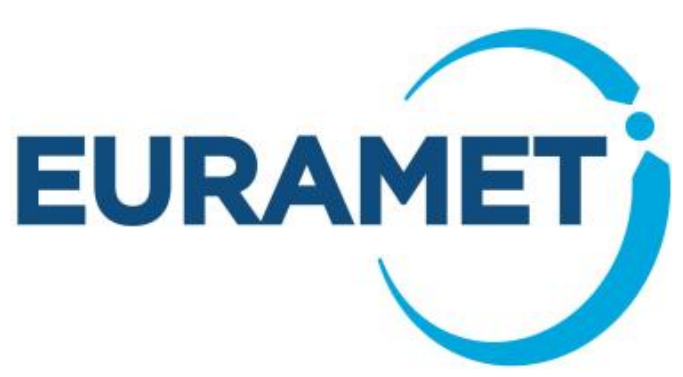

It is good practice to undertake an assessment of the quality of the extracted PP intervals to identify 'good' and 'bad' intervals before evaluating features such as those listed above. Removing bad intervals can help the ability of those features to discriminate between PPG signals obtained when AF is present or not, which in turn can lead to improved predictive performance of ML models. An algorithm for undertaking such quality assessment is described in [53].

### 2.4.3 Discrete Wavelet Transform

The CWT was introduced in section 2.3.1 as providing a time-frequency description of a signal by considering a decomposition into wavelet functions $\psi_{a,b}(t)$ that represent a localised oscillation defined by scale parameter $a$ and translation parameter $b$. In the CWT, the scale and translation parameters are continuous, whereas in a DWT they take the discrete values $a = 2^j$ and $b = ka$ for integers $j$ and $k$, and the wavelet function is written as

$$\psi_{j,k}(t) = 2^{-j/2}\psi\left(2^{-j}t - k\right).$$

The DWT gives a sparser representation of the signal, making it suitable for signal compression and other applications where computational cost is a factor.

From a signal processing perspective, the DWT can be considered as a decomposition of the signal using high and low pass filters into spaces of 'approximations' and 'details' at successive levels as illustrated in Figure 8 [54]. The high pass filters extract localised detail in the signal whereas the low pass filters smooth the signal. The properties of the decomposition are determined by the filters, which can in turn be related to the wavelet functions $\psi_{j,k}(t)$ that form a basis for the 'detail' spaces.

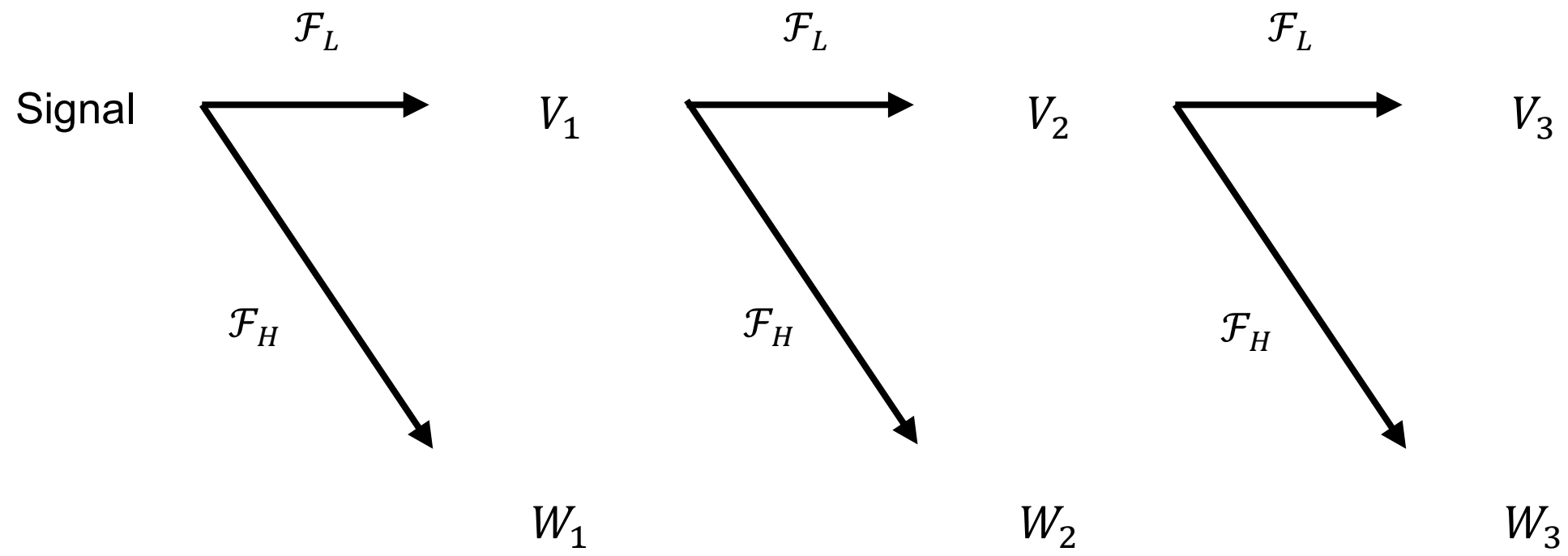

Figure 8: Schematic of a three-level decomposition of a signal into approximation spaces $V_j$ and detail spaces $W_j$ by a combination of low pass filter $\mathcal{F}_L$ and high pass filter $\mathcal{F}_H$. (Adapted from [54].)

In the DWT, the signal is decomposed into approximation and detail spaces, and then only the approximation spaces are further decomposed at successive levels. In contrast, in the DWPT [42], both the approximation and detail spaces are decomposed at each level. The advantage of this decomposition is that there is more freedom in deciding which wavelet basis functions to use to represent the signal, and there is increased flexibility in adapting the basis to the frequency content of the signal. The decomposition provides a finer level of signal analysis compared to the standard DWT, capturing more detailed information across different frequency bands.

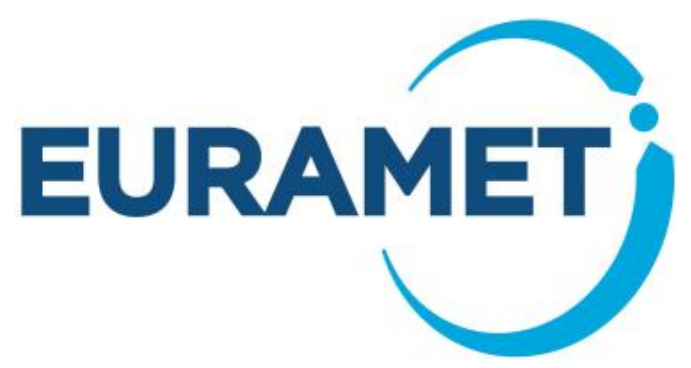

For the estimation of BP, the studies in the QUMPHY project [2] used a Daubechies wavelet of order 3, which offers a balance between time and frequency localisation, making it particularly suitable for analysing non-stationary signals. The DWPT coefficients are 'vectorised' and then can directly be used without additional processing as the input to a ML model.

### 2.4.4 Signal Pre-Processing, Feature Selection and Ranking

Raw PPG signals often show reduced quality due to motion artefacts, baseline drift, and hypoperfusion. Motion artefacts result from body movement or improper sensor attachment, introducing signal fluctuations that degrade signal quality [55]. Baseline drift, caused by respiration and body movements, shifts the PPG signal and may obscure the true pulsatile component [56]. Hypoperfusion, characterised by reduced peripheral blood flow due to vasoconstriction, weakens PPG signals, affecting the accuracy and reliability of physiological measurements [56].

While NN-based DL models can learn to deal with such effects, feature-based ML methods often benefit from signal pre-processing implemented using low pass, high pass and adaptive filters. Typically, the lower bound of the passband (at about 0.5 Hz) removes components below 0.1 Hz and respiratory influences between 0.1 Hz and 0.5 Hz, and the upper bound (at between 5 Hz and 10 Hz) retains the primary PPG signal frequency components, that are often captured using Butterworth, Chebyshev II, or finite impulse response filters [56].

For example [2], for the estimation of BP, the PPG segments from the *VitalDB* dataset (section 5.1) were pre-processed using a fourth order Butterworth infinite impulse response zero-phase band-pass filter with frequency passband of 0.4 – 7 Hz. Furthermore, for the detection of AF, the PPG segments from the *DeepBeat* dataset (section 5.2) were pre-processed using low pass, high pass, and adaptive filters. High-frequency noise and artefacts were removed using a low pass infinite impulse response filter with a cut-off frequency of 6 Hz, while baseline wander was eliminated using a high pass infinite impulse response filter with a cut-off frequency of 0.5 Hz and a fifth-order normalised least mean squares adaptive filter with a reference input of unity.

Feature selection (or dimensionality reduction) is an essential step in developing robust ML models. PPG signals provide a large number of potential features, such as time-domain fiducials, derivative-waveform metrics, and amplitude ratios, many of which can be redundant and irrelevant to the task of interest, or noisy. Dimensionality reduction studies demonstrate that while adding features can initially improve classifier performance, excessive dimensionality often degrades performance due to the 'curse of dimensionality' [57]. Feature selection methods are categorised into filter, wrapper, and embedded approaches [57]. Filter methods assess feature relevance using statistical criteria independent of the learning algorithm, such as the *Minimum Redundancy Maximum Relevance*

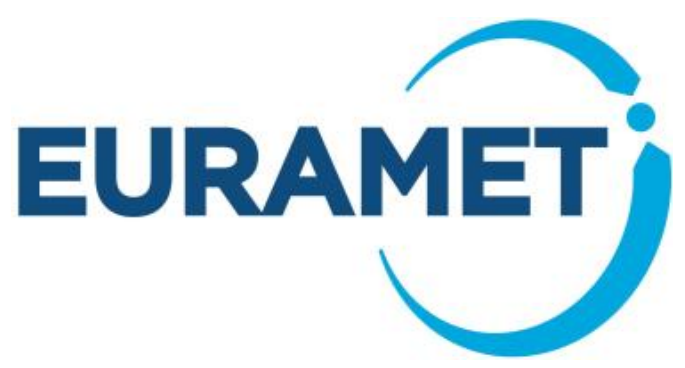

(MRMR) algorithm [58]. Wrapper methods evaluate subsets of features based on the predictive performance of a model, for example, the *Predictive Permutation Feature Selection* (PPFS) algorithm [59] identifies the most influential features by iteratively measuring the impact of feature permutation on model accuracy. Embedded methods integrate feature selection directly into model training, as exemplified by the *Least Absolute Shrinkage and Selection Operator* (LASSO) [60]. In addition, dimensionality reduction techniques such as *Principal Component Analysis* (PCA) [61], *Independent Component Analysis* (ICA) [62], and autoencoders [63] can extract compact, informative representations of high-dimensional data, improving model stability and interpretability. For example, in PPG-based blood pressure estimation studies [64][65], feature selection using the MRMR algorithm significantly reduced the number of predictors without compromising model performance.

Feature ranking provides insight into which physiological characteristics most strongly influence model predictions. Model-agnostic interpretability methods such as *SHapley Additive exPlanations* (SHAP) [66] and *Local Interpretable Model-agnostic Explanations* (LIME) [67] quantify the contribution of individual features to model outputs. While SHAP values provide a unified measure of global and local feature importance, LIME builds simplified local surrogate models to approximate the behaviour of complex models around specific predictions. For instance, SHAP-based feature ranking applied to PPG-derived features to estimate changes in blood pressure [49] revealed that features associated with reflected wave interference, indicative of arterial stiffness, were among the most influential predictors. Such interpretability techniques enhance transparency and support the physiological interpretation of learned models, which is critical for healthcare applications where trustworthiness and explainability are essential.

### 2.4.5 Machine Learning Models

The feature-based approach often involves significant dimension reduction and can be thought of as reducing the raw data to a simpler representation. This reduction means that it may not be necessary to use a deep NN, and it may make sense to consider other types of ML model. One possibility is a so-called *shallow* NN, which means a NN with a small number of layers so as not to be considered ‘deep’. Other possibilities include *support vector machines* (SVMs), and tree-based models such as random forests and gradient-boosted trees. These models all have the advantage of reducing computational time for training the model compared to deep NNs. Another option is *Gaussian Process* (GP) models, which come as standard with a principled uncertainty quantification that does not rely upon approximations, though at the cost of increased training time compared to other options. We refer the interested reader to [68] for an overview of different ML regression and classification models.

For this study, different types of ML model were used for each of the types of input features described in sections 2.4.1, 2.4.2 and 2.4.3. For the morphology features for BP estimation in section 2.4.1, a GP model was used. For the morphology features for AF classification in section 2.4.2, a particular fully-connected shallow NN known as a *Multi-Layer Perceptron* (MLP) was used. Meanwhile, for the DWT features in section 2.4.3, a deep NN was used.

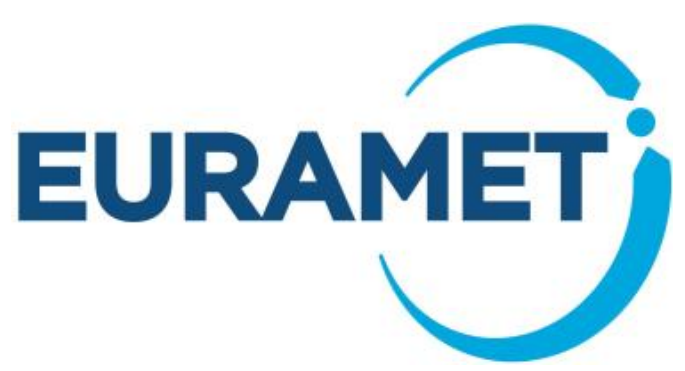

# 3 Uncertainty Quantification

## 3.1 Background

In regression tasks, the output variable corresponding to a new input value $x$ is a continuous, quantitative variable. Using the nomenclature of metrology (the science of measurement), the output variable is the *measurand* (the quantity intended to be measured) [69] and, using good metrological practice, it is modelled as a random variable, the probability distribution for which provides the most complete description of knowledge about the variable [70]. The *probability density function* (PDF) for the probability distribution can be used to summarise that knowledge, for example, in terms of an *expectation*, which provides an estimate $y$ of the quantity, a *standard deviation*, which provides the standard uncertainty $u(y)$ for the estimate, and a *coverage interval* that contains values of the variable with a stated probability. Both the standard deviation and coverage interval constitute valid expressions of (measurement) *uncertainty*, which is defined (as of 2026) in [69] as a “non-negative parameter characterising the dispersion of the quantity values being attributed to a measurand, based on the information used”, and in [70] as a “parameter, associated with the result of a measurement, that characterises the dispersion of the values that could reasonably be attributed to the measurand”.

In classification tasks, the output variable corresponding to a new input value $x$ is a discrete, categorical variable. Ways to treat such variables within a metrological context are currently (as of 2026) being developed and debated (see, for example, [71]). Modelled also as a random variable, a natural way to capture knowledge about the output variable is in terms of a *probability mass function* (PMF) comprising the probabilities $p_1, \dots, p_K$ of each class. However, the ordering of the classes is usually arbitrary and there is not such a rich notion of ‘distance’ between the possible values $\{1, \dots, K\}$ of the output variable, so different ways of summarising that knowledge are needed. For example, the *mode* of the distribution (the class having largest probability) provides an estimate $y$ of the variable that, in the case of no ‘ties’, returns a class label as the estimate that is the same whatever is the ordering of the labels. Furthermore, the *normalised entropy*, defined by

$$H(y) = -\sum_{i=1}^{K} p_i \log_K p_i \,,$$

provides a way to express uncertainty of $y$ in terms of a scale defined by two extreme cases: at one extreme, when a single class has probability one and all other classes probability zero, the normalised entropy is zero, implying no uncertainty about the class label; at the other extreme, when all classes have equal probability $1/K$, the normalised entropy is one, implying maximum uncertainty about the class label. This representation of uncertainty has been used in the recent international standard [72].

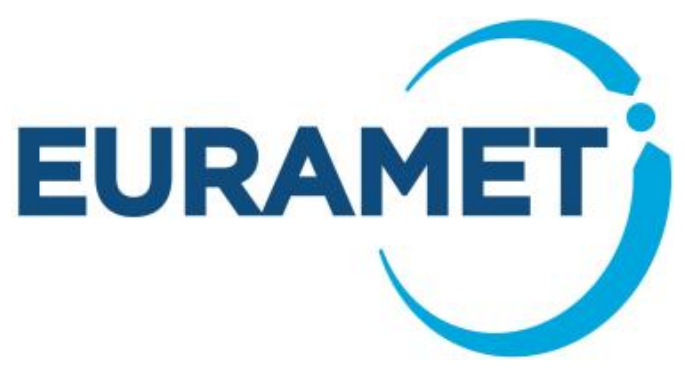

There are various sources of uncertainty which contribute to the overall uncertainty of a predicted value, including those associated with the specification of the model, the training data and the data for which a prediction is required. Good metrology practice requires the identification of all sources of uncertainty that can contribute to the uncertainty of a predicted value, and the quantification of their contributions. Furthermore, understanding the different sources can help to ensure the reliability of the overall uncertainty by informing decisions about the choice of model architecture and data collection, and also help with the interpretation of the uncertainty by the decision-maker. Two types of uncertainty in ML and DL are often distinguished [73][74]: *epistemic uncertainty*, also called 'model uncertainty', and *aleatoric uncertainty*, also called 'data uncertainty'.

Epistemic uncertainty refers to ambiguity in the model specification, which can be of two types: uncertainty concerning the choice of type of model and uncertainty concerning the parameters of a given type of model. Quantifying uncertainty concerning the choice of type of model is challenging, and it is common practice to fix the choice of model and focus on parameter uncertainty, reflecting that the training of the model is imperfect because the training data is necessarily limited in terms of its quality and/or its volume. Parameter uncertainty may be reduced by training the model on more and/or higher quality data so that the model is better equipped to detect task-specific features.

Aleatoric uncertainty refers to uncertainty that is intrinsic to the task and, in contrast to epistemic uncertainty, is irreducible. *Intrinsic uncertainty* refers to inherent ambiguity in the value of the predicted quantity that no model is able capture, however small is the model uncertainty. Such ambiguity may occur either because the input value fed into the model is insufficient to predict the output variable or because of the presence of random effects which cannot be explained.

In this section we describe and contrast different methods for UQ. Some of the methods, namely Maximum Likelihood Estimation, Maximum A Posteriori estimation, Deep Ensembles and Monte Carlo Dropout, are *model-dependent* (section 3.2). Here, they are considered with NN models and can be applied to a feed-forward NN with an arbitrary architecture. Other methods, namely calibration methods and conformal prediction, are *model-independent* (section 3.3) and can be used to generate or improve the uncertainty estimates for any regression or classification model. Calibration methods require some initial uncertainty estimates, and these are also described.

## 3.2 Model-Dependent Methods

### 3.2.1 Regression Tasks

A popular approach is to assume a parameterised probability distribution for the output variable, for example,

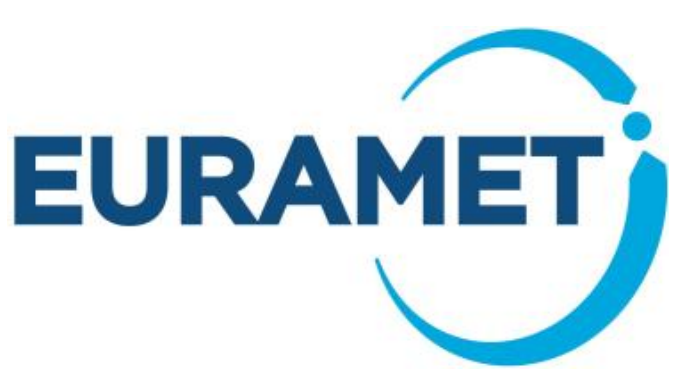

$$y \in \mathrm{N}\big(\mu(x,\theta), \sigma^2(x,\theta)\big),$$

which says that the value of the output variable is drawn from a Gaussian distribution with mean $\mu(x,\theta)$ and variance $\sigma^2(x,\theta)$, each of which depends on input value $x$ and parameters $\theta$ of a ML model. The probability of observing training labels $\{y_1, \dots, y_N\}$ given a ML model defined by parameters $\theta$ (the so-called 'likelihood') is

$$p(y_1, \dots, y_N|\theta) = \prod_{i=1}^{N} \frac{1}{\sigma(x_i,\theta)\sqrt{2\pi}} \exp\left\{-\frac{\big(y_i - \mu(x_i,\theta)\big)^2}{2\sigma^2(x_i,\theta)}\right\},$$

where it is assumed that only the labels $\{y_1, \dots, y_N\}$ are subject to uncertainty and those labels are obtained from measurements made independently.

Consider the case of a ML model $f(x,\theta)$ in the form of a NN designed to have an input layer that accepts $x$, an output layer with two nodes that provide values for $\mu$ and $\log \sigma^2$, and hidden layers in between defined by weights and biases, collectively denoted by $\theta$. The NN is trained by minimising the *negative-log-likelihood* (NLL) loss function, which takes the form

$$-\log p(y_1, \dots, y_N|\theta) = \sum_{i=1}^{N} \left\{\frac{\log \sigma^2(x_i,\theta)}{2} + \frac{\big(y_i - \mu(x_i,\theta)\big)^2}{2\sigma^2(x_i,\theta)}\right\} + N \log \sqrt{2\pi}$$

when the distribution for the output variable is Gaussian (and it is then referred to as the Gaussian NLL), giving the *Maximum Likelihood Estimate* (MLE) $\hat{\theta}$ of $\theta$ [74][75]. Note that specific knowledge about data uncertainty is not needed to train the NN but instead data uncertainty is estimated statistically [74]. Here, usual good practice in the training of the NN is followed, including the use of validation and test sets to evaluate the performance of the network during and after the training phase in terms of established metrics. Then, the probability distribution for the output variable corresponding to an arbitrary input value $x$ is $\mathrm{N}(\mu(x,\hat{\theta}), \sigma^2(x,\hat{\theta}))$, which provides a complete description of the prediction and specifically both the estimate $\mu(x,\hat{\theta})$ and its standard uncertainty $\sigma(x,\hat{\theta})$.

In a Bayesian treatment, a prior distribution is specified for the model parameters that captures specialist, domain knowledge such as physical constraints that the parameters are known to satisfy. For example, if a prior distribution of the form $\mathrm{N}(0,\eta^2)$ is assumed for each of the $n$ model parameters $\theta_j$, the *Maximum A Posteriori* (MAP) estimate $\hat{\theta}$ for $\theta$ is obtained by minimising the modified loss function

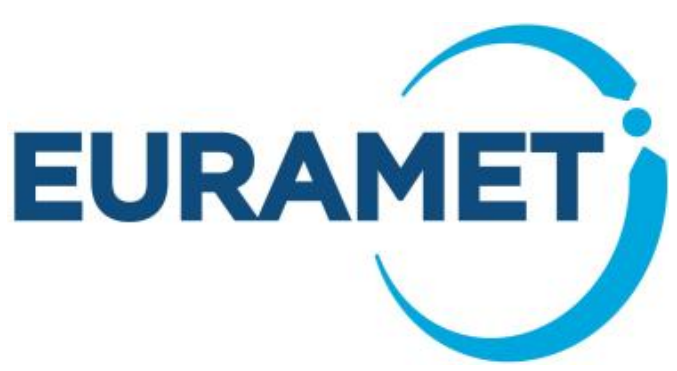

$$-\log\{p(y_1,\dots,y_N|\theta)p(\theta)\} = \sum_{i=1}^{N}\left\{\frac{\log\sigma^2(x_i,\theta)}{2} + \frac{\big(y_i-\mu(x_i,\theta)\big)^2}{2\sigma^2(x_i,\theta)}\right\} + N\log\sqrt{2\pi} + \sum_{j=1}^{n}\frac{\theta_j^2}{2\eta^2}$$

that incorporates the knowledge encapsulated by those prior distributions [3]. As $\eta \to \infty$ and the prior distributions become less informative about the parameters, so the MAP estimate tends towards the MLE, but maintains its 'Bayesian' interpretation of representing an updated belief about the model parameters to incorporate the evidence of observed data. Variants of the MLE and MAP approaches can be formulated to treat different distributions to describe uncertainty in the labels as well as to include uncertainty in the input values. However, in all these cases, the probability distribution defined by the outputs $\mu(x,\hat{\theta})$ and $\sigma^2(x,\hat{\theta})$ from the NN only captures aleatoric uncertainty because it uses a point estimate of the model parameters and models intrinsic uncertainty in the class label.

*Deep Ensembles* (DEs) [75] provide one way to include epistemic uncertainty arising from a lack of knowledge about the model $f$. The idea is to train a model starting with $M$ different random initialisations of the weights to obtain an ensemble of trained models $f_m(x,\theta_m), m = 1,\dots,M$, and then combine the predictions from those individual models. For example, suppose each model is a NN with, as above, the output of the $m^{\text{th}}$ model comprising the mean $\mu_m(x,\hat{\theta}_m)$ and variance $\sigma_m^2(x,\hat{\theta}_m)$ of the predicted distribution for input value $x$. The probability distribution for the prediction corresponding to an input value $x$ is taken to be a uniformly-weighted mixture of the distributions from the individual models, i.e.,

$$y \in \frac{1}{M}\sum_{m=1}^{M}\mathrm{N}\left(\mu_m(x,\hat{\theta}_m),\sigma_m^2(x,\hat{\theta}_m)\right).$$

This mixture distribution is also Gaussian $\mathrm{N}(\mu_*(x),\sigma_*^2(x))$, and assuming independence between the distributions from the individual models, its mean and variance are

$$\mu_*(x) = \frac{1}{M}\sum_{m=1}^{M}\mu_m(x,\hat{\theta}_m), \qquad \sigma_*^2(x) = \frac{1}{M}\sum_{m=1}^{M}\left(\sigma_m^2(x,\hat{\theta}_m) + \mu_m^2(x,\hat{\theta}_m)\right) - \mu_*^2(x),$$

and the distribution again provides both an estimate $\mu_*(x)$ of the output variable and its total uncertainty $\sigma_*(x)$. It is important that each model is trained using a *proper scoring rule* [76], which is a metric whose expected value is optimal when the predicted distribution matches the true distribution. Proper scoring rules can be decomposed into two parts [77]: one which measures *calibration* (the extent to which uncertainties align with observed dispersion) and one which measures *sharpness* (related to the magnitude of the uncertainty). The Gaussian NLL loss function is an example of a proper scoring rule. Larger values for the number $M$ of models in the ensemble may deliver more reliable values for the uncertainty but will incur greater computational costs associated with training more models. A pragmatic approach is to evaluate uncertainty reliability for various values of $M$ up to the value that is feasible practically and then report results for the number that produced optimal reliability.

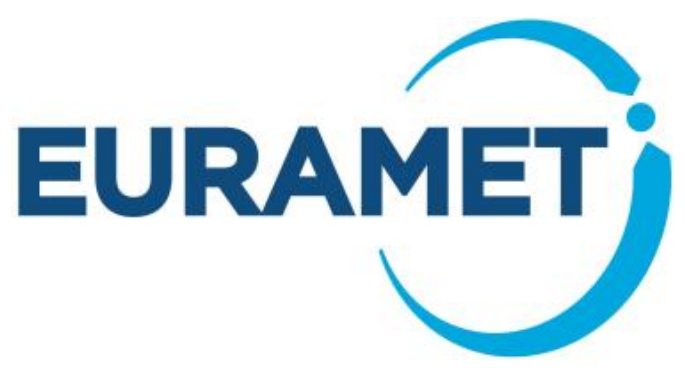

It is noted that $\sigma_*^2$ can be decomposed as

$$\sigma_*^2 = \frac{1}{M}\sum_{m=1}^{M}\left(\mu_m(x,\hat{\theta}_m) - \mu_*(x)\right)^2 + \frac{1}{M}\sum_{m=1}^{M}\sigma_m^2(x,\hat{\theta}_m),$$

in which the first term, which captures variability in the estimates provided by the individual models, represents epistemic uncertainty, and the second term, which depends on the uncertainties estimated by the individual models, represents aleatoric uncertainty.

*Monte Carlo Dropout* (MCD) [78] provides another way to include epistemic uncertainty but avoids the need to train different models in an ensemble. The operation of dropout involves randomly selecting nodes in the hidden layers of the NN and deactivating those nodes, i.e., removing the nodes and their connections to other nodes. Given a NN trained using MCD, the idea is to apply dropout $M$ times to obtain for $m = 1, \ldots, M,$ values for the mean $\mu_m(x)$ and variance $\sigma_m^2(x)$ of the predicted distribution for input value $x$. As is done above for the case of DE, those results are then aggregated to obtain a single estimate $\mu_*(x)$ of the output quantity with an uncertainty $\sigma_*(x)$ that can be decomposed into components representing epistemic and aleatoric uncertainty.

One disadvantage of MCD is that the diversity of the models produced can be low as each model is derived from the same 'parent model'. It is hypothesised that this corresponds to sampling from a single mode of the underlying posterior distribution, which means that the approach does not capture the full variability of the posterior. In contrast, DE consider a 'parent architecture' and can generate trained models that may correspond to different modes of the underlying posterior distribution so they produce a greater diversity of models. However, variability local to those modes is not captured. Like the choice of ensemble models in DE, results from MCD can be sensitive to the choice of hyperparameters, such as the dropout rate (or dropout probability) that determines the numbers of nodes that are deactivated, as well as weight regularisation.

### 3.2.2 Classification Tasks

In the case of classification, a natural distribution for the class label is the discrete categorical distribution defined by class probabilities $p_k, k = 1, \ldots, K,$ which are constrained to sum to unity. For a binary classification problem, the distribution reduces to the Bernoulli distribution defined by two class probabilities $p_1$ and $p_2 = 1 - p_1$. Then, the likelihood for the training set is

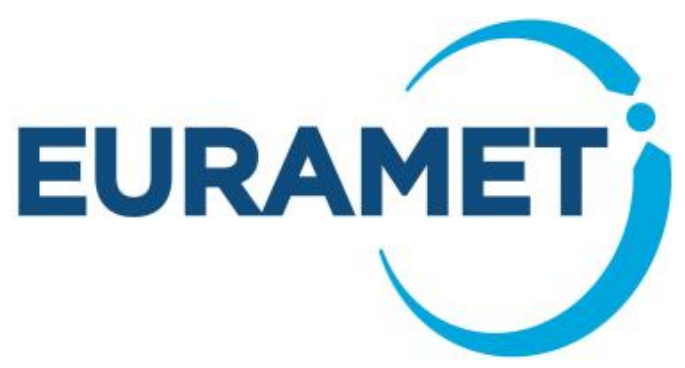

$$p(y_1,\dots,y_N|\theta) = \prod_{i=1}^{N}\prod_{k=1}^{K} p_k(x_i,\theta)^{\mathbb{I}(y_i=k)},$$

where $\mathbb{I}(\cdot)$ is the indicator function taking the value of one when its argument is true and zero otherwise, which makes explicit the dependence of the class probabilities on the input value $x$ and the parameters $\theta$ of a ML model.

Consider the case of a ML model $f(x,\theta)$ in the form of a NN designed to have an input layer that accepts $x$, an output layer with $K$ nodes that provides values for $p_1,\dots,p_K$, with hidden layers in between defined by weights and biases, collectively denoted by $\theta$. Note that, typically, the class probabilities are obtained by applying the *softmax* activation function to a set of raw scores (known as *logits*), denoted here by $v_k$, which are the output from the penultimate layer of the NN. The *softmax* function is defined by

$$p_k = \frac{\exp v_k}{\sum_{k=1}^{K}\exp v_k}, \qquad k = 1,\dots,K,$$

which ensures that each $p_k \in [0,1]$ and they sum to one. The MLE and MAP estimates are obtained by minimising, respectively, the NLL loss function [79]

$$-\log p(y_1,\dots,y_N|\theta) = -\sum_{i=1}^{N}\sum_{k=1}^{K}\mathbb{I}(y_i = k)\log p_k(x_i,\theta)\,,$$

and the modified loss function [3]

$$-\log\{p(y_1,\dots,y_N|\theta)p(\theta)\} = -\sum_{i=1}^{N}\sum_{k=1}^{K}\mathbb{I}(y_i = k)\log p_k(x_i,\theta) + \sum_{j=1}^{n}\frac{\theta_j^2}{2\eta^2},$$

with, in the latter, prior distributions for the $n$ model parameters $\theta_j$ expressed as independent Gaussian distributions. Then, the probability distribution for the output variable corresponding to an arbitrary input value $x$ is the categorical distribution defined by the class probabilities $p_1(x,\hat{\theta}),\dots,p_K(x,\hat{\theta})$, which provides a complete description of the prediction. As discussed in section 3.1, a representation of the measurement result can be provided in terms of the mode and normalised entropy evaluated in terms of the class probabilities for that distribution.

Another approach to capturing aleatoric uncertainty is described in [74], which allows for the incorporation of an explicit model for the aleatoric uncertainty of the logits. Similarly to the case of regression described in section 3.2.1, the architecture of the NN is modified, but here in such a way that at the penultimate layer of the NN there are two nodes corresponding to each logit $v_k$, one providing a mean $\mu_k(x,\theta)$ and the other a variance $\sigma_k^2(x,\theta)$ of an assumed Gaussian distribution for the logit. These distributions together with the *softmax* function that transforms the logits to class probabilities imply distributions for the class probabilities that can be approximated numerically. Specifically, a random sample of size $T$ of values for the $K$ logits is evaluated from

$$v_{kj}(x,\theta)=\mu_k(x,\theta)+\sigma_k(x,\theta)z_{kj},\qquad k=1,\dots,K,\qquad j=1,\dots,T,$$

where $z_{kj}\in \mathrm{N}(0,1)$. Then, for each sample $j$, the *softmax* function is applied to the logit values $\{v_{kj}(x,\theta): k=1,\dots,K\}$, to obtain corresponding class probabilities $\{p_{kj}(x,\theta): k=1,\dots,K\}$. Finally, mean class probabilities are evaluated from

$$\bar{p}_k(x,\theta)=\frac{1}{T}\sum_{j=1}^{T}p_{kj}(x,\theta),\qquad k=1,\dots,K.$$

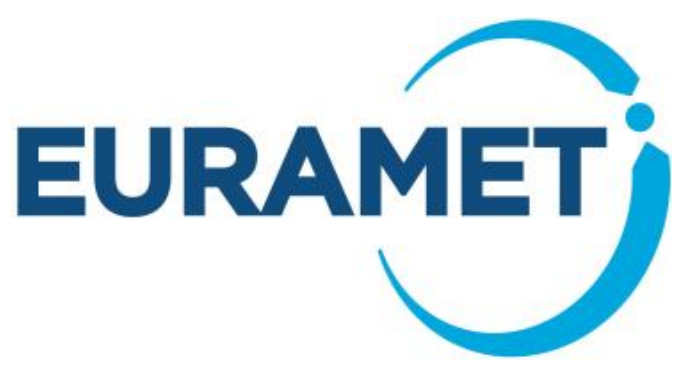

For the training of the NN, these mean class probabilities are used in the MLE or MAP loss functions defined above to obtain estimates $\hat{\theta}$ of the ML model parameters $\theta$.

As for the case of regression described in section 3.2.1, whichever approach is used for calculating the class probabilities only aleatoric uncertainty is captured, and DE and MCD can be used to include epistemic uncertainty. For DE, an ensemble of such NNs is trained using the same parent architecture and different random initialisations of the weights. For MCD, a single parent model is trained using MCD.

For the prediction of the class label for a new input value $x$, class probabilities $p_{km}(x,\hat{\theta}_m)$ (or mean class probabilities $\bar{p}_{km}(x,\hat{\theta}_m)$ in the alternative approach) are obtained for each class $k$ and each ML model $m=1,\dots,M$. For DE, the ML models are the separately trained NNs. For MCD, they are the models obtained by applying dropout to the single trained NN. The ensemble probability distribution for the prediction is taken to be a uniformly-weighted mixture of the distributions from the individual models, which corresponds to the categorical distribution with probabilities defined by the averages of the (mean) class probabilities for the individual models:

$$p_{k*}(x)=\frac{1}{M}\sum_{m=1}^{M}p_{km}(x,\hat{\theta}_m),\qquad k=1,\dots,K.$$

The categorical distribution defined in this way provides complete information about the prediction, which as before can be summarised by its mode $y$ and normalised entropy $H(y)$. Here, $H(y)$ captures epistemic uncertainty through the ‘ensembling’ process and aleatoric uncertainty through the use of the MLE or MAP loss functions to train the individual NNs. Intuitively, a single estimate of the aleatoric uncertainty is provided by averaging the normalised entropies provided by the individual models, which permits an approximate disentangling of the aleatoric uncertainty from the total uncertainty. The entropy of the average probability across all models, however, cannot be considered as epistemic uncertainty alone, because the class probabilities themselves represent aleatoric uncertainty. That entropy could be considered as an expression of total uncertainty. Note however, that popular methods generate aleatoric and epistemic uncertainties that are significantly correlated [80], which is not ideal or realistic.

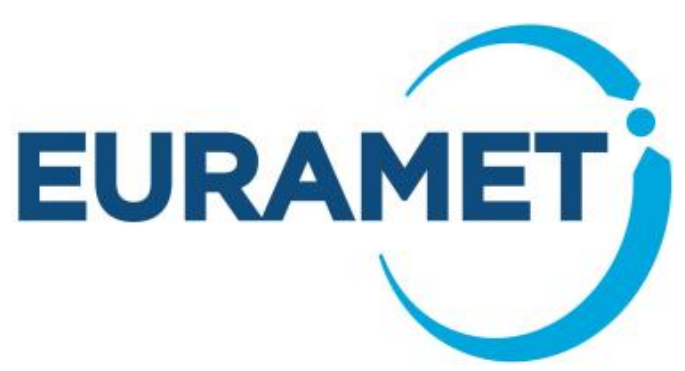

## 3.3 Model-Independent Methods

### 3.3.1 Calibration Methods

The methods presented in this section are described as 'post-hoc' methods for UQ in the sense that they adjust and improve the values of uncertainty provided by a pre-trained ML model. Calibration aims to capture the extent to which calculated uncertainties are aligned with the observed prediction errors. Along with other post-hoc techniques, the methods have the advantage of not requiring the training and validation of a new ML model, but they require a calibration set that is separate from the training and validation sets and is used to make the adjustment (see section 2.1). There are many different calibration methods, including 'binning-based' approaches, such as histogram binning, isotonic regression and Bayesian binning into quantiles, and 'scaling-based' approaches, such as temperature scaling, logistic calibration, β-calibration and g-layers. Below only a selection of those methods is described, initially for a classification task and then for a regression task. See [81][82][83] and the references therein for further information.

#### 3.3.1.1 'Binning-based' approaches

To begin with, consider a classification task and a trained ML model that provides for an input $x$ an estimate $\hat{y}$ of the true class label $y$ with uncalibrated probability $\hat{p}$. For example, $\hat{y}$ might be taken to be the class label $k$ associated with the maximum of the model outputs $p_1(x,\hat{\theta}),\ldots,p_K(x,\hat{\theta})$ evaluated using estimates $\hat{\theta}$ of the model parameters obtained from training and validating the model, and $\hat{p}$ to be the maximum of those outputs. The idea behind a *calibration method* is to adjust the model outputs so that $\hat{p}$ provides a quantification of the ground-truth probability that the true class label for $x$ is $y$, i.e., the model outputs are *calibrated* in the sense that for any $p \in [0,1]$,

$$\mathrm{Prob}(\hat{y} = y|\hat{p} = p) = p.$$

In other words, and viewed from a frequentist perspective, if $n$ inputs $x_i$ are assigned the label $\hat{y}$ with the same probability $\hat{p}$, that probability is 'calibrated' if the number of those inputs that are *correctly* labelled is $\hat{p}n$ in the limit as $n \to \infty$.

Given a sample $x_i, i = 1,\ldots,n$, of inputs with corresponding true class labels $y_i$, the left and right-hand-sides of the above expression can be approximated in the following way. Define $M$ bins $B_m$ by the partition $0 = b_0 < b_1 < \cdots < b_M = 1$ of the interval $[0, 1]$ of possible values of uncalibrated probability. Here, the bins might be chosen to have equal lengths or to contain equal numbers of probabilities. Let $\hat{y}_i$ be the estimate of the class label $y_i$ with probability $\hat{p}_i$ for input $x_i$, $I_m$ the set of indices of those samples for which $\hat{p}_i \in B_m$, and $|I_m|$ the number of samples in $I_m$. Then, for $m = 1, \dots, M$,

$$A(B_m) \equiv \frac{1}{|I_m|} \sum_{i \in I_m} \mathbb{I}(\hat{y}_i = y_i)$$

approximates $\mathrm{Prob}(\hat{y} = y | \hat{p} \in B_m)$, and

$$C(B_m) \equiv \frac{1}{|I_m|} \sum_{i \in I_m} \hat{p}_i,$$

approximates $p \in B_m$. Following [81], $A(B_m)$ and $C(B_m)$ are described, respectively, as the 'accuracy' and 'confidence' of bin $B_m$, and a calibrated model is one for which $A(B_m) = C(B_m), m = 1, \dots, M$. Plotting $A(B_m)$ against $C(B_m)$ produces a *reliability diagram* that provides a visualisation of the degree of calibration of a ML model. For a calibrated model, the graph of $A(B_m)$ against $C(B_m)$ should approximate the identity function. Metrics such as the *Expected Calibration Error* (ECE), defined by

$$\mathrm{ECE} = \sum_{m=1}^{M} \frac{|I_m|}{n} |A(B_m) - C(B_m)|,$$

and the *Maximum Calibration Error* (MCE), defined by

$$\mathrm{MCE} = \max_{m=1,\dots,M} |A(B_m) - C(B_m)|,$$

can be used to quantify the degree of calibration of a ML model (see section 4).

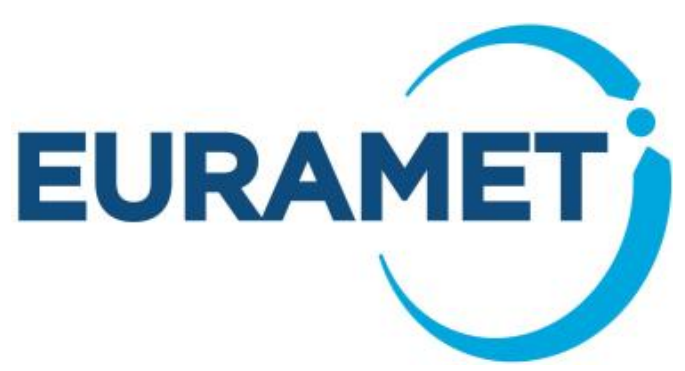

*Histogram binning* and *isotonic regression* are two non-parametric methods based on establishing a mapping from the values $C(B_m)$ to the values $A(C_m)$. Here, they are described in the context of a binary classification task, but the methods may be extended to multi-class classification tasks. Specifically, it is assumed that for each input $x_i, i = 1, \dots, n$, in the calibration set, the ML model provides the uncalibrated probability $\hat{p}_i$ that the class label is $y_i = 1$ corresponding to the 'positive' class and it is desired to provide a calibrated probability, denoted here by $\hat{q}_i$.

Histogram binning is a simple, non-parametric method. The calibration samples are binned as described above, and the calibrated probability for those samples $I_m$ in bin $B_m$ is set as $q_m = A(B_m)$, which is the *observed* proportion of samples in $B_m$ with class label $y_i = 1$. When used to make a prediction for a new input $x$, if the corresponding uncalibrated probability $\hat{p}$ provided by the ML model falls into bin $B_m$ then the calibrated probability is $\hat{q} = q_m$.

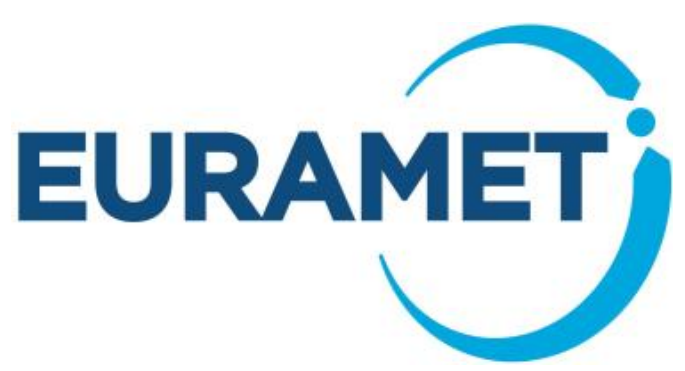

Isotonic regression is another non-parametric method, which can be viewed as a generalisation of histogram binning. In histogram binning, the number of bins and the partitioning of the interval of possible values of uncalibrated probability $\hat{p}$ is pre-selected and fixed. In contrast, in isotonic regression, a non-decreasing function in the form of a piecewise constant function that maps uncalibrated probability $\hat{p}$ to calibrated probability $\hat{q}$ is determined based on a partition that is also estimated from the calibration set. An advantage of isotonic regression is that it ensures monotonicity of the calibrated probabilities. It also has the ability to adapt to the characteristics of the calibration set.

In the context of a regression task, it is assumed that the trained NN has an output layer with two nodes that provide for an input $x$ a prediction $\hat{y}$ for the output variable with uncalibrated variance $\sigma^2$, as in section 3.2.1. Then, histogram binning works by grouping model predictions with similar output variances into bins and adjusting the variances within each bin based on the *observed* errors evaluated using the calibration set. Define $M$ bins $B_m$ by the partition $b_0 < b_1 < \cdots < b_M$ of the interval $[b_0, b_M]$ of possible values of uncalibrated variance $\sigma^2$. Let $\hat{y}_i$ be the estimate of the value $y_i$ of the output variable with uncalibrated variance $\sigma_i^2$ for input $x_i$, $I_m$ the set of indices of those samples for which $\sigma_i^2 \in B_m$, and $|I_m|$ the number of samples in $I_m$. Then, for $m = 1, \ldots, M$,

$$A_{\mathrm{R}}(B_m) \equiv \frac{1}{|I_m|} \sum_{i \in I_m} (\hat{y}_i - y_i)^2$$

is the observed mean squared error, and

$$C_{\mathrm{R}}(B_m) \equiv \frac{1}{|I_m|} \sum_{i \in I_m} \sigma_i^2$$

is the average output variance. These quantities play similar roles to the ‘accuracy’ and ‘confidence’ of bin $B_m$ from the classification case and can be used as the basis of visualising the degree of calibration of a model through a reliability diagram and of metrics to quantify that degree of calibration.

In histogram binning, the calibrated variance for those samples $I_m$ in bin $B_m$ is set as $\hat{\sigma}_m^2 = A_{\mathrm{R}}(B_m)$. When used to make a prediction for a new input $x$, if the corresponding uncalibrated variance $\sigma^2$ provided by the ML model falls into bin $B_m$ then the calibrated variance is $\hat{\sigma}^2 = \hat{\sigma}_m^2$. Isotonic regression, on the other hand, fits a piecewise-constant function to the ‘accuracy’ and ‘confidence’ data that is constrained to be non-decreasing. This method provides a smoother calibration than simple histogram binning and often works better when sufficient calibration data is available.

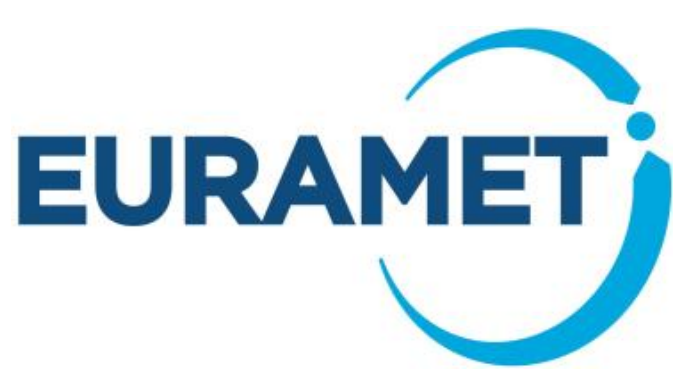

#### 3.3.1.2 'Scaling-based' approaches

For a classification task, *temperature scaling* is a parametric method that operates on the logits $v_k, k = 1, \dots, K,$ provided at the penultimate layer of a NN model rather than on the class probabilities $p_k$ obtained by applying the *softmax* activation function to those logits. The method does not alter the relative ranking of the class probabilities and so has the benefit that it does not alter the classification accuracy of the model. A scalar parameter, known as the 'temperature' and denoted by $T > 0$, is introduced and applied to the logits before passing them to the *softmax* activation function to provide calibrated probabilities

$$q_k = \frac{\exp\left(\frac{v_k}{T}\right)}{\sum_{k=1}^{K} \exp\left(\frac{v_k}{T}\right)}, \qquad k = 1, \dots, K.$$

An estimate of the temperature $T$ is obtained by minimising a loss function, such as the NLL loss function

$$-\log p(y_1, \dots, y_n | T) = -\sum_{i=1}^{n} \sum_{k=1}^{K} \mathbb{I}(y_i = k) \log q_k(x_i, \hat{\theta}, T)\,,$$

evaluated for the calibration set $\{(x_i, y_i): i = 1, \dots, n\}$ using the pre-trained model defined by parameter estimates $\hat{\theta}$. For $T > 1$, the effect of the temperature parameter is to 'soften' the *softmax* activation function, with higher values of $T$ yielding a more uniform distribution of class probabilities representing greater normalised entropy and greater uncertainty in the predicted class label. A limitation of the method is that it assumes that a single scalar parameter can describe the calibration of class probabilities, which may not be sufficient when the nature of their miscalibration is more complex.

In the context of a regression task, temperature scaling is sometimes known as *variance scaling*. As for the 'binning-based' approaches described above, it is assumed that the trained NN has an output layer with two nodes that provide for an input $x$ values for the mean $\mu(x, \theta)$ and the variance $\sigma^2(x, \theta)$ defining a Gaussian probability distribution for the output variable, as in section 3.2.1. The uncalibrated variance $\sigma^2(x, \theta)$ is scaled by the temperature $T > 0$ and is estimated by minimising a loss function, such as the NLL loss function

$$-\log p(y_1, \dots, y_n | T) = \sum_{i=1}^{n} \left\{ \frac{\log T\sigma^2(x_i, \hat{\theta})}{2} + \frac{\left(y_i - \mu(x_i, \hat{\theta})\right)^2}{2T\sigma^2(x_i, \hat{\theta})} \right\} + n \log \sqrt{2\pi}\,,$$

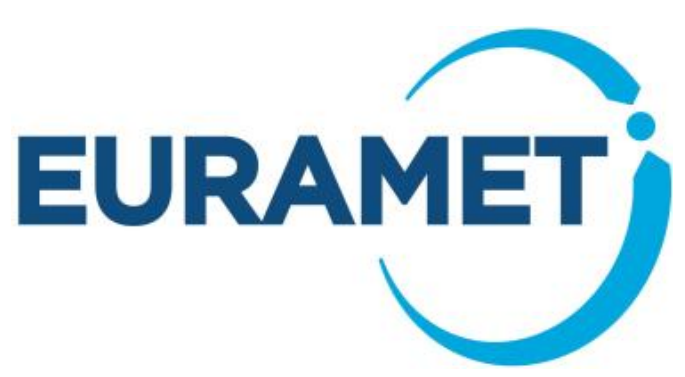

evaluated for the calibration set $\{(x_i, y_i): i = 1, \ldots, n\}$. Then, for a new input $x$, the calibrated variance is evaluated as $\hat{T}\sigma^2(x, \hat{\theta})$ in terms of the estimated temperature $\hat{T}$ and the uncalibrated variance $\sigma^2(x, \hat{\theta})$ output by the trained NN.

### 3.3.2 Conformal Prediction

The description of conformal prediction given here is largely based on the introduction [84] and the lecture [85]. The focus is on *split conformal prediction* in which the calibration set is kept distinct from the set of data used to train the ML model, which is in contrast to *full conformal prediction* that is more statistically efficient but generally much more computationally expensive and is rarely used in practice.

Conformal prediction is a framework for providing the uncertainty for a prediction obtained from any pre-trained model that already has an inherent notion of 'uncertainty'. For a classification task and arbitrary input value $x$, the uncertainty is expressed in the form of a *prediction set* $\mathcal{C}(x)$ of possible values of the class label $y \in \{1, 2, \ldots, K\}$. For a regression task, the uncertainty is expressed in the form of a *prediction interval* $\mathcal{C}(x)$ of possible values of the output quantity $y \in \mathcal{R}$. The uncertainty is statistically valid in the sense that, under the assumption of exchangeability, $\mathcal{C}(x)$ contains the true output value $y^*$ corresponding to $x$ with an approximately user-specified probability $1 - \alpha$. Specifically,

$$1 - \alpha \leq \mathrm{Prob}\big(y^* \in \mathcal{C}(x)\big) \leq 1 - \alpha + \frac{1}{n + 1},$$

where $n$ is the size of the calibration set used in the calculation of $\mathcal{C}(x)$. Importantly, the approach is distribution-free in the sense that the uncertainty is statistically valid even without distributional assumptions about the data or assumptions about the quality of the pre-trained model.

Given a pre-trained model and the calibration set $\{(x_i, y_i): i = 1, \ldots n\}$, the framework of conformal prediction is based on the following steps. Firstly, define a *score function* $s(x, y)$ that captures an inherent notion of 'uncertainty' in the sense that the score is high when confidence in the output value $y$ for input value $x$ is low. For example, for a classification task with class label $y \in \{1, 2, \ldots, K\}$, the score might be evaluated as $1 - \hat{p}$ where $\hat{p}$ is the *softmax* output provided by the model for class label $y$. The score approaches one as $\hat{p}$ approaches zero and approaches zero as $\hat{p}$ approaches one. Secondly, calculate scores $s_i = s(x_i, y_i), i = 1, \ldots, n$, using the samples in the calibration set. For example, for a classification task and the score function defined above, $s_i = 1 - \hat{p}_i$ where $\hat{p}_i$ is the *softmax* output for the true class $y_i$. Thirdly, evaluate $\hat{q}$ as the $\lceil (n + 1)(1 - \alpha) \rceil / n$ empirical quantile of the scores $s_i, i = 1, \ldots, n$, where $\lceil \cdot \rceil$ is the 'ceiling' function that rounds up its argument to the nearest integer. Finally, calculate the prediction set or interval $\mathcal{C}(x) = \{y: s(x, y) \leq \hat{q}\}$. For example, for a classification task and the score function defined above, $\mathcal{C}(x)$ contains those class labels $y \in \{1, 2, \ldots, K\}$ for which $s(x, y) \leq \hat{q}$. In other words, the prediction set is composed only of those class labels for which the *softmax* output is high enough (or the score function is low enough).

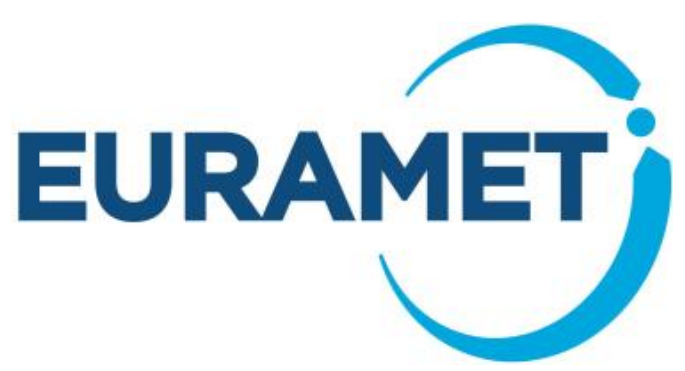

Conformal prediction constructs a different prediction set (or interval) $\mathcal{C}(x)$ adaptively for each input value $x$. The set becomes larger (or the interval wider) when the model is uncertain, or it is intrinsically difficult to make a prediction for the input value. These are desirable properties, because the size of the set (or width of the interval) gives a natural indicator of the uncertainty of the model prediction. Furthermore, $\mathcal{C}(x)$ is interpreted as a set of plausible classes (or output values) $y$ for the input value $x$.

Although the prediction set (or interval) is always statistically valid, in the sense described above, the usefulness of the prediction set depends primarily on the choice of score function. This follows because the score function incorporates almost all the information about the problem and data, including the underlying model itself. There are many possible score functions for an underlying model, and they can have different properties. Constructing the right score function is an important choice when using conformal prediction.

In the context of regression tasks, one approach is to use the score function $s(x,y) = \left|y - f(x,\hat{\theta})\right|$ equal to the absolute residual deviation, where $f(x,\hat{\theta})$ is the point prediction returned by the pre-trained model for input value $x$. Then, following the steps above,

$$\mathcal{C}(x) = \left\{y: \left|y - f\left(x,\hat{\theta}\right)\right| \leq \hat{q}\right\} = \left\{y: f\left(x,\hat{\theta}\right) - \hat{q} \leq y \leq f\left(x,\hat{\theta}\right) + \hat{q}\right\},$$

where $\hat{q}$ is an empirical quantile of the scores $s_i = \left|y_i - f(x_i,\hat{\theta})\right|, i = 1, \dots, n$, calculated in terms of the calibration set. The score function has the required property that it is high when the absolute residual deviation is high and confidence in the prediction is low. However, the approach has the undesirable property that the width of the prediction interval is the same for all input values $x$, and it lacks the property of local adaptivity that captures the idea that the uncertainty of a prediction can be different for different input values $x$.

Two ways of addressing this property are described as follows. Consider a NN model whose outputs are the estimate $\mu(x,\hat{\theta})$ and its variance $\sigma^2(x,\hat{\theta})$ in a Gaussian model $\mathrm{N}(\mu\left(x,\hat{\theta}\right), \sigma^2\left(x,\hat{\theta}\right))$ for the output variable, as in section 3.2.1. The score function $s(x,y)$ is defined to be the 'studentised' absolute residual given by

$$s(x,y) = \frac{\left|y - \mu\left(x,\hat{\theta}\right)\right|}{\sigma(x,\hat{\theta})},$$

and the prediction interval is

$$\mathcal{C}(x) = \left\{y: \frac{\left|y - \mu\left(x,\hat{\theta}\right)\right|}{\sigma(x,\hat{\theta})} \leq \hat{q}\right\} = \left\{y: \mu\left(x,\hat{\theta}\right) - \hat{q}\sigma\left(x,\hat{\theta}\right) \leq y \leq \mu\left(x,\hat{\theta}\right) + \hat{q}\sigma(x,\hat{\theta})\right\},$$

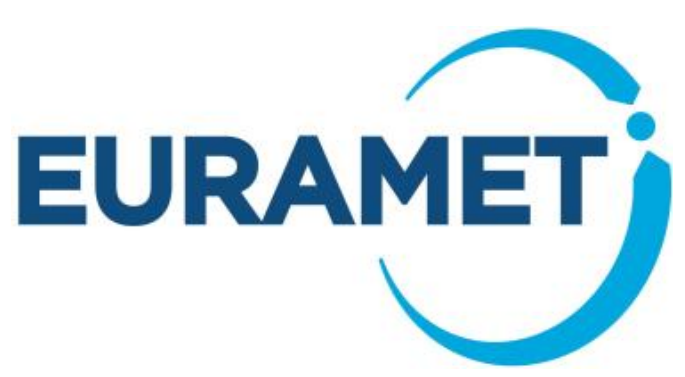

whose width adapts according to the predicted uncertainty $\sigma(x,\hat{\theta})$. Note that the interval is statistically valid in the sense described above irrespective of the actual distribution of the output quantity.

Alternatively, consider the method of *quantile regression* (QR) in which a model is trained to directly estimate quantiles of the distribution of the output quantity. For example, the model can be trained to estimate the quantiles $t_{\alpha/2}(x)$ and $t_{1-\alpha/2}(x)$ corresponding to probabilities $\alpha/2$ and $1-\alpha/2$ that define the interval

$$\{y: t_{\alpha/2}(x) \le y \le t_{1-\alpha/2}(x)\}$$

to contain the output quantity with approximate probability $1-\alpha$. In 'conformalised' QR [86], the score function is defined as the difference between $y$ and the nearest quantile, i.e.,

$$s(x,y) = \max\{t_{\alpha/2}(x) - y, y - t_{1-\alpha/2}(x)\},$$

and the prediction interval is

$$\mathcal{C}(x) = \{y: t_{\alpha/2}(x) - \hat{q} \le y \le t_{1-\alpha/2}(x) + \hat{q}\}$$

whose width also adapts to the predicted uncertainty.

Returning to the classification task used as the example at the start of the section, the use of simple likelihood scores as in the example produces prediction sets with the smallest average size [87], but they can tend to 'undercover' for hard cases and 'overcover' for easy cases. Alternative score functions have been proposed to address this characteristic of the method, such as cumulative likelihood scores in the *adaptive prediction sets* (APS) method. However, these methods are generally used in the context of multi-class classification tasks with many classes and can be difficult to interpret in the case of binary classification problems. Venn-Abers prediction [88] is another approach for classification tasks that provides prediction sets with statistical guarantees, is distribution-free and can be applied to binary classification tasks. The approach can be used to transform the (point) predicted probability provided by a pre-trained model into a *probability interval* that represents the uncertainty of the prediction. It is based on applying isotonic regression to the calibration set augmented with the new input value $x$ and the label '1' or '2' (for a binary classification problem) to obtain two functions that are used to construct the probability interval. See [88] for further details.

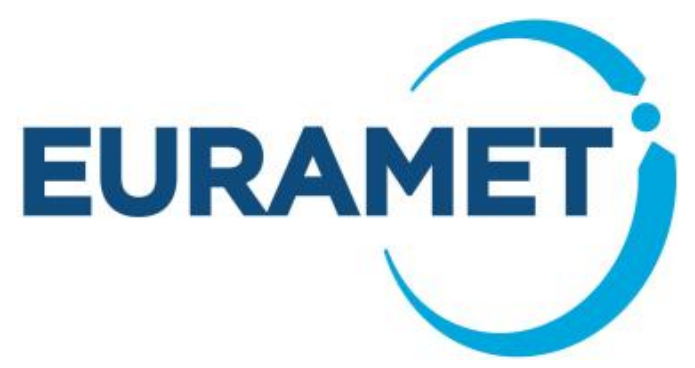

# 4 Validation of Uncertainty

## 4.1 Background

As discussed in section 1.1, decision-making based on the prediction provided by a ML model, for example, to make a diagnosis about the state of health of a patient, requires knowledge of the uncertainty of the prediction. Furthermore, making a *good* decision depends on knowing that the uncertainty is reliable because it is then possible to have confidence in the decision and to quantify the risk of the decision being wrong. The reliability of the uncertainty is established through a framework for validating the uncertainty, which is the topic of this section.

The idea of well-calibrated uncertainties (specifically, well-calibrated variances or confidence/credible intervals in the case of regression tasks and well-calibrated class probabilities in the case of classification tasks) was introduced in section 3.3.1. *Calibration* captures the extent to which calculated uncertainties reflect the distribution of observed prediction errors. For example, in a regression task, a calculated uncertainty expressed as an uncertainty interval for specified probability $p$ (0.95, say) might be considered well-calibrated if the interval contains precisely a fraction $p$ of observed prediction errors for those test samples having that uncertainty.

An assessment of the reliability of the uncertainties accompanying the predictions provided by a ML model should consider how those predictions will be used in practice. For example, if a clinical diagnosis is to be based on a single prediction, then what matters is the reliability of the uncertainty of that prediction ('individual reliability') and not an aggregated reliability assessed over the complete test set ('global reliability') or a subset of the test set ('local reliability'). Such considerations can affect the choice of reliability metric.

Local reliability metrics are often applied to subsets of the test set obtained by binning the samples according to the magnitude of their uncertainties and are sometimes referred to as 'size-stratified metrics'. However, it can be helpful to consider subsets obtained by binning the samples based on other characteristics and then applying local reliability metrics to each subset. For example, for a classification task, it may be of interest to assess the reliability of the uncertainties evaluated for each individual class (so, binning by class) and, for a regression task, to understand how the reliability depends on signal quality (so, binning by signal-to-noise ratio of the input signals). Evaluating reliability metrics for different subsets provides an assessment of the adaptivity of a ML model.

The use of local calibration techniques for regression models is motivated by the need to go beyond global calibration and address small-scale local reliability. Indeed, it is possible to have an uncertainty quantification that is globally well-calibrated, but which is not sufficiently discriminative because the uncertainties are large (sometimes due to poor predictive performance of the model). *Sharpness* metrics, which capture the extent to which calculated uncertainty intervals concentrate around predicted values, can be used in conjunction with calibration metrics to address this issue. It has been argued that both calibration and sharpness are important in validating uncertainty [77], and a desirable UQ method is one that maximises sharpness subject to the constraint of being well-calibrated.

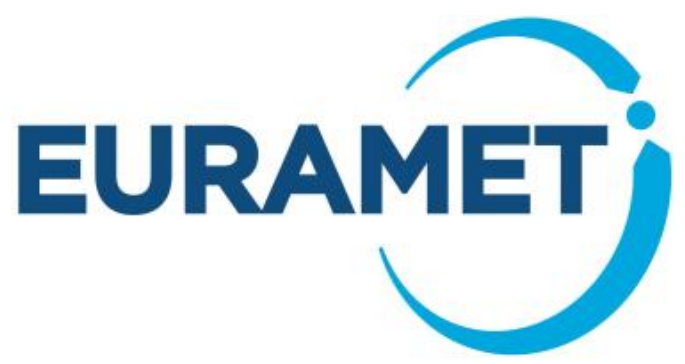

Practical issues that arise when applying different evaluation metrics concern the binning of the samples for evaluating local reliability metrics and how UQ methods use different ways to express and report uncertainty.

The binning strategy can have an appreciable effect on the interpretation and comparison of validation results. The Expected Normalised Calibration Error (ENCE) is a popular reliability metric used for regression tasks (section 4.3), but it is noted in [89] that it has the undesirable property that it scales with the square root of the number of bins for well calibrated or nearly calibrated test sets. The Expected Calibration Error (ECE) is a popular reliability metric used for classification tasks (section 4.2), which is usually implemented with bins of equal length, but it is noted in [90] that such a choice means that often only a few bins contribute to the value of the metric. Consequently, metrics based on alternative and adaptive binning strategies, or that use kernel density estimation, can be preferable.

For classification tasks, uncertainty is sometimes reported as the probability of the single predicted class and sometimes as the normalised entropy of the probability distribution defined by all the class probabilities. For binary classification, there is a nonlinear correspondence between the two expressions, and different reliability metrics are used to assess them: for example, Expected Calibration Error (ECE) for uncertainties expressed as a single class probability, and Uncertainty Calibration Error (UCE) for uncertainties expressed as a normalised entropy. Normalised entropy is generally preferred for reporting uncertainty for the case of multi-class classification, because it captures how the probabilities are spread across the complete set of classes in a normalised fashion.

For regression tasks, uncertainty is sometimes reported as a predicted variance and sometimes as a prediction interval, and these are different expressions of the dispersion of possible values of the measurand. The former is a summary statistic (the second central moment) of the probability distribution for the measurand and measures the average squared deviation from the expected value, whereas the latter is an interval that contains values of the measurand with a stated probability and is defined by quantiles of the distribution. Converting between the two expressions requires knowledge of the probability distribution of the measurand, and that distribution is often assumed to be Gaussian for that purpose. For regression tasks, some reliability metrics are 'variance-based' and assess how well a predicted variance describes the sample variance of observed prediction errors. Other metrics are 'coverage-based' and assess how well a prediction interval approximates certain quantiles of the empirical distribution of observed prediction errors. Finally, some metrics assume a distribution for the measurand and, again, it is often convenient to choose a Gaussian distribution for that purpose, which helps to support the comparability of validation results.

In the following sections, various reliability metrics are summarised for classification and regression tasks. The metrics consider the complementary characteristics of calibration and sharpness and assess local and global reliability as well as adaptivity. See [3][4] and the references therein for further information.

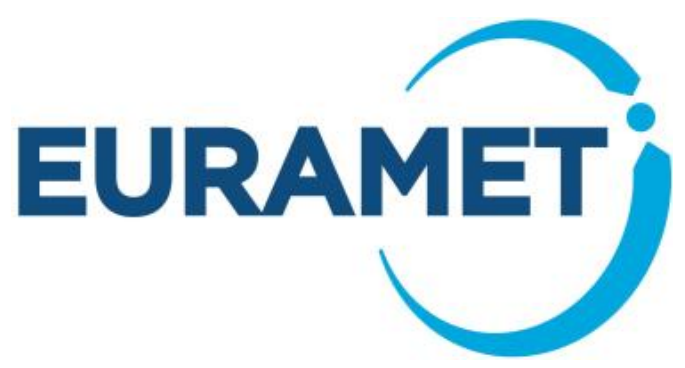

## 4.2 Evaluation Metrics for Classification Tasks

The local reliability metric of *Expected Calibration Error* (ECE) [81], defined by

$$\mathrm{ECE} = \sum_{m=1}^{M} \frac{|I_m|}{n} |A(B_m) - C(B_m)|,$$

was introduced in section 3.3.1.1 in the context of binning-based approaches to calibration methods for UQ. The metric is based on partitioning the interval $[0, 1]$ of possible values of the predicted class probability into bins $B_m, m = 1, \ldots, M,$ and forming the weighted average of the absolute differences between the accuracy $A(B_m)$ and confidence $C(B_m)$ for each bin. Here, $n$ is the number of samples in the test set, $I_m$ are the indices of those samples for which the predicted class probability falls into bin $B_m$, and $|I_m|$ is the number of samples in $B_m$. If $\hat{y}_i$ is the estimate of the class label $y_i$ with probability $\hat{p}_i$ for test sample $x_i$, recall from section 3.3.1.1 that for each bin $B_m$,

$$A(B_m) \equiv \frac{1}{|I_m|} \sum_{i \in I_m} \mathbb{I}(\hat{y}_i = y_i)$$

approximates $\mathrm{Prob}(\hat{y} = y | \hat{p} \in B_m)$, and

$$C(B_m) \equiv \frac{1}{|I_m|} \sum_{i \in I_m} \hat{p}_i,$$

approximates $p \in B_m$.

A disadvantage of ECE is that it is computed using only the probability for the single predicted class, which is the largest of the output probabilities. The metric does not consider the reliability of the output probabilities for the other $K - 1$ classes. For a clinician, a diagnosis informed by the second or third highest class probabilities may also be informative and for that it is important that the other output probabilities are also reliable. Another disadvantage is that ECE is usually implemented with bins of equal length. However, the distribution of predicted class probability is often skewed, with bins at the lower end of the interval $[0, 1]$ being sparsely populated and those at the upper end being densely populated. Consequently, in practice only a few of the bins contribute to the value of ECE. Additionally, there is a 'bias-variance trade-off' associated with the choice of the number of bins because that number determines how many samples fall into each bin and therefore the quality of the estimates of accuracy and confidence. The trade-off is further exacerbated by the use of bins of equal length, which produces large differences between the numbers of samples falling into the different bins.

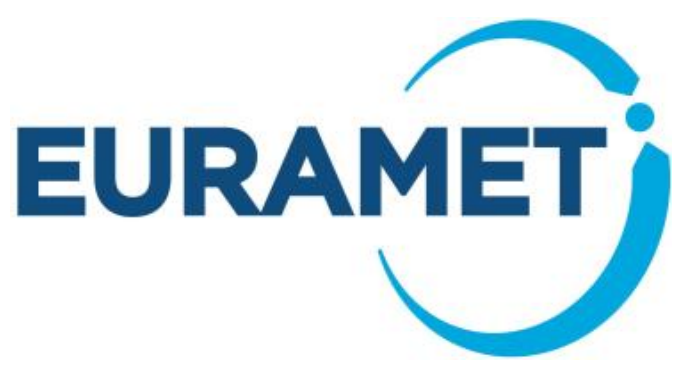

The local reliability metric of *Adaptive Calibration Error* (ACE) [90] goes some way to addressing these disadvantages. It is based on selecting bins $B_{m,k}$ *adaptively* for each class $k = 1, \dots, K$, to contain approximately the same number of samples, and is evaluated using

$$\mathrm{ACE} = \frac{1}{KM} \sum_{k=1}^{K} \sum_{m=1}^{M} \left| A\left(B_{m,k}\right) - C\left(B_{m,k}\right) \right|,$$

where $A(B_{m,k})$ and $C(B_{m,k})$ are, respectively, the accuracy and confidence for the $m^{\text{th}}$ bin selected for the $k^{\text{th}}$ class. The motivation for the metric is twofold. Firstly, the differences between accuracy and confidence are averaged over bins (as for ECE) but then over classes. Secondly, there is a focus on selecting bins that contain similar numbers of predictions.

The reliability metric of *smoothed Expected Calibration Error* (smECE) [91] provides another way to address the disadvantage of a 'binning-based' approach, such as ECE, to assess the calibration of a ML model. It gives a better-behaved measure of (mis)calibration by applying kernel smoothing to data for the test set, for example using a radial basis function or Gaussian kernel, rather than binning the data. Consider a binary classification task and let $\hat{y}_i \in \{1,2\}$ be the estimate of the class label $y_i = 1$ with probability $\hat{p}_i = \mathrm{Prob}(\hat{y}_i = 1)$ for test sample $x_i, i = 1, \dots, n$. Then, the calculation of ECE is based on using binning (or histogram regression) to replace the data $(\hat{p}_i, \hat{y}_i), i = 1, \dots, n$, by data $\left(C(B_m), A(B_m)\right), m = 1, \dots, M$, representing the positions and heights of a set of histogram bars. It then measures the departure of the histogram defined by those bars from the idealised calibration function corresponding to a straight-line passing through the origin and having a slope of unity. The idea behind the metric smECE is to instead replace the data $(\hat{p}_i, \hat{y}_i), i = 1, \dots, n$, by a smooth function and to measure the departure of that smooth function from the idealised calibration function. In practice, the smoothing is applied to the data $(\hat{p}_i, \hat{y}_i - \hat{p}_i), i = 1, \dots, n$. The approach includes a principled choice for the bandwidth parameter of the kernel function, which means that the metric can be considered as being hyperparameter-free, simplifying its application and comparison across different ML models. Other advantages of the metric are that it is a continuous function of the model predictions (i.e., small changes in the predictions lead to small changes in the value of the metric), and it provides a smooth and visually interpretable representation of the (mis)calibration of a ML model. Furthermore, it comes with some theoretical properties, for example, that it is a *consistent* calibration metric in the sense described in [92], and it corresponds to a natural notion of distance, i.e., if the value of the metric is $\epsilon$ then the model can be post-processed to make it perfectly calibrated without perturbing the model by more than $\epsilon$ as measured in the $L_1$ norm.

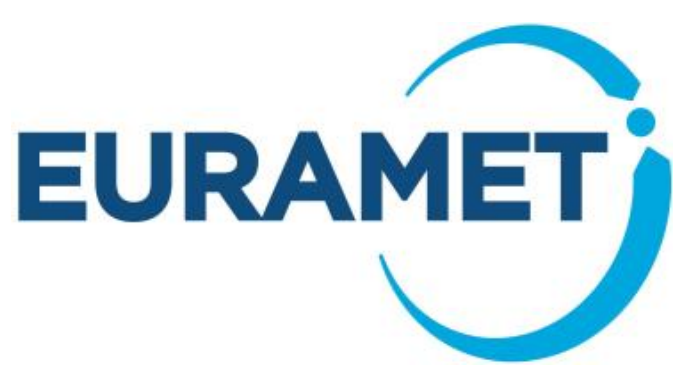

Another local reliability metric is *Uncertainty Calibration Error* (UCE) [93], which considers normalised entropy in place of predicted class probability as the expression of uncertainty. Using bins selected in a similar way as for ECE, the metric is calculated from

$$\mathrm{UCE} = \sum_{m=1}^{M} \frac{|I_m|}{n} |E(B_m) - U(B_m)|,$$

where, for $m = 1, \dots, M$,

$$E(B_m) \equiv \frac{1}{|I_m|} \sum_{i \in I_m} \mathbb{I}(\hat{y}_i \neq y_i)$$

is the 'error' (or 'inaccuracy' or 'misclassification rate') of bin $B_m$, which approximates $\mathrm{Prob}(\hat{y} \neq y | H(\hat{y}) \in B_m)$, and

$$U(B_m) \equiv \frac{1}{|I_m|} \sum_{i \in I_m} H(\hat{y}_i),$$

is the 'uncertainty' of bin $B_m$, which approximates the normalised entropy $H(y) \in B_m$, and is equal to the average of the normalised entropies for test samples for which their normalised entropy falls into bin $B_m$. In contrast to ECE, but similarly to ACE, the metric considers the output probabilities for all classes. However, in contrast to ACE, the metric is not as highly sensitive to the number of bins and, moreover, provides a consistent ranking of different models for the same classification task.

While some justification is given in [93] for the use of UCE, it is not clear why the error $E(B_m)$ and uncertainty $U(B_m)$ as defined above are expected to be equal for well-calibrated uncertainties. In [94] a family of evaluation metrics is defined in which the predicted and observed ranked class probabilities are compared using a general measure of variation such as normalised entropy. The resulting metric, referred to as *Variation Calibration Error* (VCE), provides a generalisation of ECE, which is recovered when the measure of variation is taken to be the predicted class probability. The reference [94] presents some simple numerical experiments using simulated predictions whose uncertainties are well-calibrated by design. It is shown that the uncertainties are well-calibrated according to both ECE and VCE, but poorly calibrated according to UCE. The experiments provide numerical evidence that the more intuitive metric of VCE is a more useful indicator of the reliability of uncertainties than UCE.

The *negative-log-likelihood* (NLL) loss function was introduced in section 3.2.2 as the basis for training a ML model such as a NN for classification tasks and is commonly used to measure the performance of such models. The loss function is minimised when the class probabilities estimated in the training yield estimates of the class labels that match the true class labels. However, making the NLL small can lead to overconfident predictions because the loss function penalises a small class probability assigned to the correct class (leading to overestimation of that probability) and large class probabilities assigned to incorrect classes (leading to underestimation of those probabilities). Despite that disadvantage, NLL is often used as a metric to assess the reliability of uncertainties expressed as class probabilities, and for a binary classification task takes the form

$$\mathrm{NLL} = -\sum_{i=1}^{N} y_i \log \hat{p}_i + (1 - y_i) \log(1 - \hat{p}_i)\,,$$

which is called the binary cross entropy (BCE). NLL is a proper scoring rule, which means that it captures both calibration and sharpness.

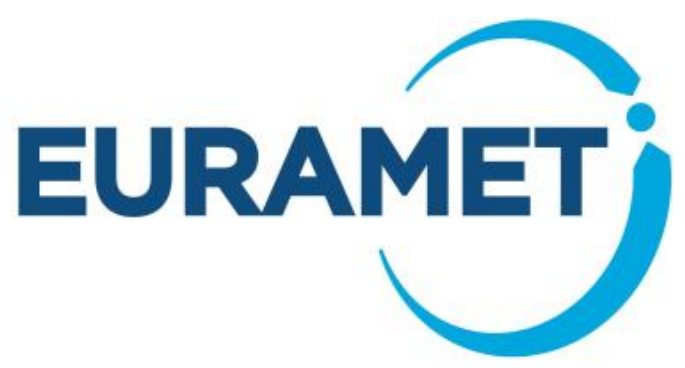

## 4.3 Evaluation Metrics for Regression Tasks

The local reliability metric of *Expected Normalised Calibration Error* (ENCE) [95], defined by

$$\mathrm{ENCE} = \frac{1}{M} \sum_{m=1}^{M} \frac{\left|\sqrt{A_{\mathrm{R}}(B_m)} - \sqrt{C_{\mathrm{R}}(B_m)}\right|}{\sqrt{C_R(B_m)}},$$

is based on the metric ECE introduced in section 4.2 in the context of classification tasks. The idea is to partition the interval of possible values of predicted variance $\sigma^2$ into bins $B_m, m = 1, \dots, M$, and form the weighted average of the absolute differences between the root mean square error and root mean variance for each bin. Here, as before, $I_m$ are the indices of those test samples for which the predicted variance $\sigma^2$ falls into bin $B_m$, and $|I_m|$ is the number of samples in $B_m$. If $\hat{y}_i$ is the estimate of the value $y_i$ of the output variable with variance $\sigma_i^2$ for input $x_i$, recall from section 3.3.1.1 that for each bin,

$$A_{\mathrm{R}}(B_m) \equiv \frac{1}{|I_m|} \sum_{i \in I_m} (\hat{y}_i - y_i)^2$$

is the observed mean squared error, and

$$C_{\mathrm{R}}(B_m) \equiv \frac{1}{|I_m|} \sum_{i \in I_m} \sigma_i^2$$

is the average predicted variance. As for other local metrics that use binning, there are different ways to construct the bins, ranging from the simple approach of choosing bins of equal width to adaptive strategies that aim to balance the numbers of samples within the bins. A disadvantage of the metric is that it is not resistant to the effect of outliers [89]. Note that plotting predicted variance against prediction error can provide insight into small-scale and local reliability and allow the identification of outliers.

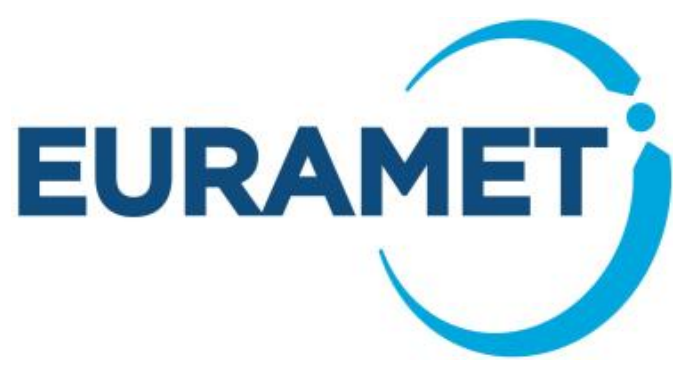

ENCE considers the case when uncertainty is reported as a predicted variance $\sigma^2$. In contrast, 'coverage-based' metrics treat the cases when uncertainty is expressed either (a) as a prediction interval $I_p$ containing values of the output variable $Y$ with a stated probability $p$:

$$\mathrm{Prob}\left(Y \in I_p\right) = p,$$

or (b) as a quantile $y_p$, which is the value of $Y$ such that the probability that $Y$ is less than the value is a stated probability $p$:

$$\mathrm{Prob}\left(Y \leq y_p\right) = p,$$

or (c), more generally, as a distribution function $p = F(\eta)$ or its inverse, the quantile function $\eta = F^{-1}(p)$, that satisfy

$$F(\eta) = \mathrm{Prob}(Y \leq \eta), \qquad p = \mathrm{Prob}(Y \leq F^{-1}(p))$$

for general probabilities $p, 0 \leq p \leq 1$. Note that both $I_p$ and $y_p$ can straightforwardly be derived from the quantile function.

For a specific probability $p$, the coverage-based metric *Prediction Interval Coverage Probability* (PICP) [96] is defined by

$$\mathrm{PICP}(p) = \frac{1}{n}\sum_{i=1}^{n} \mathbb{I}(\hat{y}_i \in I_{p,i}),$$

when the uncertainty is expressed as a prediction interval $I_{p,i}$ for the $i$th test sample, or by

$$\mathrm{PICP}(p) = \frac{1}{n}\sum_{i=1}^{n} \mathbb{I}(\hat{y}_i \leq y_{p,i}),$$

when the uncertainty is expressed as the quantile $y_{p,i}$ for the $i$th test sample. In either case, a reliable uncertainty corresponds to $\mathrm{PICP}(p) \approx p$. The metric can be evaluated for different choices of $p$, and particular choices are $p = 0.6826$ and $p = 0.9545$, which, for a Gaussian distribution, correspond to prediction intervals with half-widths equal to, respectively, one and two standard deviations.

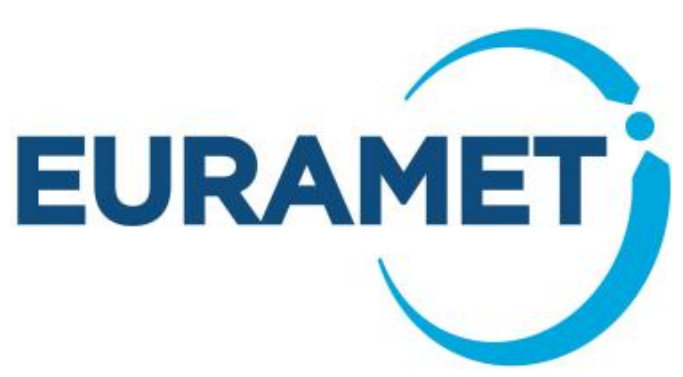

When quantiles $y_{j,i}$ are reported for $n_p > 1$ probabilities $p_j, j = 1, \dots, n_p$, and the $n$ test samples $x_i, i = 1, \dots, n$, then the *Coverage Calibration Error* (CCE) combines that information into a single coverage-based metric, defined by

$$\mathrm{CCE} = \sum_{j=1}^{n_p} \left(\rho_j - p_j\right)^2.$$

Here, $\rho_j$ is the observed probability calculated from

$$\rho_j = \frac{1}{n} \sum_{i=1}^{n} \mathbb{I}(\hat{y}_i \leq y_{j,i}),$$

where $y_{j,i} = F_i^{-1}(p_j)$ is evaluated from the quantile function $F_i^{-1}(p)$ for the $i$th test sample. Reliable uncertainties correspond to values of CCE that are close to zero.

Finally, when the distribution function $F_i(\eta)$ is reported for the $i$th test sample $x_i$, the metric *Continuous Ranked Probability Score* (CRPS) [97] is defined by

$$\mathrm{CRPS}(F_i, y_i) = \int_{-\infty}^{\infty} (F_i(\eta) - \mathbb{I}(\eta \leq y_i))^2 \,\mathrm{d}\eta,$$

where $\mathbb{I}(\eta \leq y_i)$ represents the (degenerate) distribution function based on the ground truth value $y_i$. The scores for the different test samples can then be aggregated using a simple or weighted average, for example, to obtain

$$\widehat{\mathrm{CRPS}} = \frac{1}{n} \sum_{i=1}^{n} \mathrm{CRPS}(F_i, y_i),$$

and reliable uncertainties correspond to a value of this aggregated score that is close to zero. An advantage of CRPS is that it is a proper scoring rule, which means that it captures both calibration and sharpness. Note that when the predicted probability distribution is the Gaussian distribution $\mathrm{N}(\mu, \sigma^2)$, then $\mathrm{CRPS}(F, y)$ has the algebraic form

$$\mathrm{CRPS}(F, y) = \sigma \left\{ 2\phi\left(\frac{y-\mu}{\sigma}\right) + \left(\frac{y-\mu}{\sigma}\right)\left[2\Phi\left(\frac{y-\mu}{\sigma}\right) - 1\right] - \frac{1}{\sqrt{\pi}} \right\},$$

where $\phi(\cdot)$ and $\Phi(\cdot)$ denote, respectively, the probability density function and distribution function of the standard Gaussian distribution $\mathrm{N}(0,1)$.

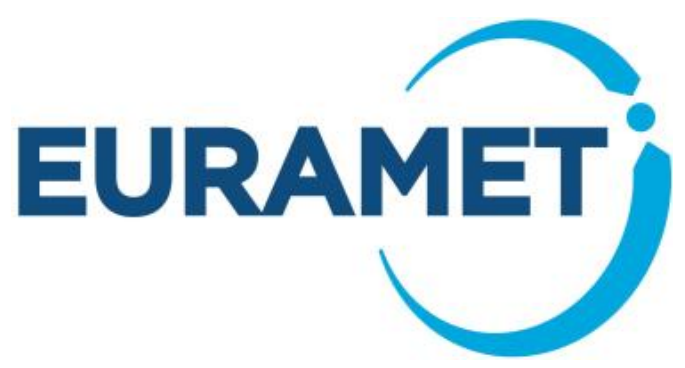

Another way to assess the sharpness of the predicted distribution or prediction interval derived from that distribution is to use the *Mean Prediction Interval Width* (MPIW) [96], which is the average width of the prediction intervals evaluated for stated probability $p$, i.e., if the prediction intervals for the $n$ test samples are denoted by $[L_i, U_i]$ then

$$\mathrm{MPIW}(p) = \frac{1}{n}\sum_{i=1}^{n}(U_i - L_i).$$

For example, if the prediction intervals are derived from a Gaussian distribution, then for probability $p = 0.6826$, the MPIW is simply twice the average of the predicted standard deviations $\sigma_i$. MPIW is often used in conjunction with a coverage-based metric such as PICP. The goal is to achieve a PICP close to $p$, indicating the frequency of prediction intervals containing the true values is as expected (well-calibrated), and a MPIW close to zero, indicating predicted values are precise (sharp).

# 5 Benchmark Datasets

The report [5] describes six benchmark problems covering both regression and classification tasks applied to PPG signals that are summarised in section 1.3. For each benchmark problem the report identifies benchmark datasets that are designed to support the development and evaluation of ML models and UQ methods for PPG signal analysis. The datasets are tailored to address the challenges of ensuring the accuracy and reliability of model results and are intended for the medical device and digital health communities to facilitate model evaluation. The report lists potential benchmark datasets and gives guidance for a selection of those about accessing and using the datasets. Here, only high-level summary descriptions of the benchmark datasets are provided.

## 5.1 Determination of Systolic and Diastolic Blood Pressure

The *AuroraBP* dataset [98] consists of PPG (and other) signals for which auscultatory or oscillometric measurements of systolic and diastolic BPs are simultaneously made. Signals were collected over a 24-hour period for 1,125 subjects with ages 21 to 85 years, an average age of 45.1 ± 11.3 years, 49.2 % of whom were female, and with multiple blood pressure categories ranging from non-elevated to elevated (hypertension). The PPG sensor was placed on the anterior surface of the arm, which is in contrast to smartwatches worn on the posterior surface, and the signals were resampled to a frequency of 500 Hz. Signals were acquired from clinical laboratory settings and community settings in daily life. The dataset contains 38,623 signals of varying durations in the range 9.1 s to 87.1 s with an average duration of 19.73 s.

The *PulseDB* dataset [99] is a collection of high-quality signals from two datasets: *VitalDB* comprising 2,938 subjects and *MIMIC-III Waveform Database Matched Subset* comprising 2,423 subjects from surgery and intensive care unit settings. The *VitalDB* dataset, which was the focus of the studies in the QUMPHY project, contains synchronised raw and filtered physiological signals from an ECG, a finger-based PPG, and reference invasive BP measurements. The dataset includes patients during perioperative periods with respect to various surgical operations (general, thoracic, urological, and gynaecological). The dataset provides information about patient age, height, weight, and body-mass index. The average age of subjects is 58.76 ± 15.02 years and 54.73 % are male. The PPG signals and continuous BP measurements were sampled at a rate of 125 Hz. The signals were segmented into 10 s windows, providing single average values of systolic BP and diastolic BP for each segment. All 10 s segments related to PPG or BP with outliers or anomalies were removed. The average duration of patient signals is 1.38 ± 1.18 h with a range of 10.08 s to 9.16 h. In addition, the dataset provides annotated fiducial points for ECGs (R peaks), PPGs (peaks and onsets), and BP measurements (peaks and onsets), which can be used to obtain beat-to-beat systolic and diastolic BP sequences.

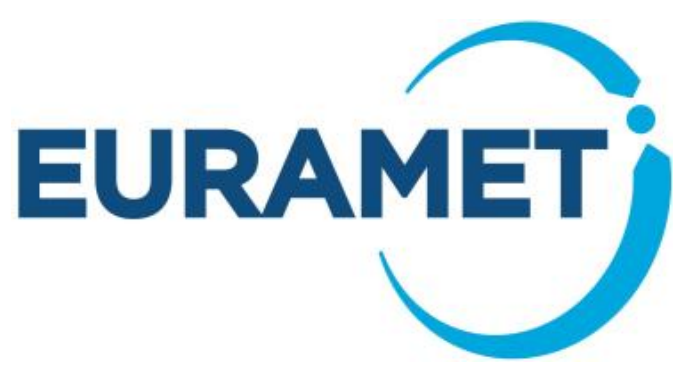

## 5.2 Detection of Atrial Fibrillation

Five benchmark datasets are considered, the first three of which (*DeepBeat* [100], *TriggersAF* [101] and *UMMC Simband* [102][103]) involve wrist-based measurements and the fourth of which (*MIMIC-III-Ext-PPG* [104]) involves fingertip-based measurements. The fifth dataset (*PPGArrhythmiaDetection* [100]) also involves finger-tip based measurements and is used within the QUMPHY project as an external test dataset.

*DeepBeat* is a preferred dataset due to its large scale and existing code support, despite some label noise. The dataset comprises over 3,336,185 PPG segments of 25 s duration sampled at 32 Hz for 167 subjects. PPG signals were collected using a wrist-based wearable device in various conditions, including before cardioversion, during an exercise stress test, and in daily life. The dataset was restructured in [2] to ensure equal numbers of cases with and without AF across training, validation, calibration, and test sets, eliminating redundancy from the original data splits and providing a more reliable representation for model evaluation. It is important to note that the AF labels in DeepBeat were based on patient history rather than reflecting active AF within windows of data and the non-AF cohort were collected in an independent manner during exercise testing.

The *TriggersAF* dataset [101] includes data from patients diagnosed with paroxysmal AF recruited from the inpatient and outpatient wards of the Cardiology Department at Vilnius University Hospital. The dataset comprises 133 patients of whom 48 are female and 85 are male with an average age of 57.9 ± 11.6 years and a body mass index of 28.4 ± 4.9 kg/m$^2$. There is no case of permanent AF, with 45 patients experiencing AF and 88 patients experiencing no AF. The average duration of recordings is 6.9 ± 1.1 days, capturing a total of 307 episodes of AF with an average of 10 ± 19 episodes. ECG and PPG signals are sampled at 500 Hz and 100 Hz, respectively.

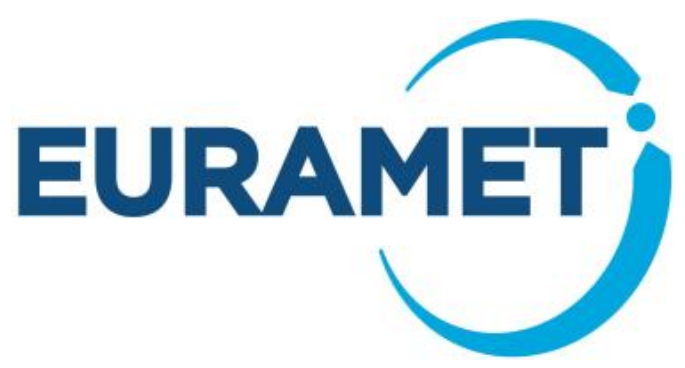

The *UMMC Simband* dataset [102][103] consists of 37 patients of whom 28 are male and 9 are female, aged between 50 and 91 years, and all diagnosed with cardiac arrhythmia. Patients wore the Samsung Simband 2 smartwatch, and their ECG was taken using a 7-lead Holter monitor. The data were pre-processed and segmented using non-overlapping 30 s windows. The ECG signals were sampled at 128 Hz and the PPG signals were down-sampled to 50 Hz. The PPG signals were labelled according to five categories: 0 (normal sinus rhythm), 1 (AF), 2 (premature atrial or ventricular contractions), 3 (uncertain between categories 0 and 2), and 5 (noisy signal). For this study, the PPG signals were labelled as AF present (category 1) or normal (categories 0, 2 and 3). Signals labelled as category 5 or without an ECG reference signal were excluded from the analysis.

*MIMIC-III-Ext-PPG* [104] is a new PPG-based benchmark dataset that has been created in the QUMPHY project. It is a large-scale dataset based on the *MIMIC III Waveform Database*, capturing over six million 30 s fingertip PPG waveform segments from more than 6,000 patients and sampled at 125 Hz. The patients were 44 % female and the metadata includes age, weight, height and ethnicity. Rhythm annotations were inferred from heart rhythm chart events in the corresponding *MIMIC-III* clinical database [105], which covers 26 different rhythm types, including AF and atrial flutter. It is particularly suited for training due to its size and diverse rhythm types. A potential second *MIMIC-III* based dataset [106] with manual annotations of AF was used for technical validation of the label quality of the *MIMIC-III-Ext-PPG* dataset and is not considered as a benchmark dataset. A limitation of the MIMIC data is that it is collected from intensive care patients who are receiving significant pharmacological and physiological treatment including cardiovascular medications, fluids that alter blood vessel volume (and hence could influence PPG signal), cardiac pacing and ventilation amongst others. Thus, models trained on this data should not be used to detect AF in healthy subjects.

## 5.3 Classification of Hypertension

The *AuroraBP* dataset [98] described in section 5.1 consists of PPG and BP signals collected for two separate cohorts of subjects distinguished by the technique used for BP measurement, namely auscultatory or oscillometric, and the presence or absence of ambulatory measurements. Based on those measurements of BP, the subjects are labelled according to the three classes of hypertension defined in section 1.3.3. Source code to perform this labelling is available from the software repository established in the QUMPHY project (see section 7).

## 5.4 Determination of Vascular Age

Two benchmark datasets are considered. The first is the *AuroraBP* dataset [98] described in section 5.1 in which subjects were separated into those considered ‘suitable’ (having no cardiovascular disease or hypertension and are not obese) and ‘unsuitable’ and then labelled by chronological age band (< 30 years, 30–40 years, 40–50 years, 50–60 years, 60–70 years, and > 70 years). The second is the *Pulse Wave Database* [107], which is a dataset of simulated waveforms comprising 4,374 healthy male subjects divided into five 10-year age groups between 25 years and 75 years of age. The dataset contains single-cycle PPG signals collected at various locations on the body.

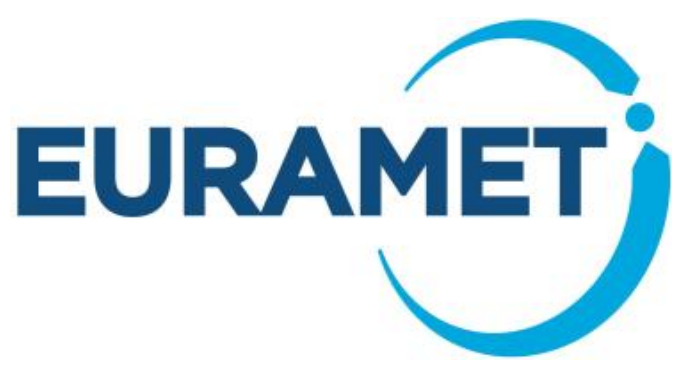

## 5.5 Detection of Sleep Apnoea

The *OSASUD* polysomnography dataset [108] involves 30 patients of whom 36.67 % are female admitted to the stroke unit at the Clinical Neurology Unit of Udine University Hospital between 2019 and 2020 due to suspected cerebrovascular events, such as ischemic and haemorrhagic stroke or transient ischemic attack. The dataset provides information about subject age (68.97 ± 11.22 years), body-mass index (29.17 ± 5.74 kg/m$^2$), apnoea-hypopnea index (25.43 ± 21.31 counts/h), and signal duration (8.91 ± 1.47 h). The dataset includes polysomnography signals such as finger-based PPG, ECG, oxygen saturation, respiratory rate, perfusion index, heart rate, nasal airflow, snoring, abdominal and thoracic movements. The sampling rate for the PPG and ECG signals is 80 Hz, and that for respiratory rate and oxygen saturation is 1 Hz. The *OSASUD* dataset was annotated by a trained sleep medicine physician based on the sleep scoring guidelines established by the American Academy of Sleep Medicine, identifying occurrences of central, obstructive, mixed apnoea, and hypopnea events.

The *MESA* dataset [109] comprises data from 2,055 patients, aged between 54 years and 95 years, of whom 53.63 % are female. It contains a total of 16,300 hours of fully annotated overnight polysomnography recordings collected during a sleep study between 2010 and 2012. The average age of subjects is 69.37 ± 9.12 years. Subjects included in the study had not used treatments for sleep apnoea, such as continuous positive airway pressure or oxygen devices, for more than a month prior to the study, or only used them less than once a week. The dataset represents a cohort featuring subjects from a variety of racial and ethnic groups, including White, African American, Hispanic, and Chinese American. The dataset includes signals from measurements of PPG, ECG, electrooculography, electroencephalography, electromyography, oxygen saturation, nasal airflow, snoring, and abdominal and thoracic movements. The sampling rate of the PPG and ECG signals is 256 Hz. Time labels and durations of apnoea and hypopnea episodes in the respiratory flow signals are annotated.

## 5.6 Determination of Respiratory Rate

The *MIMIC-III-Ext-PPG* dataset [104] (section 5.2) and the *OSASUD* dataset [108] (section 5.5) can both be used as benchmark datasets. For the former, measurements of respiratory rate are extracted for a subset of the patients to provide about 4.3 million PPG waveform segments for more than 4,600 patients of duration 30 s.

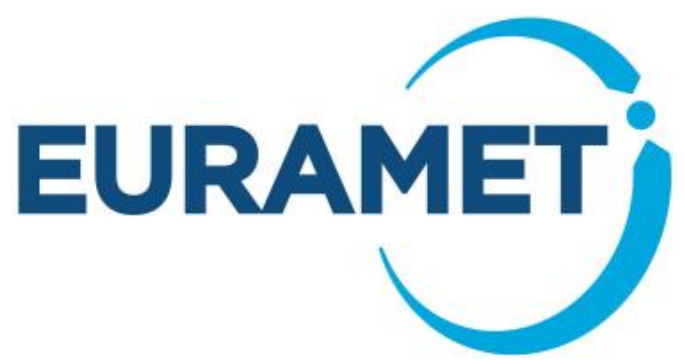

*MIMIC PERform Large* [110] is a newly created dataset, which is an extension of the *MIMIC Perform* datasets that are extracted from the *MIMIC III Waveform Database*. The dataset includes recordings of duration 32 s with reference respiratory rates.

# 6 Other Problems Studied

In addition to the six benchmark problems described above, we also studied two further classification problems, namely classification of skin tone and classification of biological sex. While neither of these is of direct clinical interest (and hence they were not included in the benchmark problems) an accurate classification indicates that the PPG signals are affected by these factors.

## 6.1 Classification of Skin Tone

As PPG signals are collected by shining light through the skin and measuring the light that is transmitted/reflected, the skin tone of the subject is likely to have a significant influence on the signals. This was confirmed during the Covid pandemic where it was observed that blood oxygenation measurements were inaccurate for patients with dark skin tones [111].

The labels used for this task were the Fitzpatrick skin tone [112] which is a six-point scale that was originally developed to classify people with white skin in order to select the correct UV dose for treating skin disease, not for classifying skin tone. However, it is now commonly used for this purpose. Another major limitation of these labels is that they are subjective and not based on any measurement, and so it is possible that the chosen class could easily differ by one from the "correct" class. Because of this, we introduced a *fuzzy accuracy* in which a predicted skin tone class is considered to be correct if it differs from the skin tone label by at most one [113].

The dataset used for this work was Aurora BP [98], which includes PPG signals together with Fitzpatrick skin tone labels for 823 participants. Three machine learning approaches were used for this classification problem, namely machine learning with interpretable features extracted from the PPG signals as input, machine learning or deep learning (using a one-dimensional CNN) with the raw signals as input, and deep learning using networks with pretrained weights which use SPAR images [43] as input. All three methods obtained modest accuracy, with the best result being 55 % using a Gradient Boosted Tree method with raw signals as input. However, much higher results were obtained using the fuzzy accuracy, with the best result being 96 %, again using Gradient Boosted Trees [113]. These results confirm that there is a clear dependence of PPG signals on skin tone.

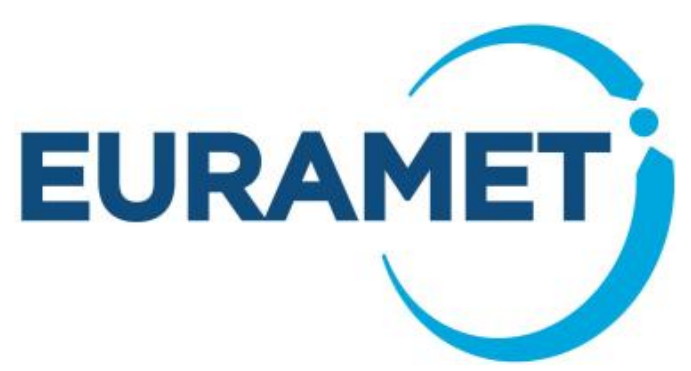

### 6.2 Classification of Biological Sex

High accuracy has been reported in the classification of biological sex using ECG signals [114], but similar work using PPG signals is limited. Clearly, in this case, there is no uncertainty in the labels for this binary classification task. We again used the Aurora BP dataset [98] and we used two approaches, namely deep learning using a one-dimensional CNN with the raw signals as input and deep learning using SPAR images [43] as input. The former method gave an accuracy of 71 % with slightly lower accuracy using SPAR images [4]. This is quite a bit lower than the classification using ECG signals, which suggests that the signature of biological sex is weaker in PPG signals than in ECG signals.

## 7 Software Repository

Software to train deep learning models (specifically, the *XResNet* and *AlexNet* models) for the AF detection and BP regression tasks, together with implementations of the UQ methods of MAP estimation, MCD, DEs, QR, conformal prediction, Venn-Abers, temperature scaling and isotonic regression can be found in [115]. This software repository also contains a notebook to apply a comprehensive validation framework for evaluating the reliability of reported uncertainties.

Implementations of the model dependent methods are found in the *train_eval_MCD_DE* directory of the repository, scripts for conformal prediction in the *conformal_prediction* directory, scripts for post-hoc calibration methods in the *post-hoc* directory, and the evaluation framework in the *uq_eval* directory. These scripts use custom versions of the *DeepBeat* and *VitalDB* benchmark datasets as described in the *README.md* file. Instructions for using the software are either apparent from the contents of the notebook or contained in the *README.md* files within the individual directories. Benchmark results obtained using the scripts are reported in [3].

The data pre-processing scripts for these experiments can be found in [116], together with other scripts for the training of models for the BP regression and AF classification tasks. Several different network architectures (without UQ) for various different input representations of the PPG signals (raw time series, image-based, and extracted features) are covered. Instructions for using the software are provided in a *README.md* file, and preliminary results obtained using the scripts are reported in [2].

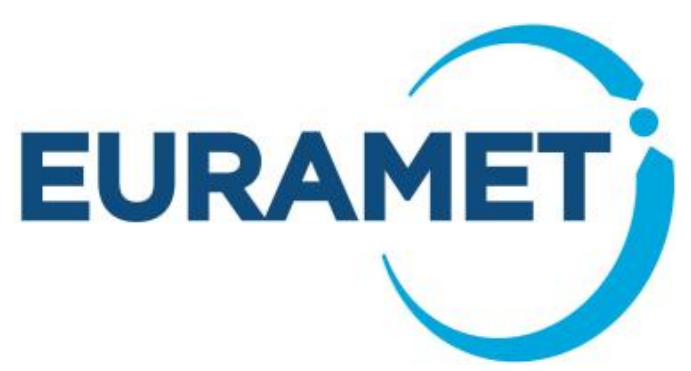

# 8 Ethics

## 8.1 Background

An ethics advisor was engaged to help the consortium for the QUMPHY project to highlight potential areas of risk, to propose ways that the outputs of the project can be delivered safely and equitably, and to adhere to existing guidelines and regulations where applicable, and with as wide a reach as possible. The ethics advisor played an important role in the project and prepared two reports during the lifetime of the project that captured their suggestions and recommendations. The ethics advisor led dedicated sessions with the consortium to discuss and address ethical issues relevant to the project and to ensure those suggestions and recommendations were integrated into the framework of the project. In this section, a summary based on extracts from those (unpublished) reports is presented.

## 8.2 Ethics Reports

While not a clinical project, the QUMPHY project can act as a flagship example of how to incorporate ethical considerations into the earliest stages of ML research applied to medical applications. The ethics reports prepared by the ethics advisor addressed the ethical considerations and implications of using UQ methods for ML in the analysis of PPG signals. The reports focussed on the following key areas:

- *Informed Consent*: Addressing ethical considerations related to obtaining informed consent from individuals whose PPG data is used for model development.
- *Regulatory Compliance*: Ensuring that the development and deployment of ML models for medical applications comply with relevant ethical guidelines and regulatory standards, such as those outlined by the European Commission (EC) as well as international bodies such as the World Health Organisation (WHO) and the International Telecommunication Union (ITU).
- *Bias and Fairness*: Evaluating the potential for bias in ML models, particularly concerning demographic subgroups such as sex, age, and skin tone, to ensure equitable performance and avoid discrimination.
- *Data Privacy and Protection*: Ensuring compliance with data protection regulations, such as the General Data Protection Regulation (GDPR) that applies in the UK and in the EU, and addressing the ethical handling of personal health data collected from PPG signals.
- *Transparency and Explainability*: Discussing the importance of transparency in ML models, including the interpretability of model predictions and the communication of uncertainty to end users, such as clinicians and patients.

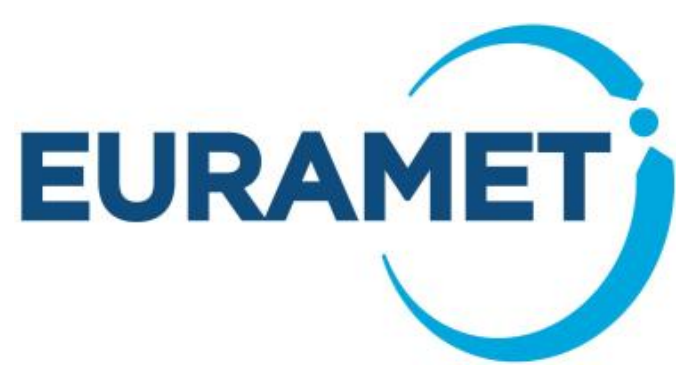

## 8.3 Ethics Self-Assessment

The outcome of a self-assessment by the consortium for the QUMPHY project undertaken in collaboration with the ethics advisor yielded the following conclusions.

### 8.3.1 Involvement of Humans, Human Cells or Tissue

In this project, no humans, human cells, or tissues are directly involved in any experimental or data collection processes. The project focuses on the development and validation of ML models using existing datasets of PPG signals. These datasets are either publicly available or synthetically generated, ensuring that no new human data is collected specifically for the project. The analysis and methodologies employed are designed to work with pre-existing data, thereby eliminating the need for direct human participation or the use of human biological materials. This approach ensures that all research activities are conducted in compliance with ethical standards and regulations, prioritising the use of non-invasive and pre-collected data to achieve the project's objectives.

### 8.3.2 Personal Data

In this project, the use of personal data is carefully managed to ensure compliance with ethical standards and data protection regulations. All data used in the project are either generated through computer simulations or derived from publicly available datasets that already conform to EU data protection guidelines. This approach ensures that no new personal data is collected directly from individuals, thereby minimising privacy concerns. For datasets used within the project that have access restrictions, these are handled with the utmost care and will only be made available to the public in strict accordance with the data access guidelines of the respective dataset. This meticulous handling of data underscores the project's commitment to maintaining high ethical standards and protecting individual privacy.

### 8.3.3 Animals

In this project, no animal is involved in any experimental or data collection processes. The project's focus is solely on the development and validation of ML models using existing datasets of PPG signals collected from consenting humans, which are either publicly available, available on request, or synthetically generated. The only exception to this involves the potential presence of emotional support animals, which may accompany individual consortium members in certain settings to provide comfort and relieve stress. However, these animals are not subjects of the research and are not involved in any data collection or experimental procedures. This approach ensures that the project adheres to ethical guidelines and regulations concerning animal welfare.

### 8.3.4 Non-EU Countries

This project does not use any facilities, materials, or experimental designs from non-EU countries. While some datasets employed in the project originate from non-EU countries, the project has not been involved in the creation of those datasets. Excepting the *MIMIC-III-Ext-PPG* benchmark dataset which has been compiled in the QUMPHY project from the established MIMIC-III dataset

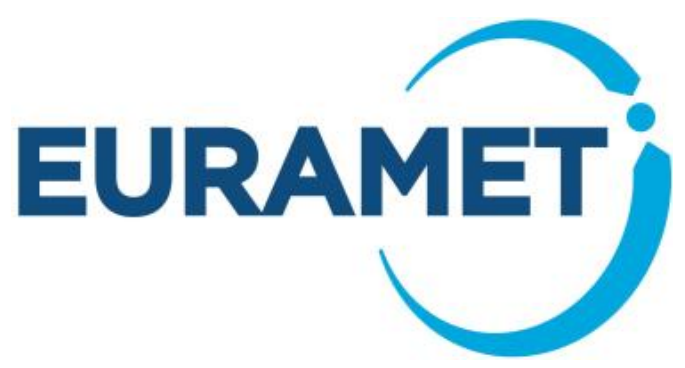

[105] (section 5.2), all datasets in use have been published prior to the project, are frequently used in European research initiatives and, as far as the consortium is aware, have been collected with similar ethical and humane standards as those stated by the EU and conform to EU data protection guidelines.

### 8.3.5 Environment, Health and Safety

In this project, there is no risk of environmental damage and, consequently, no specific precaution is necessary to address such concerns. The experimental design and structure of the project have been carefully planned to ensure that they do not meet any criteria that could potentially harm the health or safety of the individuals involved. Additionally, the technologies employed in the project have been evaluated to confirm that they do not pose any undesirable side-effects that could endanger any of the persons participating in the project.

### 8.3.6 Artificial Intelligence

*Human Agency and Oversight*: All ML methods developed and investigated in the project are critically evaluated for their performance. The project's focus on UQ for ML specifically emphasises human agency and oversight, ensuring that humans can interpret the reliability of the outputs in conjunction with the associated uncertainties. The project is centred on establishing general procedures for UQ and benchmarks ML models using specific, openly available, and anonymised datasets. Consequently, the results are not intended to inform decisions for direct clinical practice. The limitations of the employed models and the UQ methods are discussed and highlighted within the software and publications produced by the project. No software or result of the project attempts to exaggerate confidence or coerce, deceive, or manipulate individuals, as such actions are fundamentally opposed to the project's goals.

*Privacy and Data Governance*: Privacy and data governance are handled with the utmost care and adherence to international, EU, and national laws regarding anonymity, ethical treatment of subjects, and data quality. All datasets used in the project are openly available or available on request and comply with EU standards. For datasets with restricted access, these are clearly noted, and such datasets will not be published without the explicit consent of the original data holders. To ensure transparency in data processing and augmentation for ML purposes, scripts used to process restricted access datasets are made publicly available (see section 7). This approach allows for the reproduction of the project's results while maintaining data privacy. The selection of data subsets for training purposes is conducted internally to prevent or reduce inherent bias against minority groups. Network architectures, training procedures, and datasets are made publicly available to ensure reproducibility of the results. Furthermore, results are run multiple times to provide robust statistics for training procedures, ensuring the reliability and robustness of model outputs.

*Transparency*: Transparency is a cornerstone of the research methodology and dissemination strategy. All datasets used in the project are openly available or available on request, ensuring that the broader research community can access and verify the data used. For datasets with restricted access, these are clearly noted, and such datasets will not be published without the explicit consent of the original data holders. To further ensure transparency, scripts used to process restricted access datasets are made publicly available, allowing for the reproduction of the project's results and providing insight into data processing and augmentation for ML purposes. Network architectures, training procedures, and datasets are made publicly available to ensure reproducibility of the results. Results are run multiple times to provide robust statistics for training procedures, ensuring the reliability and robustness of model outputs. All software developed in the project is tracked in an online repository with version control, allowing for traceability of the development process. Automatically generated online documentation is provided to simplify access to the software library. Scripts to reproduce any published results, including network architectures, training routines, and used hardware, are also provided (see section 7).

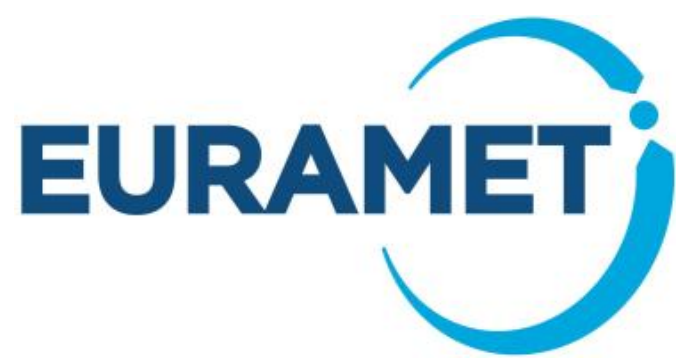

*Fairness, Diversity and Non-Discrimination*: All datasets used in the project adhere to the FAIR principles, ensuring that they are Findable, Accessible, Interoperable, and Reusable [117]. A significant focus of the project is on investigating potential biases against demographic subgroups within these datasets, with specific activities aimed at examining biases related to sex, age, and skin tone. Considerable thought is given to the limitations of the produced results with respect to their application to demographic subgroups, and these concerns are discussed with the ethics advisor to ensure a comprehensive understanding of potential biases and limitations. The outputs of the project include discussions on the limitations of the data when applied to certain minority groups, highlighting areas where biases may exist or where data may not be fully representative. All ML applications developed in the project are trained on a variety of data to reduce bias within the constraints of the given datasets. Additionally, the UQ methods employed are investigated for their potential to reflect biases against minority groups, ensuring that the project's findings are both robust and ethically sound.

*Societal and Environmental Well-Being*: From a societal perspective, the project focuses on improving the reliability and trustworthiness of ML models used in medical diagnostics through the analysis of PPG signals. By enhancing the accuracy and UQ of these models, the project aims to improve healthcare outcomes, for example in the early detection and management of conditions such as hypertension. This focus can lead to better health management and societal well-being. Environmentally, the project does not involve any activities that could harm the environment. The research is conducted using existing datasets and computational models, which do not require physical experimentation or the use of hazardous materials. The project's reliance on data analysis and ML models ensures that there is no direct environmental impact, such as pollution or resource depletion. Regarding energy use, the project trained machine and deep learning models on local computers which have a negligible environmental impact. Thus, the project aligns with principles of sustainability and ethical research, ensuring that both societal and environmental well-being are preserved and potentially enhanced.

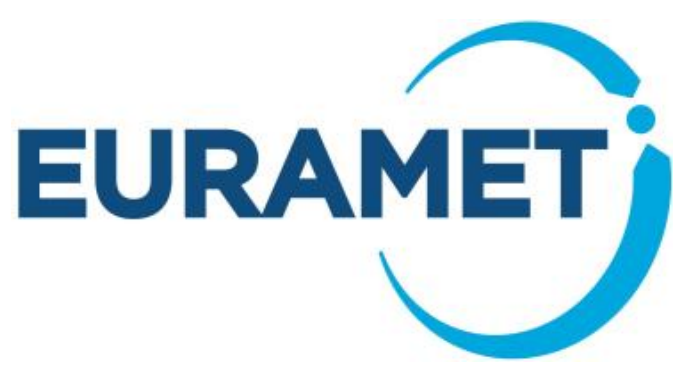

*Accountability*: Accountability is realised through a structured approach that ensures responsibility, transparency, and oversight throughout the development and operation of ML models. The project involves multiple stakeholders, including developers, researchers, and clinicians, who are collectively responsible for the functionality and outcomes of the ML systems developed. This accountability is supported by the project's commitment to transparency, as evidenced by the open availability of datasets, network architectures, training procedures, and software repositories. These resources are made publicly accessible to ensure that the processes and methodologies used can be scrutinised and validated by external parties. Furthermore, the project incorporates rigorous oversight mechanisms, including regular consultations with an ethics advisor and stakeholder committee, to ensure that the ML applications are developed and operated in an ethically sound manner. The project's focus on UQ also enhances accountability by providing a clear understanding of the reliability and limitations of the ML models' predictions. This comprehensive framework will help developers and operators of the ML software applications explain how and why the systems exhibit particular characteristics or result in certain outcomes, thereby fulfilling the criteria for accountability.

## 8.4 Potential Misuse of Results

In this project, potential misuse of the results and ML models in general is carefully considered and mitigated through several integrated measures. To prevent misuse, the project emphasises transparency and accountability by making all datasets, network architectures, training procedures, and software repositories openly available (see section 7). Furthermore, the project ensures transparency in the scope, application domain and limitations of the benchmark datasets and code frameworks made publicly available. This transparency ensures that the processes and methodologies can be scrutinised and validated by external parties, reducing the risk of unethical use. It also ensures the outputs will not be used for purposes for which they are not validated, which could cause unintended harm. Additionally, the project incorporates rigorous oversight mechanisms, including regular consultations with an ethics advisor and stakeholder committee, to ensure that the ML models are developed and operated ethically. The focus on UQ further enhances the reliability and trustworthiness of the ML models' predictions, providing a clear understanding of their limitations and potential biases. By discussing the limitations of the data and models, particularly concerning minority groups, the project aims to prevent any discriminatory or harmful applications of the results. Moreover, the project's commitment to open science practices, such as publishing results in open-access journals and presenting findings at international conferences, ensures that the broader research community can access and build upon the project's outcomes responsibly. These comprehensive measures underscore the project's dedication to ethical research practices and the responsible use of ML in medical applications.

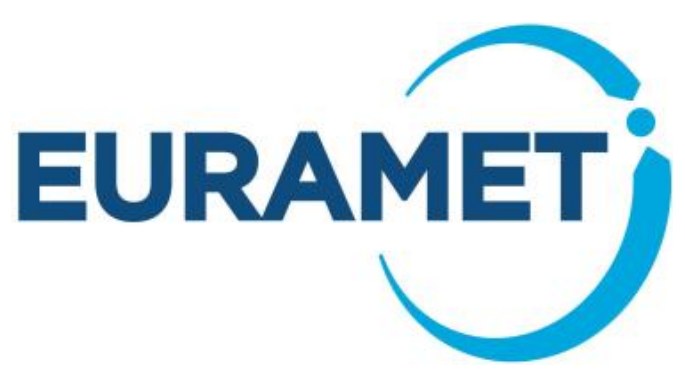

# 9 Summary

## 9.1 Background

Machine learning is increasingly being applied using PPG signal inputs in order to estimate physiological parameters in regression tasks or to detect diseases associated with the cardiac system in classification tasks. Three approaches to machine learning are described in section 2: deep learning using the raw PPG signal as input, deep learning using the PPG signal converted into an image representation, and machine learning using features extracted from the PPG signal.

Whichever approach to analysing a PPG signal is adopted, a model output in the form of a prediction is incomplete without a quantitative statement of the quality of the prediction in terms of an uncertainty. The uncertainty is needed to establish whether the prediction can be used to reliably inform diagnosis and as the basis of decision-making. Methods for the quantification of the uncertainty are described in section 3, and summarised in Table 1 in terms of their type, whether they apply to regression and/or classification tasks, whether they model epistemic and/or aleatoric uncertainty, and how the uncertainty is reported.

But it is insufficient to simply provide a quantification of the uncertainty, rather it is essential that the uncertainty is reliable and conveys meaningful information about the prediction. It follows that the practical realisation of the benefits that wearable PPG-sensing technologies could provide to patients hinges on the development of robust uncertainty quantification methods and a framework for the validation of uncertainty. Metrics that underpin that framework are described in section 4 and in Table 2 and Table 3 for, respectively, the tasks of classification and regression.

To apply the framework for the validation of uncertainty requires benchmark problems and associated benchmark datasets that are tailored to address the challenges of ensuring the accuracy and reliability of model results and are intended for the medical device and digital health communities to facilitate model evaluation. Such benchmarking datasets are described in section 5 for benchmark problems covering both regression and classification tasks. To apply the framework also depends on software, and software to assist practitioners in implementing the guidance of this GPG is indicated in section 7.

There are ethical considerations and implications of using methods of uncertainty quantification for machine learning in the analysis of PPG signals. Such considerations and implications are described in section 8, covering the issues of informed consent, regulatory compliance, bias and fairness, data privacy and protection, and transparency and explainability.

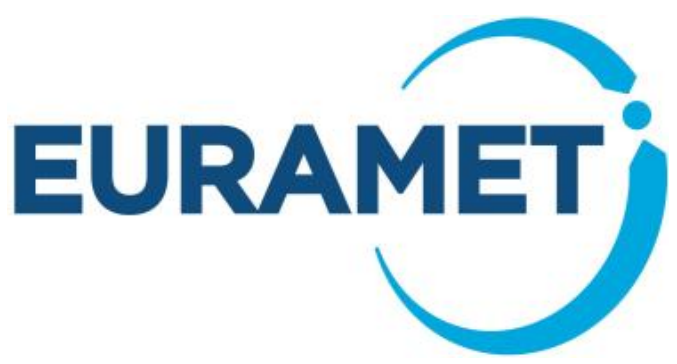

Table 1: Summary of uncertainty quantification methods. Note that frequentist techniques do not explicitly model different sources of uncertainty (and are marked '*'). Post-hoc techniques aim to improve the reliability of informative, but poorly calibrated, uncertainties provided by a model. Therefore, the sources of uncertainty encoded in the outputs depend on those captured by the base model and UQ technique. MAP and QR have output types that straightforwardly express uncertainty, and while we express this uncertainty here, this is not always implemented in practical use cases. ('R' denotes regression tasks and 'C' denotes classification tasks; 'E' denotes epistemic uncertainty and 'A' denotes aleatoric uncertainty.) (Adapted from [3].)

| UQ Method | Type | R/C | E/A | Output Type (R) | Output Type (C) |
|---|---|---|---|---|---|
| **MLE estimation** | Model-dependent | ✓/✓ | ✗/✓ | Mean and variance | Class probabilities |
| **MAP estimation** | Model-dependent | ✓/✓ | ✗/✓ | Mean and variance | Class probabilities |
| **Deep Ensembles** | Model-dependent | ✓/✓ | ✓/✓ | Mean and variance | Class probabilities and logit variances |
| **Monte Carlo Dropout** | Model-dependent | ✓/✓ | ✓/✓ | Mean and variance | Class probabilities and logit variances |
| **Quantile Regression** | Model-dependent | ✓/✗ | * | Quantiles | ✗ |
| **Histogram Binning and Isotonic Regression** | Model independent calibration | ✓/✓ | * | Same as 'uncalibrated' uncertainties | Base model output type |
| **Temperature Scaling** | Model independent calibration | ✓/✓ | * | Same as 'uncalibrated' uncertainties | Base model output type |
| **Conformal Prediction** | Model-independent calibration | ✓/✓ | * | Prediction intervals | Prediction sets |

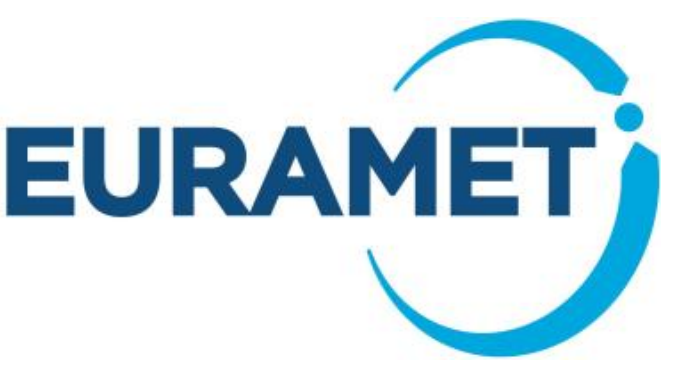

| Venn-Abers | Model-independent calibration | ✗/✓ | * | ✗ | Probability intervals |
|---|---|---|---|---|---|

Table 2: Summary of metrics for evaluating the reliability of uncertainty for classification tasks. ('L' denotes local metrics and 'G' denotes global metrics.) (Adapted from [3].)

| Metric | Expression | L/G | Visualisation | Proper Scoring Rule |
|---|---|---|---|---|
| **ECE** | Predicted class probability | L | Average confidence in bin vs. average accuracy | ✗ |
| **ACE** | Predicted class probability | L | Average confidence in bin vs. accuracy, with adaptive binning | ✗ |
| **smECE** | Predicted class probability | L | Kernel smoothed confidence vs. accuracy | ✗ |
| **UCE** | Entropy of predicted prob-ability distribution | L | Mean predicted uncertainty (expressed as entropy) vs. mean observed error in each uncertainty bin | ✗ |
| **VCE** | Compatible with several expressions of variation (here entropy is considered) | L | Mean predicted uncertainty (expressed as entropy) vs. mean observed variation (also expressed as an entropy) in each uncertainty bin | ✗ |
| **NLL** | Predicted class distribution | G | N/A | ✓ |

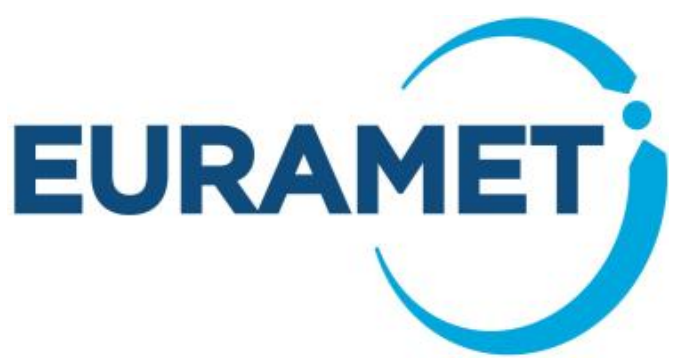

Table 3: Summary of metrics for evaluating the reliability of uncertainty for regression tasks. ('L' denotes local metrics and 'G' denotes global metrics; 'V' denotes variance-based metrics and 'C' denotes coverage-based metrics.) (Adapted from [3].)

| Metric | Input | L/G | Visualisation | Proper Scoring Rule | V/C |
|---|---|---|---|---|---|
| **ENCE** | Predicted variances | L | Root mean square of the prediction errors in a bin vs. the root mean of the corresponding predicted variances | ✗ | V |
| **PICP** | Prediction intervals or quantiles | G | N/A | ✗ | C |
| **CCE** | Quantiles | | Coverage level vs. observed coverage for various coverage levels | ✗ | C |
| **CRPS** | Predicted distribution functions | G | N/A | ✓ | C |
| **MPIW** | Prediction intervals | G | N/A | ✗ | C |

The study reported in [2] describes the benchmarking of raw signal, image-based and feature-based machine learning models applied in the analysis of PPG signals, and that in [3] describes a systematic evaluation of uncertainty quantification techniques using the example of PPG signal analysis. Both studies focus on particular regression tasks (the estimation of systolic and diastolic blood pressure) and a classification task (the detection of atrial fibrillation). Specific guidance from those studies is reproduced below, together with more general recommendations.

## 9.2 Benchmarking of Machine Learning Models

For the regression tasks, the best results were achieved using models taking raw time series as input. Feature-based approaches were not competitive in the so-called 'calibration' scenario (in which train and test datasets share subjects and models can profit from memorisation) but are, to a certain degree, competitive in the so-called ‘calibration-free’ scenario (in which train and test datasets do not share subjects). As was expected, models achieved much better performance in the ‘calibration’ setting (even though test sets were not entirely comparable). Large and complex models (such as *XResNet1d* and *Inception1d*) performed best in the ‘calibration’ setting, as overfitting to specific patients is desirable, whereas smaller models (such as *LeNet1d*) performed on par or even better than more complex models in the ‘calibration-free’ setting, where generalisation to unseen patients is key. The gap between the performance of raw time series and feature-based approaches is larger in the ‘calibration’ setting, which is quite natural as that scenario benefits from using complex models, and feature-based approaches are less complex than typical models operating on raw time series.

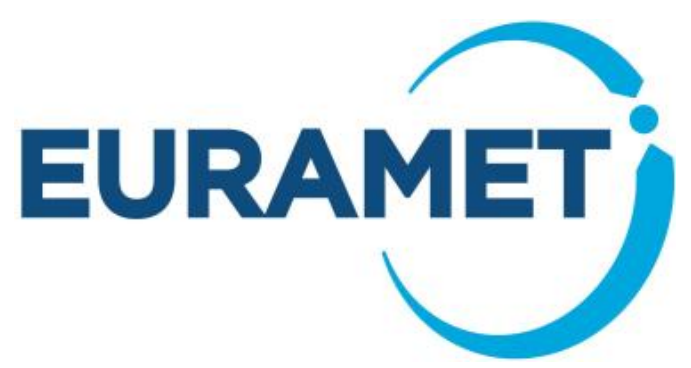

Best model performance for the classification task was achieved using modern CNN architectures (such as *XResNet1d* and *Inception1d*), which performed at least on par with, and in some cases showed advantages over, simpler models (such as *LeNet1d*). Feature-based approaches based on wavelets showed comparable performance to simple CNNs but were outperformed by large-scale CNNs operating on raw time series input. The difference in performance between raw time series models and those based on clinically interpretable features may be influenced by multiple factors, including biases in the data acquisition process, the presence of artefacts, and errors in the labelling of training samples into different classes. Deep learning-based raw time series models can use a broad range of information from the signal and automatically extract the most useful features, enhancing their ability to detect class-specific patterns. However, the drawback is that clinically *irrelevant* features, such as artefacts and biases, may be incorporated into the training process. In contrast, models that rely on predefined clinically interpretable features, such as pulse irregularity, have a more constrained feature space. While this improves interpretability and robustness in some cases, it may also make these models more vulnerable to mislabelled or noisy data. Initial experiments indicated that incorporating signal quality assessments into interpretable features can enhance prediction performance. Future work is needed to investigate the classification performance in terms of more fine-grained subgroups, in particular stratified according to signal and label quality. Such analysis could reveal if feature-based methods exhibit advantages over raw time series approaches for high-quality subsets, where the underlying assumptions of clinical validity are most clearly satisfied.

In summary, in the study [2], modern CNN architectures such as *XResNet1d* provided the best performance for the regression and classification tasks considered. A general recommendation is to use these modern CNN architectures for prediction models operating on PPG data. However, in certain cases, it may not be necessary to use such complex models (e.g., for the ‘calibration-free’ scenario and blood pressure estimation). Similarly, for certain tasks, models leveraging image representations achieved competitive results. Nevertheless, in the interest of achieving competitive results on an unknown task, the use of modern CNN architectures based on raw time series representations remains the safest choice.

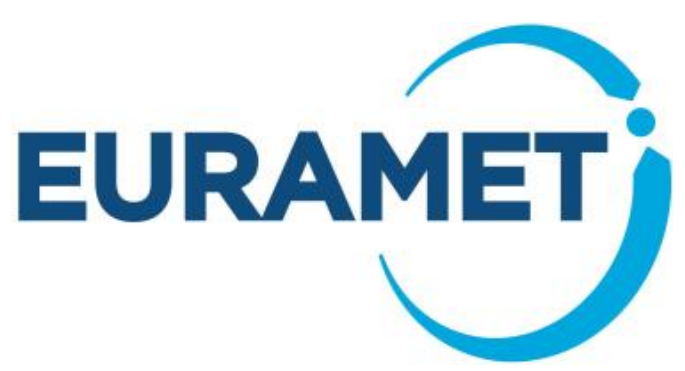

## 9.3 Benchmarking of Uncertainty Quantification Methods

The results for the classification task indicated that the considered uncertainty quantification techniques do not provide adaptive uncertainty estimates (i.e., reliable uncertainties assessed per class) but instead prioritise global reliability with a bias towards the dominant class. In particular, while model-independent calibration methods for uncertainty quantification resulted in high global reliability for the class probabilities, the corresponding adaptive per-class metrics indicated poor reliability, which is not surprising as the predictions were not calibrated to achieve adaptive reliability. This emphasises the need to use evaluation metrics that cater to the practical use of the model. Indeed, individual reliability is the most effective measure of assessing uncertainty reliability, as this reflects how the model is likely to be used in practice. However, there is a lack of robust quantitative metrics to assess individual reliability, and so there is a need for new methods to be developed. It also points to a general need for uncertainty quantification techniques that are sensitive to class imbalance, and which result in high adaptivity across classes.

In general, uncertainty reliability varied considerably across the different uncertainty quantification techniques depending on the scale (local or global) at which uncertainty reliability was assessed. The qualitative assessment of uncertainty reliability at smaller scales reported in [3] using binning-based reliability diagrams revealed these biases in uncertainty reliability, emphasising the practical utility of the approach. These diagrams also revealed local trends in whether uncertainties were over- or under-estimated, which may aid in developing strategies for using estimates to inform diagnosis.

Similar conclusions were drawn for the regression tasks where the best uncertainty quantification technique depended both on the chosen metric used to assess its reliability and the scale at which uncertainty reliability was assessed. The use of bivariate histograms reported in [3] to qualitatively assess small scale reliability revealed poor reliability for all models and uncertainty quantification techniques, where the worst reliability was exhibited in the ‘calibration-free’ setting. The reliability diagrams based on the ENCE metric helped to determine whether there were local trends of over- or under-confidence. The model-independent calibration methods for uncertainty quantification improved uncertainty reliability in the ‘calibration’ setting for some models and some metrics (mostly for quantile regression and the *ResNet* model). In the ‘calibration-free’ setting, model-independent calibration methods for UQ did not generally improve uncertainty reliability according to the chosen metrics. There are challenges associated with converting between different output types that need to be resolved to allow more meaningful comparative studies across different uncertainty quantification techniques. Note that the nature of the ‘calibration-free’ setting may violate assumptions underpinning some of the uncertainty quantification techniques considered, and it is important to be wary of whether the task and data is compatible with a given uncertainty quantification technique.

In summary, the assessment of the reliability of reported uncertainties is highly dependent upon the chosen expression of uncertainty and evaluation metric. Furthermore, any trade-off between uncertainty reliability and predictive performance should be considered when deciding on a preferred uncertainty quantification technique. A general recommendation from the study [3] for applying uncertainty quantification to deep learning models is to identify an uncertainty quantification technique that delivers the desired expression of uncertainty for the given task and then evaluate adaptive and local versions of relevant uncertainty reliability metrics for a range of model parametrisations. The uncertainty quantification technique and model parametrisation that produces the best trade-off between adaptive and/or small-scale reliability and predictive performance is selected. Reliability diagrams should be used to observe whether models exhibit local trends of over- or under-confident estimates, and this can also be used as a factor for choosing a good uncertainty quantification technique if deemed important for the practical use case of the model. Modelling both aleatoric and epistemic uncertainty appears to provide better overall uncertainty reliability, and so methods that consider both types of uncertainty are generally preferable. Future work might consider ways to develop confidence thresholds for keeping or discarding an estimate, uncertainty quantification techniques that encourage adaptivity, and quantitative individual reliability metrics.

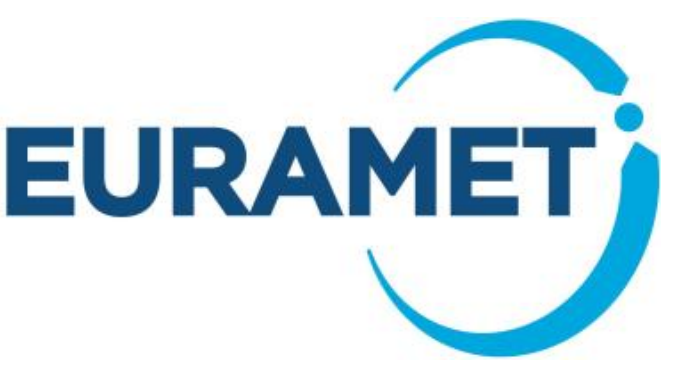

## 10 Acknowledgements

The QUMPHY consortium is grateful to Dr Peter Harris (NPL) who wrote the majority of this Good Practice Guide.

The project (22HLT01 QUMPHY) has received funding from the European Partnership on Metrology, co-financed from the European Union's Horizon Europe Research and Innovation Programme and by the Participating States. Funding for the University of Cambridge, KCL, NPL and the University of Surrey was provided by Innovate UK under the Horizon Europe Guarantee Extension, grant numbers 10091955, 10087011, 10084125, and 10084961 respectively. Peter Charlton acknowledges funding from the British Heart Foundation (BHF) grant [FS/20/20/34626].

The QUMPHY consortium acknowledges the important contribution made by the project's external ethics advisor: Dr. Jenny Venton, The Royal Surrey NHS Foundation Trust, Egerton Road, Guildford, Surrey, GU2 7XX, UK.

## References

[1] Charlton, P.H.; Kyriacou, P.; Mant, J.; Alastruey, J. Acquiring wearable photoplethysmography data in daily life: The PPG diary pilot study. *Eng. Proc.* **2**(1):80, 2020.

[2] Moulaeifard, M.; Coquelin, L.; Rinkevičius, M. et al. Machine-learning for photoplethysmography analysis: Benchmarking feature, image, and signal-based approaches. *Biomed. Sig. Proc. Ctrl.* **120**:109831, 2026.

[3] Bench, C.; Pfeffer, O.; Desai, V. et al. A systematic evaluation of uncertainty quantification techniques in deep learning: a case study in photoplethysmography signal analysis. *Mach. Learn. Hlth.* **2**(1): 015011, 2026.

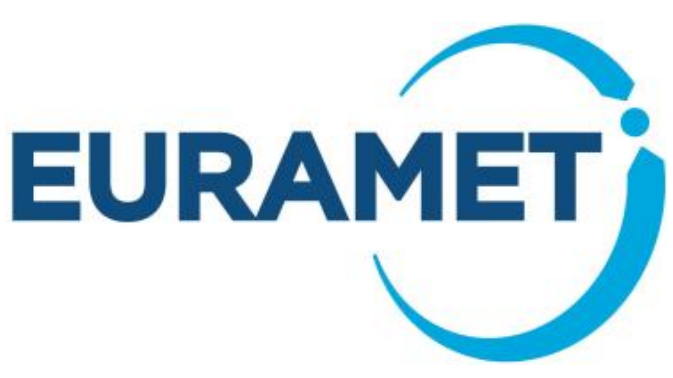

[4] QUMPHY D3 report, 2026. https://www.qumphy.ptb.de/publications

[5] Hackstein, U.; Alastruey, J.; Aston, P.J. et al. Benchmark problems and benchmark datasets for the evaluation of machine and deep learning methods on photoplethysmography signals: The D4 report from the QUMPHY project. arXiv 2604.01398, 2026.

[6] Paliakaite, B.; Petrenas, A.; Solosenko, A.; Marozas, V. Modeling of artifacts in the wrist photoplethysmogram: Application to the detection of life-threatening arrhythmias. *Biomed. Sig. Proc. Ctrl.* **66**:102421, 2021.

[7] McEvoy, J.W.; McCarthy, C.P.; Bruno, R.M. et al. ESC Scientific Document Group, 2024 ESC Guidelines for the management of elevated blood pressure and hypertension: Developed by the task force on the management of elevated blood pressure and hypertension of the European Society of Cardiology (ESC) and endorsed by the European Society of Endocrinology (ESE) and the European Stroke Organisation (ESO). *Europ. Heart J.*, **45**(38):3912–4018, 2024.

[8] Krijthe, B.P.; Kunst, A.; Benjamin, E.J. et al. Projections on the number of individuals with atrial fibrillation in the European Union, from 2000 to 2060. *Eur. Heart J.* **34**(35):2746-2751, 2013.

[9] Van Gelder, I.C.; Rienstra, M.; Bunting, K.V. et al. 24 ESC Guidelines for the management of atrial fibrillation developed in collaboration with the European Association for Cardio-Thoracic Surgery (EACTS). *Eur. Heart J.* **45**(36):3314-3414, 2024.

[10] Solosenko, A.; Petrenas, A.; Paliakaite, B.; Sornmo, L.; Marozas, V. Detection of atrial fibrillation using a wrist-worn device. *Phys. Meas.* **40**(2):025003, 2019.

[11] Petrenas, A.; Marozas, V.; Sornmo, L. Low-complexity detection of atrial fibrillation in continuous long-term monitoring. *Comp. Biol. Med.* **65**:184-191, 2015.

[12] Jones, N.R.; Taylor, C.J.; Hobbs, F.R.; Bowman, L.; Casadei, B. Screening for atrial fibrillation: A call for evidence. *Eur. Heart J.* **41**(10):1075-1085, 2020.

[13] Climie, R.E.; Alastruey, J.; Mayer, C.C. et al. Vascular ageing: moving from bench towards bedside. *Eur. J. Prev. Cardiol.* **30**(11):1101-1117, 2023.

[14] Reshetnik, A.; Gohlisch, C.; Tolle, M.; Zidek, W.; van der Giet, M. Oscillometric assessment of arterial stiffness in everyday clinical practice. *Hyperten. Res.* **40**(2):140-145, 2017.

[15] Hamczyk, M.R.; Nevado, R.M.; Barettino, A.; Fuster, V.; Andres, V. Biological versus chronological aging: JACC focus seminar. *J. Amer. Coll. Cardiol.* **75**(8):919-930, 2020.

[16] Abbasi, A.; Gupta, S.S.; Sabharwal, N. et al. A comprehensive review of obstructive sleep apnea. *Sleep Science* **14**(2):142-154, 2021.

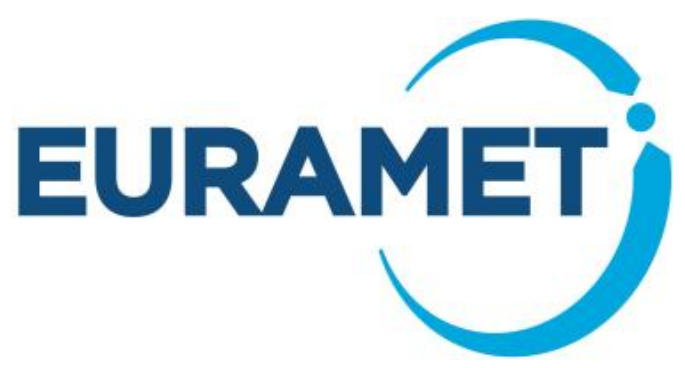

[17] Kapur, V.; Auckley, D.; Chowdhuri, S. et al. Clinical practice guideline for diagnostic testing for adult obstructive sleep apnea: An American Academy of Sleep Medicine Clinical Practice Guideline. *J. Cl. Sleep Med.* **13**:479-504, 2017.

[18] Yeghiazarians, Y.; Jneid, H.; Tietjens, J.R. et al. Obstructive sleep apnea and cardiovascular disease: A scientific statement from the American Heart Association. *Circ*. **144**(3):e56-e67, 2021.

[19] Bubu, O.M.; Andrade, A.G.; Umasabor-Bubu, O.Q. et al. Obstructive sleep apnea, cognition and Alzheimer's disease: A systematic review integrating three decades of multidisciplinary research. *Sleep Med. Rev.* **50**:101250, 2020.

[20] Smolensky, M.H.; Milia, L.D.; Ohayon, M.M.; Philip, P. Sleep disorders, medical conditions, and road accident risk. *Acc. Anal. Prev.* **43**(2):533-548, 2011.

[21] zur Nieden, P.B. The economic burden of (obstructive) sleep apnea. Costs and implications for Germany based on the results of an international systematic review. *J. Pub. Health.* **34**:409-427, 2024.

[22] Markun, L.C.; Sampat, A. Clinician-focused overview and developments in polysomnography. *Curr. Sleep. Med. Rep.* **6**:309-321, 2020.

[23] Rosen, I.; Kirsch, D.; Chervin, R. Clinical use of a home sleep apnea test: An American Academy of Sleep Medicine position statement. *J. Clin. Sleep. Med.* **13**:1205-1207, 2017.

[24] Costa, J.; Rebelo-Marques, A.; Machado, J. STOP-Bang and NoSAS questionnaires as a screening tool for OSA: Which one is the best choice? *Rev. Assoc. Med. Bras.* **66**(9):1203-1209, 2020.

[25] Evans, K.A.; Yap, T.; Turner, B. Screening commercial vehicle drivers for obstructive sleep apnea: Tools, barriers, and recommendations. *Work. H&S.* **65**(1):487-492, 2017.

[26] Braum, S.R. Respiratory rate and pattern. In *Clinical Methods: The History, Physical, and Laboratory Examinations*. Eds H. Walker, W. Hall, and J. Hurst. Third edition. Boston: Butterworths, Chapter 43, 1990.

[27] Charlton, P.H.; Bonnici, T.; Tarassenko, L. et al. An assessment of algorithms to estimate respiratory rate from the electrocardiogram and photoplethysmogram. *Phys. Meas.* **37**(4):610, 2016.

[28] Schein, R.M.H.; Hazday, N.; Pena, M.; Ruben, B.H.; Sprung, C.L. Clinical antecedents to in-hospital cardiopulmonary arrest. *Chest* **98**(6):1388–1392, 1990.

[29] Goldhill, D.; White, S.; Sumner, A. Physiological values and procedures in the 24 h before ICU admission from the ward. *Anaesthesia* **54**(6):529–534, 1999.

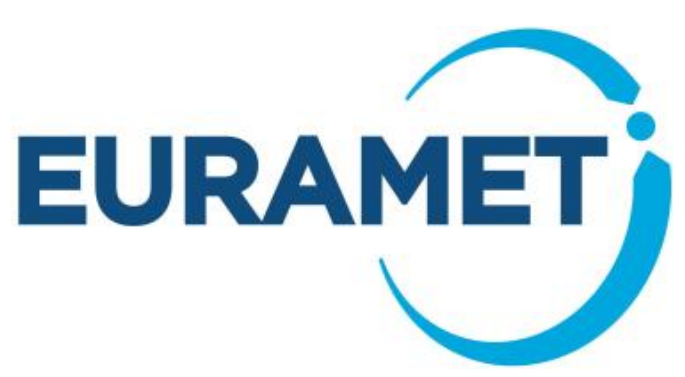

[30] Cretikos, M.; Bellomo, R.; Hillman, K. et al. A. Respiratory rate: The neglected vital sign. *Med. J. Aust.* **188**(11):657–659, 2008.

[31] Ridley, S. The recognition and early management of critical illness. Ann. R. Coll. Surg. Engl. **87**(5):315–322, 2005.

[32] Bailon, R.; Sornmo, L.; Laguna, P. ECG-derived respiratory frequency estimation. In *Advanced Methods and Tools for ECG Data Analysis*. London: Artech Houes, Chapter 8, 215–44, 2006.

[33] Meredith, D.; Clifton, D.; Charlton, P.H. et al. Photoplethysmographic derivation of respiratory rate: A review of relevant physiology. *J. Med. Eng. Technol*. **36**(1):1–7, 2012.

[34] Aston, P.J.; Adel, T.; Bench, C. et al. Machine learning tools in metrology. In *Mathematical and Computational Approaches in Measurement*, in preparation.

[35] Wagner, P.; Mehari, T.; Haverkamp, W.; Strodthoff, N. Explaining deep learning for ECG analysis: Building blocks for auditing and knowledge discovery. *Comput. Biol. Med.* **176**:108525, 2024.

[36] Fawaz, H.I.; Lucas, B.; Forestier, G. et al. Inceptiontime: Finding Alexnet for time series classification. *Data Min. Know. Disc.* **34**(6), 1936–62, 2020.

[37] Strodthoff, N.; Wagner, P.; Schaeffter, T.; Samek, W. Deep learning for ECG analysis: Benchmarks and insights from PTB-XL. *IEEE J. Biomed. Health Inform.* **25**(5):1519–1528, 2020.

[38] Krizhevsky, A.; Sutskever, I.; Hinton, G.E. Imagenet classification with deep convolutional neural networks. *Adv. Neural Inf. Process Syst.* **25**, 2012.

[39] Dempster, A.; Schmidt, D.F.; Webb, G.I. Minirocket: A very fast (almost) deterministic transform for time series classification. In Proceedings of the 27th ACM SIGKDD Conference on Knowledge Discovery and Data Mining, 248–57, 2021.

[40] Bai, S.; Kolter, J.Z.; Koltun, V. An empirical evaluation of generic convolutional and recurrent networks for sequence modeling. arXiv 1803.01271, 2018.

[41] He, K.; Zhang, X.; Ren, S.; Sun, J. Deep residual learning for image recognition. arXiv 1512.03385, 2015.

[42] Mallat, S. *A Wavelet Tour of Signal Processing*. 3rd ed. New York: Academic Press, 2009.

[43] Aston, P.J.; Christie, M.I.; Huang, Y.H.; Nandi, M. Beyond HRV: Attractor reconstruction using the entire cardiovascular waveform data for novel feature extraction. *Phys. Meas*. **39**(2):024001, 2018.

[44] Nandi, M.; Venton, J.; Aston, P.J. A novel method to quantify arterial pulse waveform morphology: Attractor reconstruction for physiologists and clinicians. *Phys. Meas.* **39**:104008, 2018.

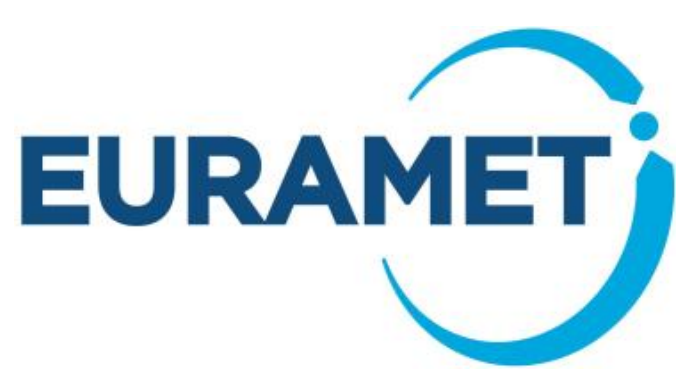

[45] Lyle, J.V.; Aston, P.J. Symmetric Projection Attractor Reconstruction: Embedding in higher dimensions. *Chaos* **31**:113135, 2021.

[46] Vardanega, S.; Segers, P.; Aston, P.J. et al. Attractor image-based deep learning of arterial pulse waves for age classification. *Comp. Cardiol.* **52**:343, 2025.

[47] Mathieu, A.J.W.; Pascual, M.S.; Charlton, P.H. et al. Advanced waveform analysis of the photoplethysmogram signal using complementary signal processing techniques for the extraction of biomarkers of cardiovascular function. *JRSM Cardiovasc. Dis.* **13**:20480040231225384, 2024.

[48] Charlton, P.H.; Celka, P.; Farukh, B.; Chowienczyk, P.; Alastruey, J. Assessing mental stress from the photoplethysmogram: A numerical study. *Phys. Meas.* **39**(5):054001, 2018.

[49] Finnegan, E.; Davidson, S.; Harford, M. et al. Features from the photoplethysmogram and the electrocardiogram for estimating changes in blood pressure. *Scientific Reports* **13**(1), 986, 2023.

[50] Kontaxis, S.; Gil, E.; Marozas, V. et al. Photoplethysmographic waveform analysis for autonomic reactivity assessment in depression. *IEEE Trans. Biomed. Engng*. **68**(4):1273–1281, 2021.

[51] Burns, E; Buttner, R. Atrial fibrillation. Life in the Fast Lane, ECG Library. https://litfl.com/atrial-fibrillation-ecg-library/ Accessed 23/02/2026.

[52] Sörnmo, L.; Petrėnas, A.; Laguna, P,; Marozas, V. Extraction of f waves. In Atrial Fibrillation from an Engineering Perspective . L. Sörnmo (Ed.). Berlin: Springer, 137-220, 2018.

[53] Sološenko, A.; Petrėnas, A.; Paliakaitė, B.; Sörnmo, L.; Marozas, V. Detection of atrial fibrillation using a wrist-worn device. *Phys. Meas*. **40**(2):025003, 2019.

[54] Lord, G.J.; Pardo-Iguzquiza, E.; Smith, I.M. A practical guide to wavelets for metrology. NPL Report CMSC 02/00, 2000.

[55] Park, J.; Seok, H.S.; Kim S.S.; Shin, H. Photoplethysmogram analysis and applications: An integrative review. *Front. Physiol.* **12**:808451, 2022.

[56] Charlton, P.H.; Paliakaitė, B.; Pilt, K. et al. Assessing hemodynamics from the photoplethysmogram to gain insights into vascular age: a review from VascAgeNet. *Am. J. Physiol.-Heart Circul. Physiol.* **322**(4):H493–H522, 2022.

[57] Jia, W.; Sun, M.; Lian, J.; Hou, S. Feature dimensionality reduction: A review. *Compl. Intell. Syst.* **8**:2663-93, 2022.

[58] Zhenyu, Z.; Anand, R.; Wang, M. Maximum relevance and minimum redundancy feature selection methods for a marketing machine learning platform. In Proceedings of the IEEE International Conference on Data Science and Advanced Analytics, 442–452, 2019.

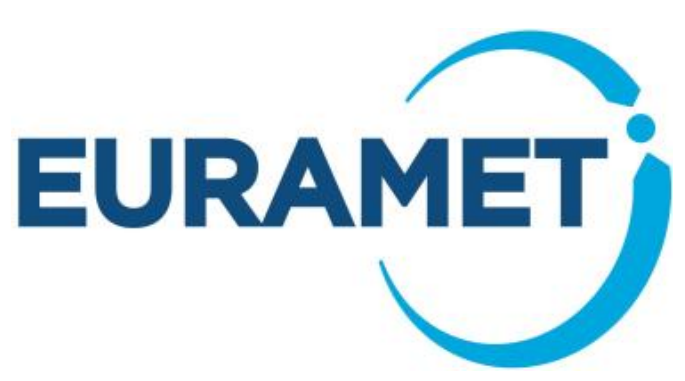

[59] Hassan, A.; Paik, J.H.; Khare, S.; Hassan, S.A. PPFS: Predictive permutation feature selection. arXiv:2110.10713v1, 2021.

[60] Tibshirani, R. Regression shrinkage and selection via the lasso. *J. R. Stat. Soc. B Methodol*. **58**(1):267–288, 1996.

[61] Gewers, F.L.; Ferreira, G.R.; De Arruda, H.F. et al. Principal Component Analysis: A natural approach to data exploration. *ACM Comp. Sur. (CSUR)* **54**(4):70, 2021.

[62] Jutten, C; Herault, J. Blind separation of sources, part I: An adaptive algorithm based on neuromimetic architecture. *Sig. Proc*. **24**:1–10, 1991.

[63] Zhai, J.; Zhang, S.; Chen, J.; He, Q. Autoencoder and its various variants. *Proc. 2018 IEEE Int. Conf. Sys., Man, Cyber. (SMC)*, 415-419, 2018.

[64] Rinkevičius, M.; Charlton, P.H.; Marozas, V. Uncertainty in photoplethysmography-based cuffless blood pressure trend monitoring: A personalized approach. *Comput. Cardiol.* **51**:098, 2024.

[65] Cisnal, A.; Li, Y.; Fuchs, B. et al. Robust feature selection for BP estimation in multiple populations: Towards cuffless ambulatory BP monitoring. *IEEE J. Biomed. Health Inform.* **28**:5768-79, 2024.

[66] Lundberg, S.M.; Lee, S. A unified approach to interpreting model predictions. In Proceedings of the 31st International Conference on Neural Information Processing Systems (NIPS 2017), **30**, 2017.

[67] Ribeiro, M.T.; Singh, S.; Guestrin, C. Why should I trust you?: Explaining the predictions of any classifier. In Proceedings of the 22nd ACM SIGKDD International Conference on Knowledge Discovery and Data Mining, 1135–1144, 2016.

[68] Hastie, T.; Tibshirani, R.; Friedman, J. *The Elements of Statistical Learning: Data Mining, Inference, and Prediction*, Second Edition. Springer Series in Statistics, Springer New York, 2009.

[69] BIPM, IEC, IFCC, ILAC, ISO, IUPAC, IUPAP and OIML. International Vocabulary of Metrology — Basic and general concepts and associated terms (VIM) (3rd edition), Joint Committee for Guides in Metrology, JCGM 200, 2012.

[70] BIPM, IEC, IFCC, ILAC, ISO, IUPAC, IUPAP and OIML. Evaluation of measurement data — Guide to the expression of uncertainty in measurement, Joint Committee for Guides in Metrology, JCGM 100, 2008.

[71] Bilson, S.; Cox, M.; Pustogvar, A.;Thompson, A. A metrological framework for uncertainty evaluation in machine learning classification models. arXiv 2504.03359, 2025.

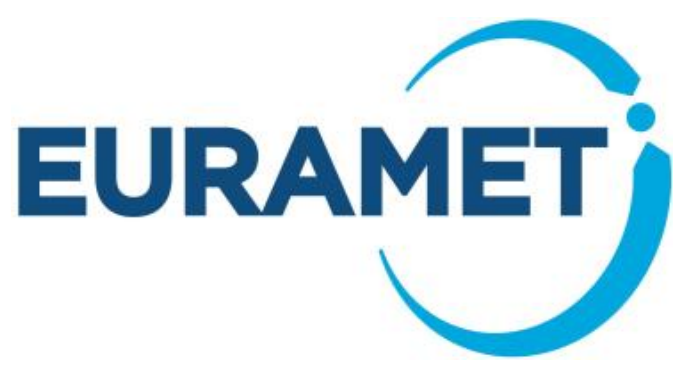

[72] ISO 33406:2024. Approaches for the production of reference materials with qualitative properties. International Organization for Standardization, Geneva, Switzerland, 2024.

[73] Bench, C.; Strodthoff, N.; Moulaeifard, M.; Aston, P.J.; Thompson, A. Towards trustworthy atrial fibrillation classification from wearables data: Quantifying model uncertainty. *Comput. Cardiol.* **51**:068, 2024.

[74] Kendall, A; Gal, Y. What uncertainties do we need in Bayesian deep learning for computer vision? In 31st Conference on Neural Information Processing Systems (NIPS 2017), **30**, 2017. arXiv 1703.04977

[75] Lakshminarayanan, B.; Pritzel, A.; Blundell, C. Simple and scalable predictive uncertainty estimation using deep ensembles. In 31st Conference on Neural Information Processing Systems (NIPS 2017), **30**, 2017.

[76] Gneiting, T.; Raftery, A.E. Strictly proper scoring rules, prediction, and estimation. *J. Amer. Stat. Assoc.* **102**(477):359–78, 2007.

[77] Gneiting, T.; Balabdaoui, F.; Raftery, A.E. Probabilistic forecasts, calibration and sharpness. *J. Roy. Stat. Soc. Series B: Stat. Meth.* **69**(2):243–68, 2007.

[78] Gal, Y.; Ghahramani, Z. Dropout as a Bayesian approximation: Representing model uncertainty in deep learning. In Proceedings of the 33rd International Conference on Machine Learning (PMLR 2016), 1050–59, 2016.

[79] Murphy, K.P. *Probabilistic Machine Learning: An Introduction*. MIT press, 2022.

[80] Mucsányi, B.; Kirchhof, M.; Oh, S.J. Benchmarking uncertainty disentanglement: Specialized uncertainties for specialized tasks. In 38th Conference on Neural Information Processing Systems (NIPS 2024), 1614, 2024

[81] Guo, C.; Pleiss, G.; Sun, Y.; Weinberger. K.Q. On calibration of modern neural networks. In Proceedings of the 34th International Conference on Machine Learning (PMLR 2017), 1321–30, 2017.

[82] Kuleshov, V.; Fenner, N.; Ermon, S. Accurate uncertainties for deep learning using calibrated regression. In Proceeding of the 35th International Conference on Machine Learning (PMLR 2018), 2796–804, 2018.

[83] Zadrozny, B.; Elkan, C. Transforming classifier scores into accurate multiclass probability estimates. In Proceedings of the 8th ACM SIGKDD International Conference on Knowledge Discovery and Data Mining, 694–699, 2002.

[84] Angelopoulos, A.N.; Bates, S. A gentle introduction to conformal prediction and distribution-free uncertainty quantification. arXiv 2107.07511, 2021.

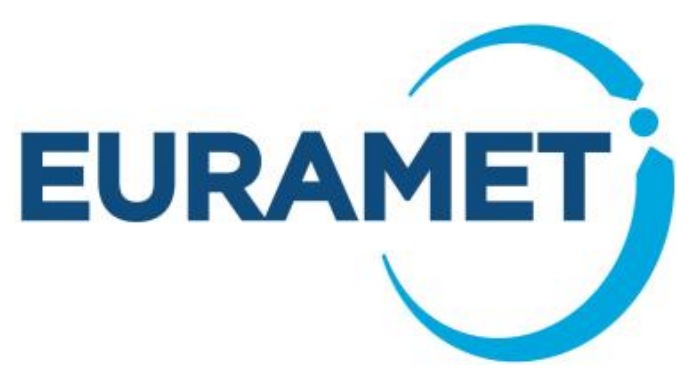

[85] Tibshirani, R. Conformal prediction. In *Advanced Topics in Statistical Learning*, UC Berkeley, 2023. https://www.stat.berkeley.edu/~ryantibs/statlearn-s23/lectures/conformal.pdf

[86] Romano, Y.; Patterson, E.; Candes, E. Conformalized quantile regression. In Proceedings of the 33rd Conference on Neural Information Processing Systems (NIPS 2019), **32**, 2019.

[87] Sadinle, M.; Lei, J.; Wasserman, L. Least ambiguous set-valued classifers with bounded error levels. *J. Amer. Stat. Ass.* **114**:223–234, 2019.

[88] Vovk, V.; Petej, I.; Fedorova, V. Large-scale probabilistic predictors with and without guarantees of validity. In Proceedings of the 29th Conference on Neural Information Processing Systems (NIPS 2015), **28**, 2015.

[89] Pernot, P. Properties of the ENCE and other MAD-based calibration metrics. arXiv 2305.11905, 2023.

[90] Nixon, J.; Dusenberry, M.W.; Zhang, L.; Jerfel, G.; Tran, D. Measuring calibration in deep learning. In Proceedings of the IEEE/CVF Conference on Computer Vision and Pattern Recognition (CVPR) Workshops, 38-41, 2019.

[91] Błasiok, J.; Nakkiran, P. Smooth ECE: Principled reliability diagrams via kernel smoothing. arXiv 2309.12236, 2023.

[92] Błasiok, J.; Gopalan, P.; Hu, L.; Nakkiran, P. A unifying theory of distance from calibration. In Proceedings of the 55th Annual ACM Symposium on Theory of Computing, 1727–1740, New York, NY, USA, 2023.

[93] Laves, M.-H.; Ihler, S.; Kortmann, K.-P.; Ortmaier, T. Uncertainty calibration error: A new metric for multi-class classification, 2021. https://openreview.net/forum?id=XOuAOv_-5Fx

[94] Thompson, A. Extending confidence calibration to generalised measures of variation. arXiv 2602.12975v1, 2026.

[95] Levi, D.; Gispan, L.; Giladi, N.; Fetaya, E. Evaluating and calibrating uncertainty prediction in regression tasks. *Sensors* **22**(15):5540, 2022.

[96] Khosravi, A.; Nahavandi, S.; Creighton, D. A prediction interval-based approach to determine optimal structures of neural network metamodels. *Expert Syst. Appl.* **37**(3):2377–87, 2010.

[97] Grimit, E.P.; Gneiting, T.; Berrocal, V.J.; Johnson, N.A. The continuous ranked probability score for circular variables and its application to mesoscale forecast ensemble verification. *Quart. J. Roy. Met. Soc.* **132**(621C):2925–42, 2006.

[98] Mieloszyk, R.; Twede, H.; Lester, J. et al. A comparison of wearable tonometry, photoplethysmography, and electrocardiography for cuffless measurement of blood pressure in an ambulatory setting. *IEEE J. Biomed. Health Inform.* **26**(7):2864–2875, 2022.

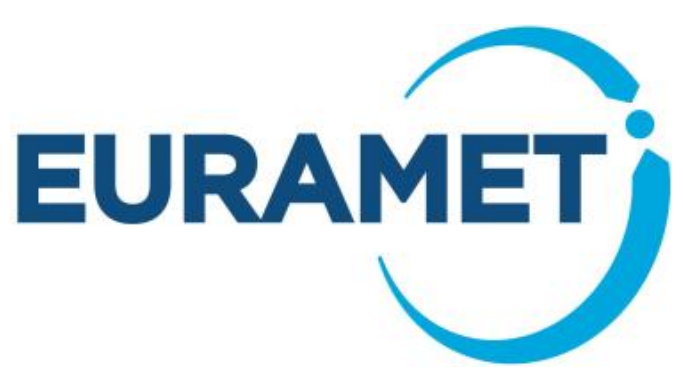

[99] Wang, W.; Mohseni, P.; Kilgore, K.L.; Najafizadeh, L. PulseDB: A large, cleaned dataset based on MIMIC-III and VitalDB for benchmarking cuff-less blood pressure estimation methods. *Front. Dig. Health* **4**:1090854, 2023.

[100] Liu, Z.; Zhou, B.; Jiang, Z. et al. Multiclass arrhythmia detection and classification from photoplethysmography signals using a deep convolutional neural network. *J. Amer. Heart Ass.* **11**(7):e023555, 2022.

[101] Bacevičius, J.; Pluščiauskaitė, V.; Abramikas, Ž. et al. Long-term electrocardiogram and wrist-based photoplethysmogram recordings with annotated atrial fibrillation episodes [Data set]". Zenodo (2024).

[102] Bashar, S.; Han, D.; Hajeb-Mohammadalipour, S. et al. Atrial fibrillation detection from wrist photoplethysmography signals using smartwatches. *Scientific Reports* **9**:15054, 2019.

[103] Han, D.; Bashar, S.K.; Mohagheghian, F. et al. Premature atrial and ventricular contraction detection using photoplethysmographic data from a smartwatch. *Sensors* **20**:5683, 2020.

[104] Moulaeifard, M.; Kutscher, M.; Aston, P.J.; Charlton, P.H.; Strodthoff, N. MIMIC-III-Ext-PPG, a PPG-based benchmark dataset for cardiovascular and respiratory signal analysis. *Scientific Data* **13**:668, 2026.

[105] Johnson, A.E.; Pollard, T.J.; Shen, L. et al. MIMIC-III, a freely accessible critical care database. *Scientific Data* **3**:1–9, 2016.

[106] Bashar, S.K.; Ding, E.; Walkey, A.J.; McManus, D.D.; Chon, K.H. Noise detection in electrocardiogram signals for intensive care unit patients. *IEEE Access* **7**:88357–68, 2019.

[107] Charlton, P.H.; Harana, J.M.; Vennin, S. et al. Modeling arterial pulse waves in healthy aging: A database for in silico evaluation of hemodynamics and pulse wave indexes. *Amer. J. Physiol.-Heart Circul. Physiol.* **317**:H1062–H1085, 2019.

[108] Bernardini, A.; Brunello, A.; Gigli, G.L.; Montanari, A.; Saccomanno, N. OSASUD: A dataset of stroke unit recordings for the detection of Obstructive Sleep Apnea Syndrome. *Scientific Data* **9:**177, 2022.

[109] Chen, X.; Wang, R.; Zee, P. et al. Racial/ethnic differences in sleep disturbances: The Multi-Ethnic Study of Atherosclerosis (MESA). *Sleep* **38**(6):877–88, 2015.

[110] Charlton, P.H. MIMIC PERform Datasets. Zenodo, July 15, 2025. doi: 10.5281/zenodo.15906524.

[111] Martin, D. The impact of skin tone on performance of pulse oximeters used by NHS England COVID Oximetry @home scheme: Measurement and diagnostic accuracy study. *BMJ* **392**:e085535, 2026.

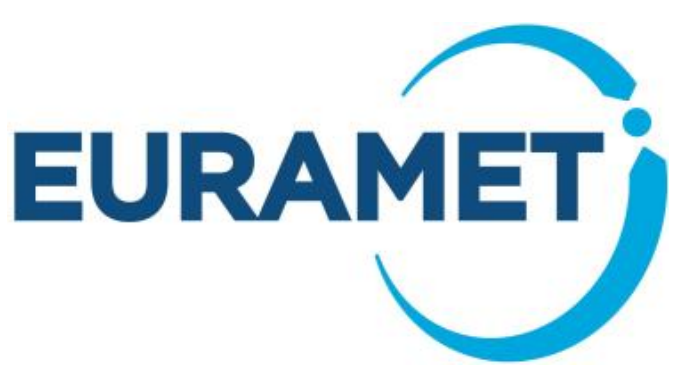

[112] Fitzpatrick, T. The validity and practicality of sun-reactive skin types I through VI. *Arch. Dermatol*. **124**:869–871, 1988.

[113] Krishnadas, P.; Hackstein, U.; Bosnjakovic, A.; Aston, P.J. Fuzzy accuracy compensates for label subjectivity in classification of skin tone using wearable photoplethysmography signals. IEEE International Conference on Fuzzy Systems, 2026.

[114] Lyle, J.V.; Nandi, M.; Aston, P.J. Symmetric Projection Attractor Reconstruction: Sex differences in the ECG. *Front. Cardiovasc. Med.* **8**:709457, 2021.

[115] QUMPHY/D2-code/Repository: https://gitlab.com/qumphy/d2-code

[116] QUMPHY/D1-code/Repository: https://gitlab.com/qumphy/d1-code

[117] Wilkinson, M.D.; Dumontier, M.; Aalbersberg, I.J. et al. The FAIR guiding principles for scientific data management and stewardship. *Scientific Data* **3**:160018, 2016.